\documentclass[nohyperref]{article}

\usepackage{microtype}
\usepackage{graphicx}
\usepackage{subfigure}
\usepackage{booktabs} 

\usepackage{hyperref}

\usepackage[accepted]{icml2023}

\usepackage{amsmath}
\usepackage{amssymb}
\usepackage{mathtools}
\usepackage{amsthm}

\usepackage[capitalize,noabbrev]{cleveref}

\theoremstyle{plain}

\theoremstyle{definition}

\theoremstyle{remark}

\usepackage[textsize=tiny]{todonotes}

\usepackage{xcolor}
\newlength\MAX  
\def\plusbar#1#2{
  +#1 {\color{black!20}\rule{0.5\MAX}{2ex}}{\color{black!60}\rule{0.001mm}{2ex}}{\color{red!100}\rule{#2\MAX}{2ex}}{\color{black!20}\rule{\MAX - #2\MAX -0.5\MAX}{2ex}}
}
\def\minusbar#1#2{
  #1 {\color{black!20}\rule{0.5\MAX + #2\MAX}{2ex}}{\color[rgb]{0,0.65,0.35}\rule{- #2\MAX}{2ex}}{\color{black!60}\rule{0.001mm}{2ex}}{\color{black!20}\rule{0.5\MAX}{2ex}}
}

\begin{document}

\twocolumn[

\icmltitle{Beyond In-Domain Scenarios: Robust Density-Aware Calibration}

\icmlsetsymbol{equal}{*}

\begin{icmlauthorlist}
\icmlauthor{Christian Tomani}{equal,tum,mcml}
\icmlauthor{Futa Waseda}{equal,tut}
\icmlauthor{Yuesong Shen}{tum,mcml}
\icmlauthor{Daniel Cremers}{tum,mcml}
\end{icmlauthorlist}

\icmlaffiliation{tum}{Technical University of Munich}
\icmlaffiliation{tut}{The University of Tokyo}
\icmlaffiliation{mcml}{Munich Center for Machine Learning}

\icmlcorrespondingauthor{Christian Tomani}{christian.tomani@tum.de}
\icmlcorrespondingauthor{Futa Waseda}{futa-waseda@g.ecc.u-tokyo.ac.jp}

\icmlkeywords{Machine Learning, Calibration, Uncertainty, Trustworthiness, ICML}

\vskip 0.3in
]

\printAffiliationsAndNotice{\icmlEqualContribution}

\begin{abstract}
Calibrating deep learning models to yield uncertainty-aware predictions is crucial as deep neural networks get increasingly deployed in safety-critical applications. While existing post-hoc calibration methods achieve impressive results on in-domain test datasets, they are limited by their inability to yield reliable uncertainty estimates in domain-shift and out-of-domain (OOD) scenarios. We aim to bridge this gap by proposing \textbf{DAC}, an accuracy-preserving as well as \textbf{D}ensity-\textbf{A}ware \textbf{C}alibration method based on k-nearest-neighbors (KNN). In contrast to existing post-hoc methods, we utilize hidden layers of classifiers as a source for uncertainty-related information and study their importance. We show that DAC is a generic method that can readily be combined with state-of-the-art post-hoc methods. DAC boosts the robustness of calibration performance in domain-shift and OOD, while maintaining excellent in-domain predictive uncertainty estimates. We demonstrate that DAC leads to consistently better calibration across a large number of model architectures, datasets, and metrics. Additionally, we show that DAC improves calibration substantially on recent large-scale neural networks pre-trained on vast amounts of data.
\end{abstract}

\section{Introduction}

Deep learning models have become state-of-the-art (SOTA) in several different fields. Especially in safety-critical applications such as medical diagnosis and autonomous driving with changing environments over time, reliable model estimates for predictive uncertainty are crucial. Thus, models are required to be accurate as well as calibrated, meaning that their predictive uncertainty (or confidence) matches the expected accuracy.
Due to the fact that many deep neural networks are generally uncalibrated \cite{guo_calibration_2017}, post-hoc calibration of already trained neural networks has received increasing attention in the last few years.

\begin{figure}[t]
\centering
\begin{tabular}{cc}
\includegraphics[width=0.17\textwidth]{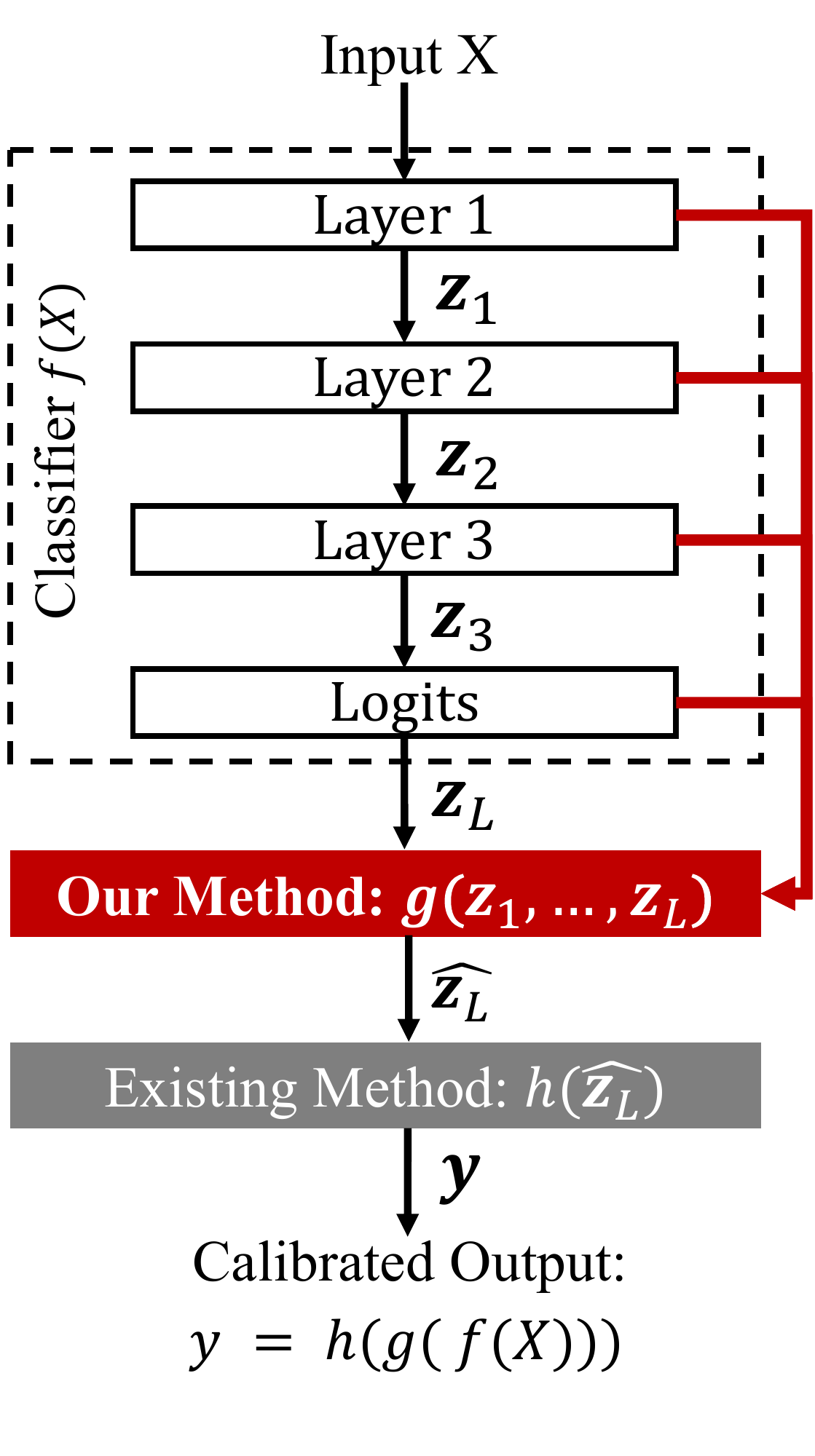} &
\includegraphics[width=0.18\textwidth]{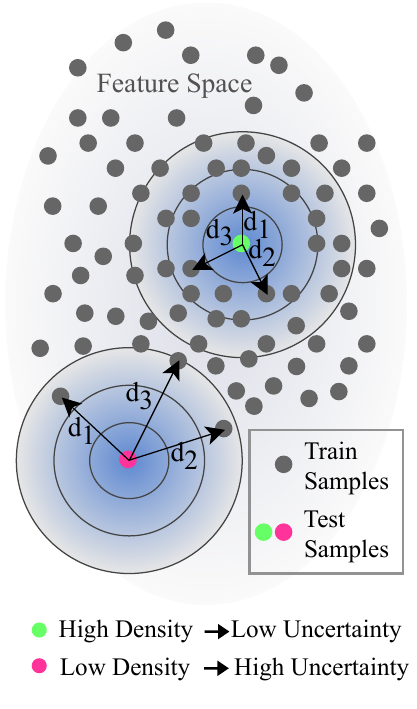} \\
\end{tabular}
\caption{\textbf{Left:} Our density-aware calibration method (DAC) $g$ can be combined with existing post-hoc methods $h$ leading to robust and reliable uncertainty estimates.  
To this end, DAC leverages information from feature vectors $z_1...z_L$ across the entire classifier $f$.
\textbf{Right:} DAC is based on KNN, where predictive uncertainty is expected to be high for test samples lying in low-density regions of the empirical training distribution and vice versa.}
\label{fig:teaser_fig}
\end{figure}

In order to tackle the miscalibration of neural networks, researchers have come up with a plethora of post-hoc calibration methods \cite{guo_calibration_2017,zhang2020mix,rahimi2020intra,milios2018dirichlet,tomani2022parameterized,gupta2020calibration}. These current approaches are particularly designed for in-domain calibration, where test samples are drawn from the same distribution as the network was trained on. Although these approaches perform almost perfect in-domain, recent works \cite{tomani2021post} have shown that they lack substantially in providing reliable confidence scores in domain-shift and out-of-domain (OOD) scenarios (up to 1 order of magnitude worse than in-domain), which is unacceptable, particularly in safety-critical real-world scenarios where calibration of neural networks matters in order to prevent unforeseeable failures. 
To date, the only post-hoc methods that have been introduced to mitigate this shortcoming in domain-shift and OOD settings, use artificially created data or data from different sources in order to estimate the potential test distribution \cite{tomani2021post,yu2022robust,wald2021calibration,yu2022robust,gong2021confidence}. However, these methods are not generic in that they require domain knowledge about the dataset and utilize multiple domains for calibration. Additionally, they might only work well for a narrow subset of anticipated distributional shifts because they rely heavily on strong assumptions towards the potential test distribution. Furthermore, they can hurt in-domain calibration performance.

To mitigate the issue of miscalibration in scenarios where test samples are not necessarily drawn from dense regions of the empirical training distribution or are even OOD, we introduce a density-aware method that extends the field of post-hoc calibration beyond in-domain calibration. Contrary to the aforementioned existing works which focus on particularly crafted training data, our method DAC does not depend on additional data and does not rely on any assumptions about potentially shifted or out-of-domain test distributions, and is even domain agnostic. The proposed method can, therefore, simply be added to an existing post-hoc calibration pipeline, because it relies on the exact same training paradigm with a held-out in-domain validation set as current post-hoc methods do.

Previous works on calibration have focused primarily on post-hoc methods that solely take softmax outputs or logits into account \cite{guo_calibration_2017,zhang2020mix,rahimi2020intra,milios2018dirichlet,tomani2022parameterized,gupta2020calibration}. However, we argue that prior layers in neural networks contain valuable information for recalibration too. 
Moreover, we report, which layers our method identified as particularly relevant for providing well-calibrated predictive uncertainty estimates. 

Recently developed large-scale neural networks that benefit from pre-training on vast amounts of data \cite{kolesnikov2020big, wslimageseccv2018}, have mostly been overlooked when benchmarking post-hoc calibration methods. 
One explanation for that could be because, e.g., vision transformers (ViTs) \cite{dosovitskiy2020image} are well calibrated out of the box \cite{minderer2021revisiting}. Nevertheless, we show that also these models can profit from post-hoc methods and in particular from DAC through more robust uncertainty estimates. 

\subsection{Contribution}
\begin{itemize} 
    \item We propose DAC, an accuracy-preserving and density-aware calibration method that can be combined with existing post-hoc methods to boost domain-shift and out-of-domain performance while maintaining in-domain calibration.\footnote{Source code available at: \url{https://github.com/futakw/DensityAwareCalibration}} 
    \item We discover that the common practice of using solely the final logits for post-hoc calibration is sub-optimal and that aggregating intermediate outputs yields improved results. 
    \item We study recent large-scale models, such as transformers, pre-trained on vast amounts of data and encounter that also for these models our proposed method yields substantial calibration gains.  
\end{itemize}

\section{Related Work}

Calibration methods can be divided into post-hoc calibration methods and methods that adapt the training procedure of the classifier itself. The latter one includes methods such as self-supervised learning \cite{hendrycks2019using}, Bayesian neural networks \cite{gal2016dropout, wen2018flipout}, Deep Ensembles \cite{lakshminarayanan2017simple}, label smoothing \cite{muller2019does}, methods based on synthesized feature statistics \cite{wang2020transferable} or mixup techniques \cite{thulasidasan2019mixup,zhang2017mixup} as well as other intrinsically calibrated approaches \cite{sensoy_evidential_2018, tomani2021falcon, ashukha2020pitfalls}. Similar to post-hoc methods, Ovadia et al. \yrcite{snoek2019can} have found that intrinsic methods suffer as well from miscalibration in domain-shift scenarios.

Post-hoc calibration methods, on the other hand, can be applied on top of already trained classifiers and do not require any retraining of the underlying neural network. Wang et al. \yrcite{wang2021rethinking} argue in favour of a unified framework comprising of main training and post-hoc calibration. Rahimi et al. \yrcite{rahimi2020post} provide a theoretical basis for post-hoc calibration schemes in that learning calibration functions post-hoc using a proper loss function leads to calibrated outputs.

These post-hoc calibration methods include non-parametric approaches such as histogram binning \cite{zadrozny2001obtaining}, where uncalibrated confidence scores are partitioned into bins and are assigned a respective calibrated score via optimizing a bin-wise squared loss on a validation set. Isotonic regression \cite{zadrozny2002transforming}, which is an extension to histogram binning, fits a piecewise constant function to intervals of uncalibrated confidence scores and Bayesian Binning into Quantiles (BBQ) \cite{naeini2015obtaining} is different from isotonic regression in that it considers multiple binning models and their combination. In addition, Zhang et al. \yrcite{zhang2020mix} introduce an accuracy-preserving version of isotonic regression beyond binary tasks, which they call multi-class isotonic regression (IRM). Moreover, Wenger et al. \yrcite{wenger2020non} and Milios et al. \yrcite{milios2018dirichlet} propose Gaussian processes-based calibration methods.

Approaches for training a mapping function include Platt scaling \cite{Platt99probabilisticoutputs}, matrix as well as vector scaling and temperature scaling \cite{guo_calibration_2017}. 
Temperature scaling (TS) transforms logits by a single scalar parameter in an accuracy-preserving manner, since re-scaling does not affect the ranking of the logits. Moreover, Ensemble Temperature scaling (ETS) \cite{zhang2020mix} extends temperature scaling by two additional calibration maps with a fixed temperature of 1 and $\infty$, respectively. 
More recent and advanced approaches include Dirichlet-based scaling \cite{kull2019beyond} and Parameterized Temperature scaling \cite{tomani2022parameterized}, where a temperature is calculated sample-wise via a neural network architecture. Rahimi et al. \yrcite{rahimi2020intra} designed a post-hoc neural network architecture for transforming classifier logits that represent a class of intra-order-preserving functions, and Gupta et al. \yrcite{gupta2020calibration} introduce a method for obtaining a calibration function by approximating the empirical cumulative distribution of output probabilities with the help of splines.

These post-hoc calibration methods are trained on a hold-out calibration set. Although there has been a surge of research on these approaches in recent years, Tomani et al. \cite{tomani2021post} have discovered that post-hoc calibration methods yield highly over-confident predictions under domain-shift and are, therefore, not well suited for OOD scenarios. They introduce a strategy where samples are perturbed in the calibration set before performing the post-hoc calibration step. However, such an approach makes a strong distributional assumption on potential domain shifts during testing by perturbing training samples in a particular way, which may not necessarily hold in each case. To date, post-hoc calibration methods that are themselves capable of distinguishing in-domain samples from  gradually shifted or out-of-domain samples without any distributional assumptions have not yet been addressed.

\section{Method}

\subsection{Definitions}

We study the multi-class classification problem, where $X \in \mathbb{R}^D$ denotes a $D$-dimensional random input variable and $Y \in \{1,2,\dots, C\}$ denotes the label with $C$ classes with a ground truth joint distribution $\pi(X, Y ) =
\pi(Y |X)\pi(X)$. The dataset $\mathbb{D}$ contains $N$ i.i.d. samples $\mathbb{D} = \{(X_n, Y_n)\}_{n=1}^N$ drawn from  $\pi(X, Y)$. 

Let the output of a trained neural network classifier $f$ be $f(X) = (y,\mathbf{z}_L)$, where $y$ denotes the predicted class and $\mathbf{z}_L$ the associated logits vector. The softmax function $\sigma_{SM}$ as $p=\max_c \sigma_{SM}(\mathbf{z}_L)^{(c)}$ is then needed to transform $\mathbf{z}_L$ into a confidence score or predictive uncertainty $p$ w.r.t $y$. In this paper, we propose an approach to improve the quality of the predictive uncertainty $p$ by recalibrating the logits $\mathbf{z}_L$ from $f(X)$ via a combination of two calibration methods: 
\begin{eqnarray}
p=h(g(f(X)))
\label{eq:combine_cali}
\end{eqnarray}
where $g$ denotes our density-aware calibration method DAC rescaling logits for boosting domain-shift and OOD calibration performance and $h$ denotes an existing state-of-the-art in-domain post-hoc calibration method (Fig.~\ref{fig:teaser_fig}).  

Following Guo et al.~\yrcite{guo_calibration_2017}, perfect calibration is defined such that confidence and accuracy match for all confidence levels:
\begin{eqnarray}
\mathop{\mathbb{P}}(Y=y| P=p) = P, \; \; \forall P \in [0,1]
\label{eq:cali}
\end{eqnarray}
Consequently, miscalibration is defined as the difference in expectation between accuracy and confidence.
\begin{eqnarray}
\mathop{\mathbb{E}}_{P}\left[\big\lvert\mathop{\mathbb{P}}(Y=y| P=p) - P\big\rvert \right]
\label{eq:miscal}
\end{eqnarray}

\subsection{Measuring Calibration} 
The expected calibration error (ECE) \cite{naeini2015obtaining} is frequently used for quantifying miscalibration. ECE is a scalar summary measure estimating miscalibration by approximating equation \eqref{eq:miscal} as follows. 
In the first step, confidence scores $\hat{\mathcal{P}}$ of all samples are partitioned into $M$ equally sized bins of size $1/M$, and secondly, for each bin $B_m$ the respective mean confidence and the accuracy is computed based on the ground truth class $y$. Finally, the ECE is estimated by calculating the mean difference between confidence and accuracy over all bins: 
\begin{eqnarray}
\mathrm{ECE}^d = \sum_{m=1}^M \frac{\lvert B_m\rvert}{N}\left\| \mathrm{acc}(B_m) - \mathrm{conf}(B_m)\right\|_d
\end{eqnarray}\label{eq:ece}
with $d$ usually set to 1 (L1-norm).

\subsection{Density-Aware Calibration (DAC)}
\label{sec:DAC}
Our main idea for our proposed calibration method $g$ stems from the fact that test samples lying in high-density regions of the empirical training distribution can generally be predicted with higher confidence than samples lying in low-density regions. 
For the latter case, the network has seen very few, if any, training samples in the neighborhood of the respective test sample in feature space, and is thus not able to provide reliable predictions for those samples. 
Leveraging this information about density through a proxy can result in better calibration.

In order to estimate such a proxy for each sample, we propose to utilize non-parametric density estimation using k-nearest-neighbor (KNN) based on feature embeddings extracted from the classifier. 
KNN has successfully been applied in out-of-distribution detection \cite{sun2022out}. In contrast to Sun et al. \yrcite{sun2022out}, who only take the penultimate layer into account, we argue that prior layers yield important information too, and therefore, incorporate them in our method as follows. We call our method Density-Aware Calibration (DAC).

Temperature scaling \cite{guo_calibration_2017} is a frequently used calibration method, where a single scalar parameter $T$ is used to re-scale the logits of an already trained classifier in order to obtain calibrated probability estimates $\hat{Q}$ for logits $\mathbf{z}_{L}$ using the softmax function $\sigma_{SM}$ as 
\begin{equation}
\hat{Q} = \sigma_{SM}(\mathbf{z}_{L}/T)   
\label{eq:ts}
\end{equation}
Similar to temperature scaling, our method is also accuracy preserving, in that we use one parameter $S(\mathbf{x},w)$ for re-scaling the logits of the classifier: 
\begin{eqnarray}
    \hat{Q}(\mathbf{x},w) = \sigma_{SM}(\mathbf{z}_{L}/S(\mathbf{x},w))
\label{eq:qscore}
\end{eqnarray} 
In contrast to temperature scaling, in our case, $S(\mathbf{x},w)$ is sample-dependent with respect to $\mathbf{x}$ 
and is calculated via a linear combination of density estimates $s_l$ as follows: 
\begin{equation}
S(\mathbf{x},w) = \sum_{l=1}^{L}w_ls_l+w_0  
\label{eq:weighting_scheme}
\end{equation}
with $w_1...w_L$ being the weights for every layer in $L$ and $w_0$ being a bias term. 
Note that only positive weights are valid because negative weights would assign high confidence to outliers.
Thus, we constrain the weights to be positive to tackle overfitting.

For each feature layer $l$, we compute the density estimate $s_l$ in the neighborhood of the empirical training distribution of the respective test sample $\mathbf{x}$ with the $k$-th nearest neighbor distance: 
First, we derive the test feature vector $\mathbf{z}_l$ from the trained classifier $f$ given the test input sample $\mathbf{x}$ and average over spatial dimensions as well as normalize it. We then use the normalized training feature vectors $Z_{N_{Tr},l}=(\mathbf{z}_{1,l},\mathbf{z}_{2,l},...,\mathbf{z}_{N_{Tr},l})$, which we gathered from the training dataset $X_{N_{Tr}}=(\mathbf{x}_1,\mathbf{x}_2,...,\mathbf{x}_{N_{Tr}})$ to calculate the euclidean distance between $\mathbf{z}_l$ and each element in $Z_{N_{Tr},l}$ for each sample $i$ in the training set $N_{Tr}$ as follows: 
\begin{equation}
d_{i,l} = \|\mathbf{z}_{i,l}-\mathbf{z}_l\|   
\label{eq:eucledian}
\end{equation}
The resulting sequence $D_{N_{Tr},L}=(d_{1,l},d_{2,l},...,d_{N_{Tr},l})$ is reordered. Finally, $s_l$ is given by the $k$-th smallest element (=$k$-th nearest neighbor) in the sequence: $s_l=d_{(k)}$, with $(k)$ indicating the index in the reordered sequence $D_{N_{Tr},L}$. For determining $k$, we follow Sun et al. \yrcite{sun2022out}, who did a thorough analysis and concluded that a proper choice for $k$ is 50 for CIFAR10, 200 for CIFAR100 for all training samples and 10 for ImageNet for 1\% training samples.

We fit our method for a trained neural network $f(X)=(y,\mathbf{z}_L)$ by optimizing a squared error loss $L_w$ w.r.t. $w$. 
\begin{eqnarray}
L_w = \sum_{c=1}^C(I_{c}-\sigma_{SM}(\mathbf{z}_{L}/S(\mathbf{x},w))^{(c)})^2
\label{eq:lossece}
\end{eqnarray}
where $I_{c}$ indicates a binary variable, which is $1$ if the respective sample has true class $c$, and $0$ otherwise. We accumulate $L_w$ over all samples in the validation set.\\

The rescaled logits $\mathbf{\hat{z}}_{L}(\mathbf{x},w)=\mathbf{z}_{L}/S(\mathbf{x},w)$ and consequently the recalibrated probability estimates $\hat{Q}(\mathbf{x},\mathbf{w})$ can directly be fed to another post-hoc method. Thus, DAC can be applied prior to other existing in-domain post-hoc calibration methods for robustly calibrating models in domain-shift and OOD scenarios (Fig.~\ref{fig:teaser_fig}). 

DAC uses KNN (a non-parametric method) to compute a density proxy per layer and combines these proxies linearly across layers, whereas other methods like Parameterized Temperature scaling \cite{tomani2022parameterized} and models that utilize intra-order-preserving functions for calibration \cite{rahimi2020intra} are parametric methods using a neural network. Moreover, DAC uses intermediate features from hidden layers and is particularly designed with domain shift and OOD calibration behavior in mind. 

Our method has the following advantages: 
\begin{itemize}
    \item \textbf{Density aware:} Due to distance-based density estimation across feature layers, our method is capable of inferring how close or how far a test sample is in feature space with respect to the training distribution and can adjust the predictive estimates of the classifier accordingly. 
    \item \textbf{Domain agnostic:} Since we use KNN, a non-parametric method for density estimation, no distributional assumptions are imposed on the feature space, and our method is therefore applicable to any type of in-domain, domain-shift, or OOD scenario. 
    \item \textbf{Backbone agnostic:} Adapts easily to different underlying classifier architectures (e.g., CNNs, ResNets, and more recent models like transformers) 
    because during training, DAC automatically figures out the informative feature layers regarding uncertainty calibration.
\end{itemize}

\begin{table*}
\vskip 0.15in
\caption{Mean expected calibration error across all test domain-shift scenarios. For each model, the macro-averaged ECE ($\times 10^2$) (with equal-width binning and 15 bins) is computed across all corruptions from severity=0 (in-domain) until severity=5 (heavily corrupted). Post-hoc calibration methods paired with our method are consistently better calibrated than simple post-hoc methods. (lower ECE is better)}
\label{tab:mean_ece}
\begin{center}
\begin{small}
\begin{sc}
\begin{tabular}{l|c|ccccc|ccccc}
    \toprule
    & \multicolumn{1}{|c|}{\bfseries Uncal} &
    \multicolumn{5}{c|}{\bfseries Baseline Calibration Methods} &
    \multicolumn{5}{c}{\bfseries Combination with DAC (Ours)} \\
    & - & TS & ETS & IRM & DIA & SPL & TS & ETS & IRM & DIA & SPL\\
    \midrule
        C10 ResNet18 & 19.27 &          4.96 &             4.96 &  5.93 &  6.39 &             6.27 & \underline{4.49} &    \textbf{4.39} &             4.90 &          4.61 &             4.74 \\
       C10 VGG16 & 19.05 &          6.25 &             6.30 &  7.45 &  9.82 &             7.57 &    \textbf{5.66} &    \textbf{5.66} &             6.30 &          6.33 &             5.69 \\
 C10 DenseNet121 & 19.26 &          5.21 &             5.21 &  6.66 &  7.81 &             6.60 & \underline{4.59} & \underline{4.59} &             5.69 &          6.60 &    \textbf{4.42} \\
    \hline
   C100 ResNet18 & 16.44 &         11.37 &            10.56 & 12.26 &  9.24 &            10.40 &            10.66 & \underline{8.96} &            10.04 & \textbf{8.47} &             9.74 \\
      C100 VGG16 & 34.41 &         11.54 &            11.54 & 13.24 & 14.62 &            10.76 & \underline{6.49} &    \textbf{6.48} &             8.11 &         10.26 &             7.45 \\
C100 DenseNet121 & 23.83 &          8.80 & \underline{8.76} & 12.07 & 11.02 &             9.73 &             8.77 &    \textbf{8.40} &            10.05 &         15.01 &             9.15 \\
\hline
   IMG ResNet152 & 10.50 &          4.47 &             4.01 &  5.20 &  7.17 &             5.56 & \underline{3.48} &    \textbf{3.34} &             3.50 &          3.64 &             3.63 \\
 IMG DenseNet169 & 13.28 &          6.59 &             6.34 &  7.37 &  8.44 &             7.12 &             4.81 &    \textbf{3.87} & \underline{4.53} &          6.31 &             4.60 \\
    IMG Xception & 30.49 &          8.81 &             8.40 & 12.93 &  9.83 &            10.80 &             8.79 &    \textbf{7.99} & \underline{8.38} &          8.99 &             8.49 \\
    \hline
       IMG BiT-M & 11.71 &          7.17 &             6.56 &  6.93 &  7.45 &             6.62 &             4.40 & \underline{3.98} &             4.21 &          5.51 &    \textbf{3.76} \\
 IMG ResNeXt-WSL & 15.44 &          8.03 &             8.03 &  8.04 &  8.32 &             6.16 &             7.32 & \underline{5.63} &             5.75 &          6.32 &    \textbf{3.90} \\
       IMG ViT-B &  3.78 &          4.23 &             3.72 &  4.24 &  5.85 &             3.93 &             3.80 &    \textbf{3.34} &             3.99 &          5.52 & \underline{3.56} \\
    \bottomrule
\end{tabular}
\end{sc}
\end{small}
\end{center}
\vskip -0.1in
\end{table*}

\section{Experimental Setup} \label{sec:exp_setup}

\paragraph{Models and Datasets} 
In our study, we quantify the performance of our proposed model for various model architectures and different datasets. We consider 3 different datasets to evaluate our models on CIFAR10/100 \cite{krizhevsky2009learning_cifar}, and ImageNet-1k \cite{deng2009imagenet}. In particular, for measuring performance on CIFAR10 and CIFAR100, we train ResNet18 \cite{he2016deep_resnet}, VGG16 \cite{simonyan2014very_vgg} and DenseNet121 \cite{huang2017densely_densenet}, and for ImageNet, we use 3 pre-trained models, namely ResNet152 \cite{he2016deep_resnet}, DenseNet169 \cite{huang2017densely_densenet}, and Xception\cite{chollet2017xception}. 
We further investigate post-hoc calibration methods applied to new state-of-the-art architectures as well as modern training schemes. To this end, we include the following models in our study, which are all finetuned on ImageNet-1k: 

\begin{itemize}
    \item \textbf{BiT-M} \cite{kolesnikov2020big}: Is a ResNet-based architecture (ResNetV2) \cite{he2016identity_resnetV2} pre-trained on ImageNet-21k.
    \item \textbf{ResNeXt-WSL} \cite{wslimageseccv2018}: Is a ResNeXt-based architecture (ResNeXt101 32x8d) \cite{xie2017aggregated_resnext}, which is weakly supervised pre-trained with billions of hashtags of social media images.
    \item \textbf{ViT-B} \cite{dosovitskiy2020image_vit}: Is a transformer-based architecture pre-trained on ImageNet-21k.
\end{itemize}

We quantify calibration performance for in-domain, domain-shift, and OOD scenarios. In order to ensure a gradual domain shift in our evaluation pipeline, we use ImageNet-C as well as CIFAR-C \cite{hendrycks2019benchmarking_imagenet_c}, which were specifically developed to produce domain shift and were incorporated in many related studies since. Both datasets have 18 distinct corruption types, each having 5 different levels of severity, mimicking a scenario where the input data to a classifier gradually shifts away from the training distribution.
Additionally, we test our models on a real-world OOD dataset, namely ObjectNet-OOD. ObjectNet \cite{barbu2019objectnet} is a dataset consisting of 50,000 test images with a total of 313 classes, of which 200 classes are out-of-domain with respect to ImageNet. Hence, we make use of these 200 classes for our OOD analysis. 

\paragraph{Post-hoc Calibration Methods}
We consider the currently best performing post-hoc calibration methods for benchmarking as well as for combining them with DAC: Temperature scaling (TS) \cite{guo_calibration_2017},  Ensemble Temperature scaling (ETS) \cite{zhang2020mix}, accuracy preserving version of Isotonic Regression (IRM) \cite{zhang2020mix}, Intra-order preserving calibration (DIA) \cite{rahimi2020intra}, Calibration using Splines (SPL) \cite{gupta2020calibration}. Additionally, we show results for Isotonic Regression (IR) \cite{zadrozny2002transforming}, Parameterized Temperature scaling (PTS) \cite{tomani2022parameterized} and Dirichlet calibration (DIR) \cite{kull2019beyond} in Appendix~\ref{appendix:additional_methods}.  

Our proposed method DAC does not solely rely on logits for calibration as other post-hoc calibration approaches do; it rather takes various layers at certain positions of the network into account. Even though DAC could use every layer in a classifier due to its weighting scheme, we opt for a much simpler and faster version. 
That is, we follow a structured approach for choosing layers, e.g., after each ResNet or transformer block. 
A detailed description of which layers we use can be found in Appendix~\ref{appendix:DAC}. In the results, we show that our selective approach produces similar results compared to taking all layers into account.

\paragraph{Measuring Calibration}
Our evaluation is based on various calibration measures. Throughout the paper, we provide results for ECE and Brier scores using equal-width binning with 15 bins. Although ECE is the most commonly used metric for evaluating and comparing post-hoc calibration methods, it bears several limitations. That is why, in Appendix~\ref{appendix:additional_measures}, we show that our results hold for different kinds of calibration measures, including ECE based on kernel density estimation (ECE-KDE) \cite{zhang2020mix}, ECE using equal-mass binning and class-wise ECE\cite{kull2019beyond} as well as we demonstrate consistency with likelihood. 

\setlength{\tabcolsep}{0.5pt} 
\renewcommand{\arraystretch}{1} 
\begin{figure*}[t!]
    \centering
    \begin{tabular}{cccc}
    \includegraphics[width=0.47\textwidth]{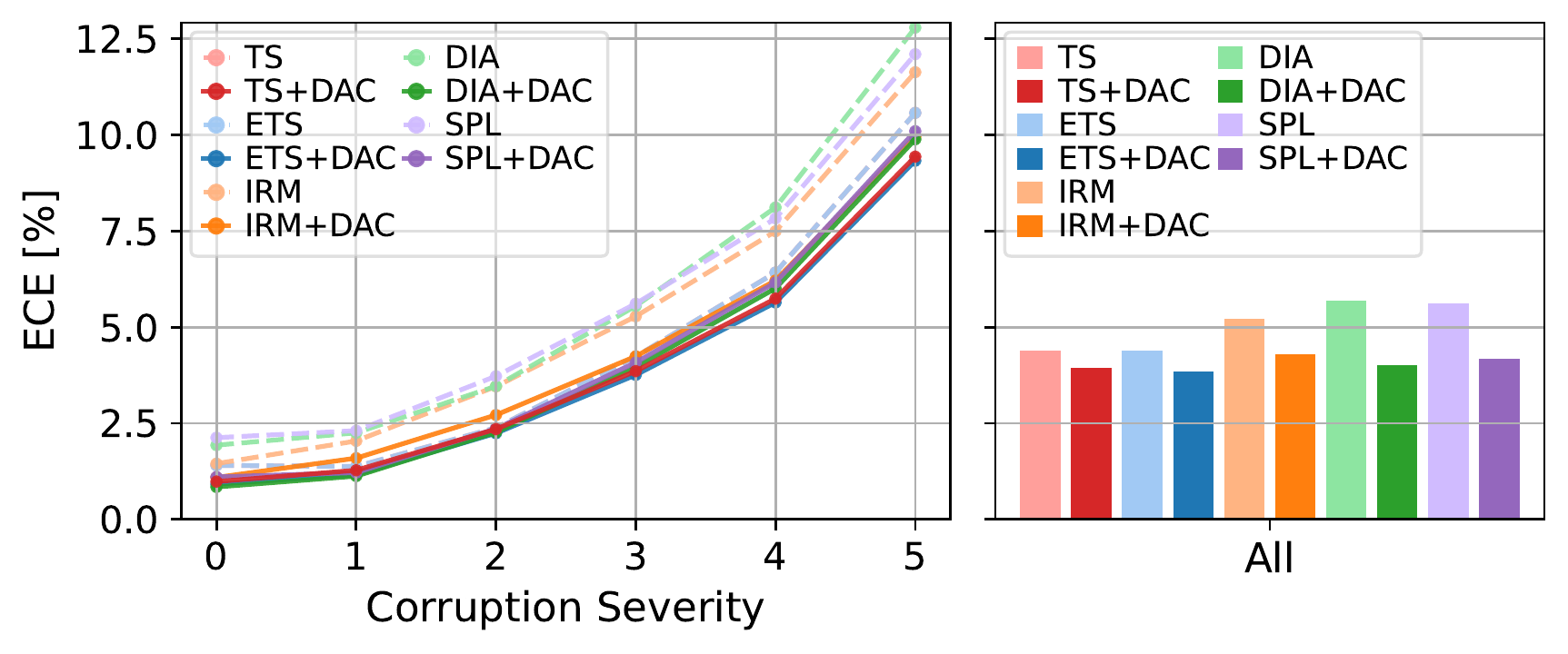} &
    \includegraphics[width=0.47\textwidth]{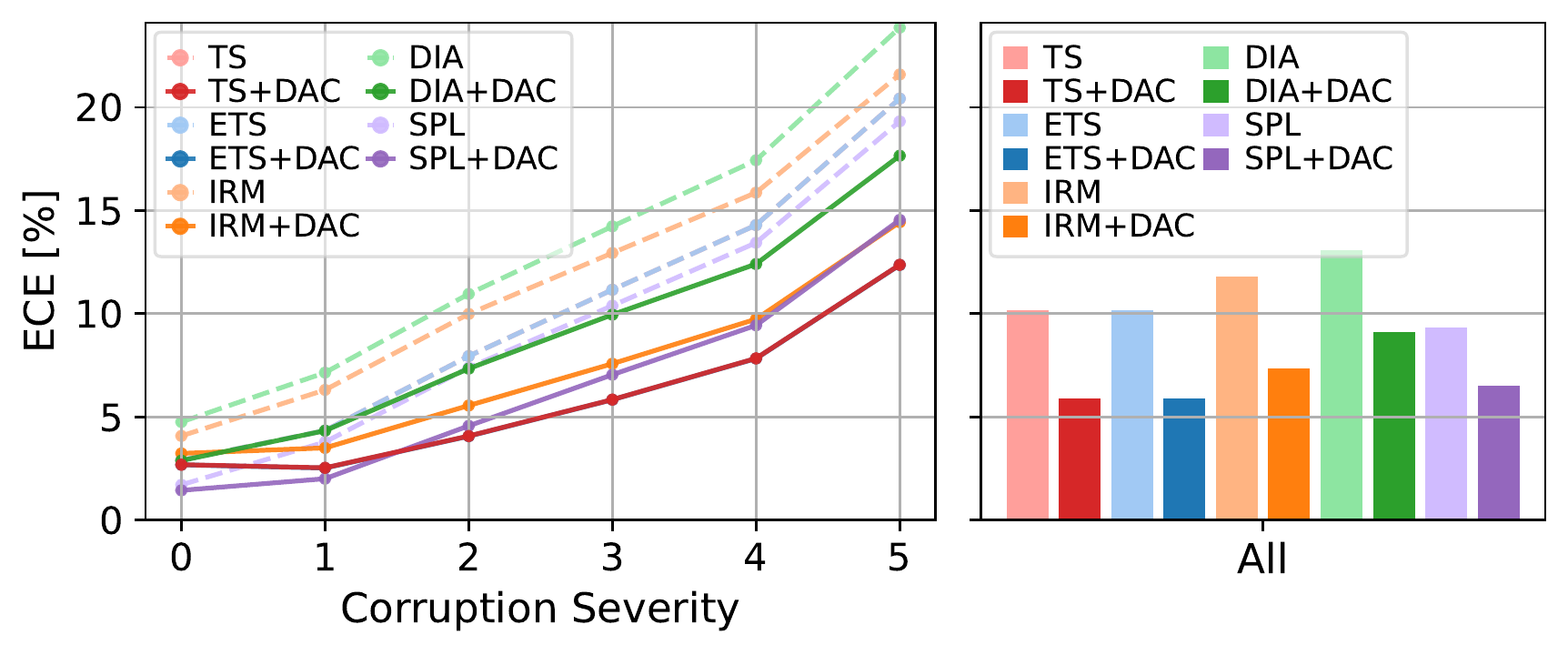} \\  
    {CIFAR10-ResNet18} & {CIFAR100-VGG16}\\  

    \includegraphics[width=0.47\textwidth]{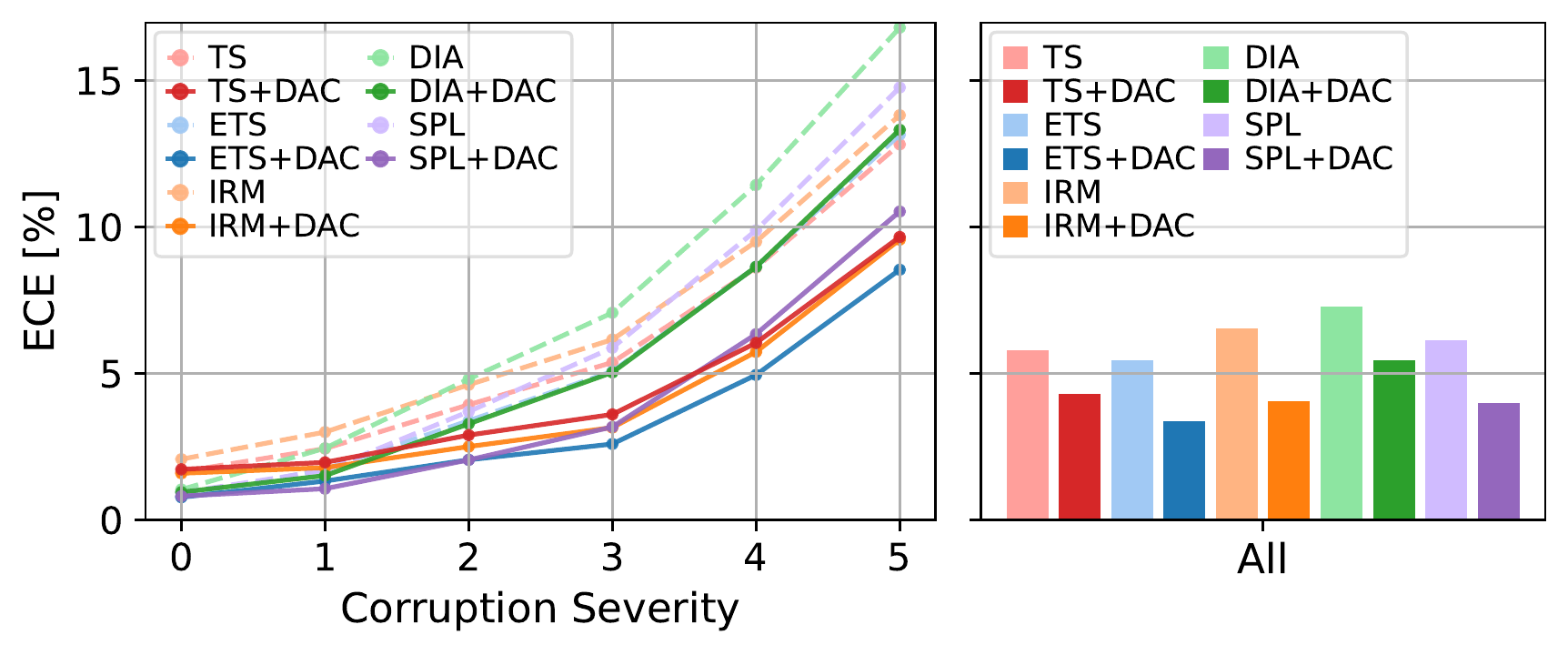} &
    \includegraphics[width=0.47\textwidth]{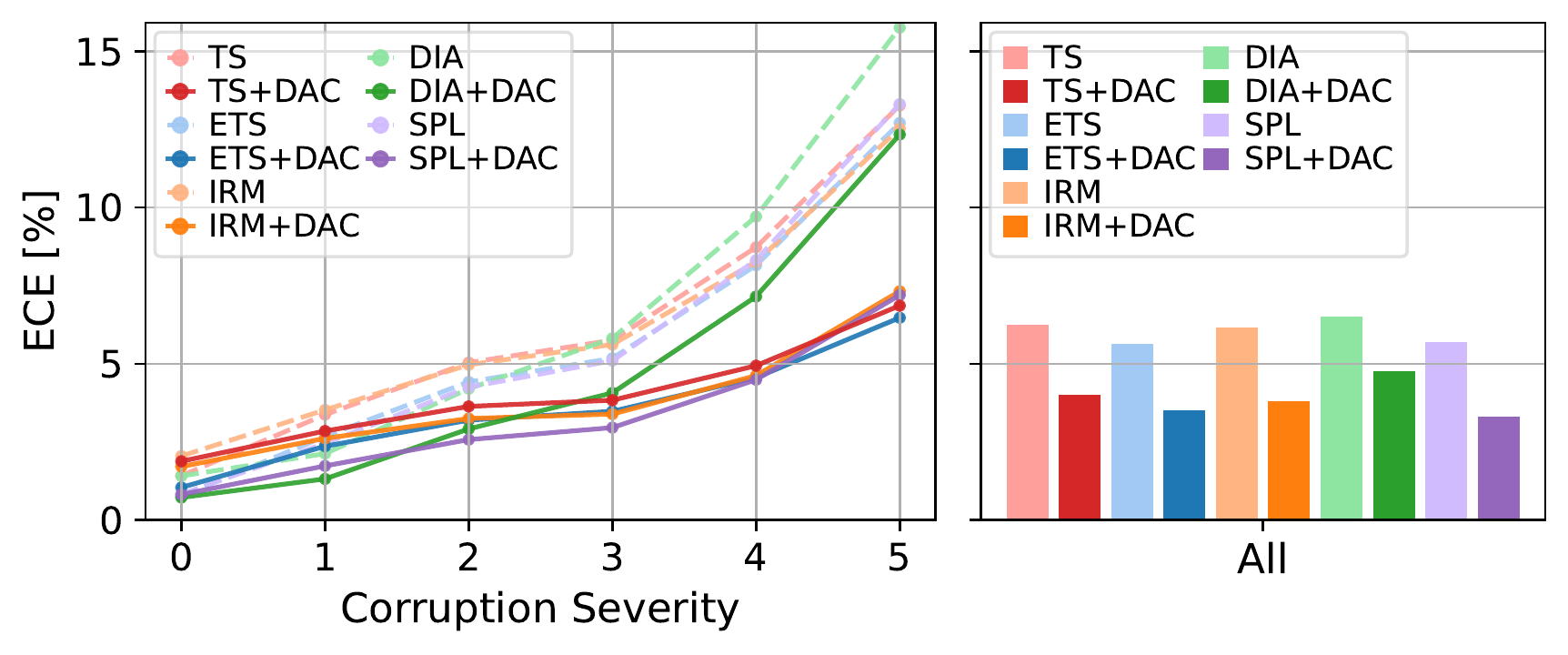} \\  
    {ImageNet-DenseNet169} & {ImageNet-BiT-M}\\  

    \includegraphics[width=0.47\textwidth]{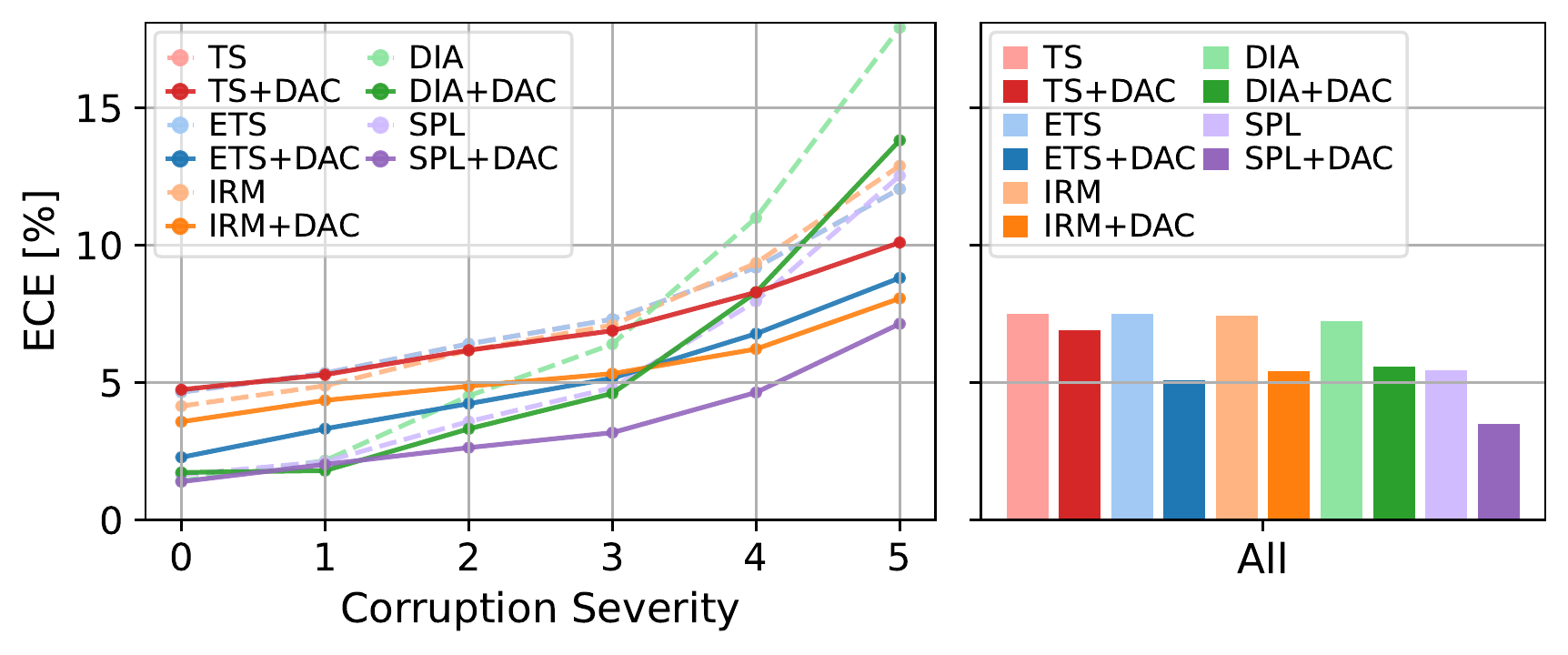} &
    \includegraphics[width=0.47\textwidth]{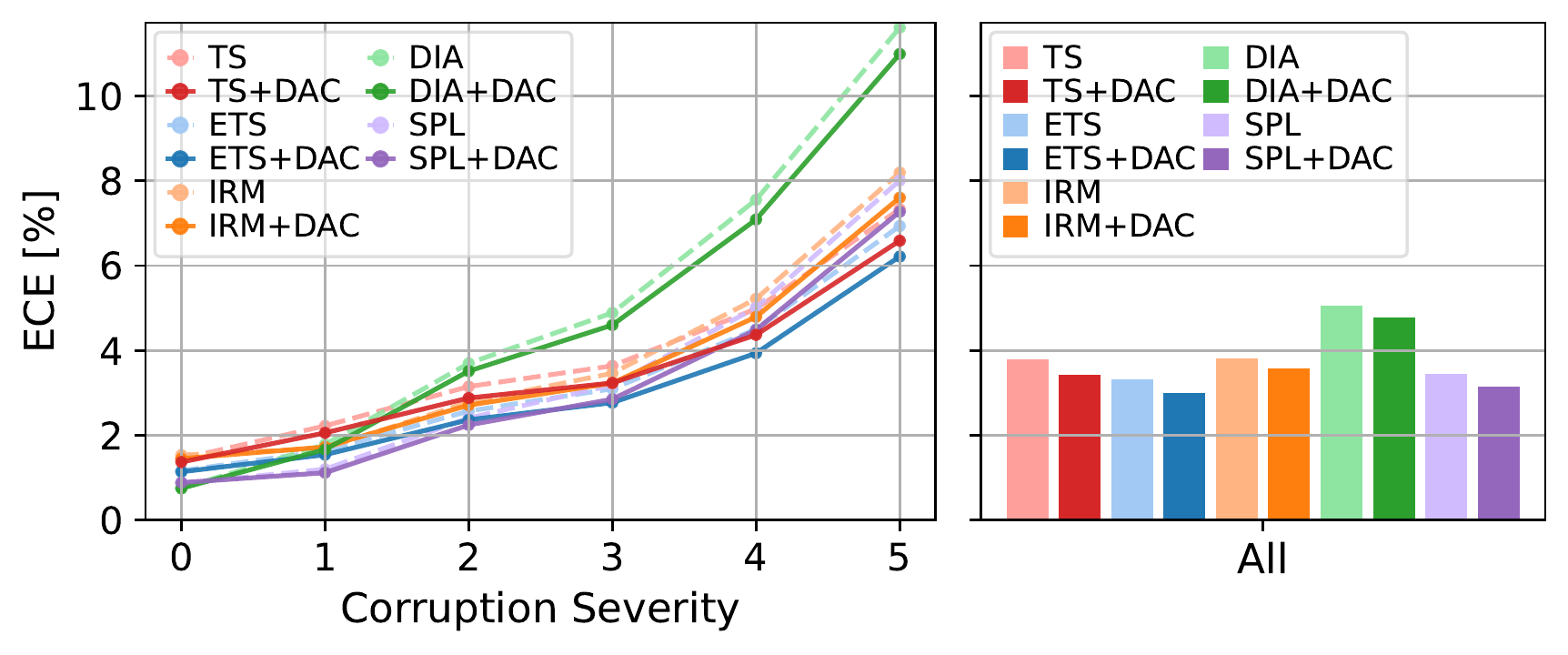} \\  
    {ImageNet-ResNeXt-WSL} & {ImageNet-ViT-B}\\  
    \end{tabular}
	\caption{Expected calibration error ($\times 10^2$) of post-hoc methods with and without our method DAC for different model and dataset combinations. \textbf{Line plots}: Macro-averaged ECE across all corruption types shown for each corruption severity, from in-domain to severity=5 (heavily corrupted). \textbf{Bar plots}: Macro-averaged ECE across all corruption types as well as across all severities. Our model captures domain-shift scenarios reliably and thus increases calibration (=decreases ECE) across the whole spectrum of corruptions.}
\label{fig:ece_line_bar_combined_assessment}
\end{figure*}
\setlength{\tabcolsep}{6pt} 
\renewcommand{\arraystretch}{1}

\setlength{\tabcolsep}{0.5pt}
\renewcommand{\arraystretch}{0.5}
\begin{table*}[h]
\centering
\vskip 0.15in
\caption{Difference in expected calibration error ($\times 10^2$) of post-hoc calibration methods with and without our method DAC. \textbf{$InD$}: In-domain, \textbf{$SEV. 5$}: Heavily corrupted (severity of 5) and \textbf{$ALL$}: Macro-averaged ECE across all corruptions from severity=0 (in-domain) until severity=5 (heavily corrupted). For \textbf{$ALL$}, we additionally report the ratio to indicate overall performance gain. 
Post-hoc calibration methods combined with DAC consistently improve calibration in cases of heavy corruption as well as overall calibration (relative improvement of around 5-40\%), while preserving in-domain performance (negative deltas are better).}
\vskip 0.15in
\label{tab:delta_ece}
    \begin{small}
    \begin{sc}
    \begin{tabular}{crrr}
    \toprule
     & \multicolumn{1}{c}{InD} & \multicolumn{1}{c}{Sev. 5} & \multicolumn{1}{c}{All} \\ 
    \midrule
    TS & \minusbar{-0.42}{-0.0725}& \minusbar{-1.14}{-0.1959}& \minusbar{-0.46}{-0.0789}10\% \\
    ETS & \minusbar{-0.49}{-0.0845}& \minusbar{-1.25}{-0.2144}& \minusbar{-0.56}{-0.0961}13\% \\
    IRM & \minusbar{-0.36}{-0.0620}& \minusbar{-1.67}{-0.2879}& \minusbar{-0.92}{-0.1590}18\% \\
    DIA & \minusbar{-1.09}{-0.1872}& \minusbar{-2.91}{-0.5000}& \minusbar{-1.67}{-0.2872}29\% \\
    SPL & \minusbar{-1.02}{-0.1758}& \minusbar{-2.00}{-0.3446}& \minusbar{-1.45}{-0.2489}26\% \\
    \bottomrule
    \multicolumn{4}{c}{a.) CIFAR10 ResNet18}
    \end{tabular}
    \end{sc}
    \end{small}
\medskip
    \begin{small}
    \begin{sc}
    \begin{tabular}{crrr}
    \toprule
    & \multicolumn{1}{c}{InD} & \multicolumn{1}{c}{Sev. 5} & \multicolumn{1}{c}{All} \\ 
    \midrule
    TS & \minusbar{-0.22}{-0.0134}& \minusbar{-8.05}{-0.4995}& \minusbar{-4.29}{-0.2660}42\% \\
    ETS & \minusbar{-0.22}{-0.0137}& \minusbar{-8.06}{-0.5000}& \minusbar{-4.29}{-0.2664}42\% \\
    IRM & \minusbar{-0.85}{-0.0526}& \minusbar{-7.15}{-0.4439}& \minusbar{-4.46}{-0.2764}38\% \\
    DIA & \minusbar{-1.87}{-0.1160}& \minusbar{-6.20}{-0.3848}& \minusbar{-3.97}{-0.2461}30\% \\
    SPL & \minusbar{-0.27}{-0.0165}& \minusbar{-4.77}{-0.2962}& \minusbar{-2.83}{-0.1756}30\% \\
    \bottomrule
    \multicolumn{4}{c}{b.) CIFAR100 VGG16}
    \end{tabular}
    \end{sc}
    \end{small}
\medskip
    \begin{small}
    \begin{sc}
    \begin{tabular}{crrr}
    \toprule
    & \multicolumn{1}{c}{InD} & \multicolumn{1}{c}{Sev. 5} & \multicolumn{1}{c}{All} \\ 
    \midrule
    TS & \plusbar{0.08}{0.0088}& \minusbar{-3.16}{-0.3427}& \minusbar{-1.49}{-0.1614}26\% \\
    ETS & \plusbar{0.03}{0.0031}& \minusbar{-4.61}{-0.5000}& \minusbar{-2.08}{-0.2257}38\% \\
    IRM & \minusbar{-0.48}{-0.0521}& \minusbar{-4.25}{-0.4607}& \minusbar{-2.47}{-0.2679}38\% \\
    DIA & \minusbar{-0.09}{-0.0098}& \minusbar{-3.49}{-0.3781}& \minusbar{-1.81}{-0.1965}25\% \\
    SPL & \minusbar{-0.15}{-0.0168}& \minusbar{-4.23}{-0.4590}& \minusbar{-2.15}{-0.2332}35\% \\
    \bottomrule
    \multicolumn{4}{c}{c.) ImageNet DenseNet169}
    \end{tabular}
    \end{sc}
    \end{small}

    \begin{small}
    \begin{sc}
    \begin{tabular}{crrr}
    \toprule
    & \multicolumn{1}{c}{InD} & \multicolumn{1}{c}{Sev. 5} & \multicolumn{1}{c}{All} \\ 
    \midrule
    TS & \plusbar{0.47}{0.0363}& \minusbar{-6.40}{-0.5000}& \minusbar{-2.26}{-0.1766}36\% \\
    ETS & \plusbar{0.28}{0.0220}& \minusbar{-6.23}{-0.4865}& \minusbar{-2.13}{-0.1663}38\% \\
    IRM & \minusbar{-0.34}{-0.0263}& \minusbar{-5.20}{-0.4064}& \minusbar{-2.34}{-0.1827}38\% \\
    DIA & \minusbar{-0.69}{-0.0537}& \minusbar{-3.42}{-0.2668}& \minusbar{-1.75}{-0.1367}27\% \\
    SPL & \minusbar{-0.01}{-0.0010}& \minusbar{-6.11}{-0.4770}& \minusbar{-2.41}{-0.1885}42\% \\
    \bottomrule
    \multicolumn{4}{c}{d.) ImageNet BiT-M}
    \end{tabular}
    \end{sc}
    \end{small}
\medskip
    \begin{small}
    \begin{sc}
    \begin{tabular}{crrr}
    \toprule
    & \multicolumn{1}{c}{InD} & \multicolumn{1}{c}{Sev. 5} & \multicolumn{1}{c}{All} \\ 
    \midrule
    TS & \plusbar{0.09}{0.0083}& \minusbar{-1.95}{-0.1809}& \minusbar{-0.58}{-0.0537}\;\;8\% \\
    ETS & \minusbar{-2.37}{-0.2201}& \minusbar{-3.24}{-0.3004}& \minusbar{-2.39}{-0.2221}32\% \\
    IRM & \minusbar{-0.57}{-0.0530}& \minusbar{-4.82}{-0.4477}& \minusbar{-2.02}{-0.1875}27\% \\
    DIA & \plusbar{0.30}{0.0279}& \minusbar{-4.10}{-0.3805}& \minusbar{-1.64}{-0.1526}23\% \\
    SPL & \minusbar{-0.25}{-0.0231}& \minusbar{-5.39}{-0.5000}& \minusbar{-1.94}{-0.1801}36\% \\
    \bottomrule
    \multicolumn{4}{c}{e.) ImageNet ResNeXt-WSL}
    \end{tabular}
    \end{sc}
    \end{small}
\medskip
    \begin{small}
    \begin{sc}
    \begin{tabular}{crrr}
    \toprule
    & \multicolumn{1}{c}{InD} & \multicolumn{1}{c}{Sev. 5} & \multicolumn{1}{c}{All} \\ 
    \midrule
    TS & \minusbar{-0.08}{-0.0533}& \minusbar{-0.74}{-0.5000}& \minusbar{-0.38}{-0.2546}10\% \\
    ETS & \minusbar{-0.03}{-0.0175}& \minusbar{-0.72}{-0.4829}& \minusbar{-0.32}{-0.2166}10\% \\
    IRM & \minusbar{-0.08}{-0.0567}& \minusbar{-0.59}{-0.3976}& \minusbar{-0.23}{-0.1541}\;\;6\% \\
    DIA & \minusbar{-0.00}{-0.0004}& \minusbar{-0.62}{-0.4171}& \minusbar{-0.28}{-0.1880}\;\;6\% \\
    SPL & \plusbar{0.02}{0.0119}& \minusbar{-0.74}{-0.4958}& \minusbar{-0.30}{-0.2045}\;\;9\% \\
    \bottomrule
    \multicolumn{4}{c}{f.) ImageNet ViT-B}
    \end{tabular}
    \end{sc}
    \end{small}
\end{table*}
\setlength{\tabcolsep}{6pt} 
\renewcommand{\arraystretch}{1} 

\section{Results}

First, we show that combining our method DAC with existing post-hoc calibration methods increases calibration performance across the entire spectrum, from in-domain to heavily corrupted data distributions, for different datasets and various model architectures, including transformers. 
Secondly, we show the calibration performance of our method on purely OOD scenarios. Lastly, we conduct additional experiments, such as a layer importance analysis and a data efficiency analysis.

\subsection{DAC Boosts Calibration Performance Beyond In-Domain Scenarios}

We begin by systematically assessing whether the performance of state-of-the-art post-hoc calibration methods can be improved when extended by our proposed DAC method. In particular, we are interested in scenarios under domain shift. To this end, we show calibration performance on CIFAR-C and ImageNet-C for severity levels from 1 to 5 and additionally provide results for in-domain scenarios (severity=0). 
Fig.~\ref{fig:ece_line_bar_combined_assessment} illustrates a comparison between stand-alone post-hoc methods and combined methods with DAC for various classifiers. The line charts reveal that DAC consistently improves the calibration performance of post-hoc methods in domain-shift cases. Tab.~\ref{tab:delta_ece} underpins this performance increase even further by revealing a substantial decrease in the absolute ECE for heavily corrupted data (severity=5) scenarios when DAC is used. When optimizing for domain-shift performance, a sharp decline in in-domain performance is generally observed for other existing methods \cite{tomani2021post}. This is, however not the case for our method, where we even observe slight improvements of in-domain ECE for many classifier and post-hoc method configurations, and for the few cases where in-domain ECE marginally increases (in the order of $10^{-4}$ to $10^{-3}$), the ECE in heavily corrupted scenarios decreases in the order of far more than a magnitude in comparison. Finally, in order to measure the overall improvement of DAC across the entire spectrum of corruption from severity 0 to 5, we calculate the macro-averaged ECE across all corruption types and levels of severity for each method. We discover a consistent improvement for all post-hoc methods combined with DAC as opposed to stand-alone methods visualized in the bar charts in Fig.~\ref{fig:ece_line_bar_combined_assessment} and in Tab.~\ref{tab:delta_ece}. Moreover, Tab.~\ref{tab:mean_ece} further reveals that DAC consistently boosts calibration performance across diverse architectures and datasets. In Tab.~\ref{tab:mean_brier} we demonstrate that improved calibration performance is also consistent with better Brier scores.

\begin{table*}[!h]
\footnotesize
\vskip 0.15in
\caption{Mean \textbf{Brier-score} computed across all corruptions from severity=0 (in-domain) until severity=5 (heavily corrupted). Note that since SPL only calibrates the highest predicted confidence, it is not directly possible to evaluate Brier scores.}
\label{tab:mean_brier}
\begin{center}
\begin{small}
\begin{sc}
\begin{tabular}{l|c|cccc|cccc}
    \toprule
    & \multicolumn{1}{|c|}{\bfseries Uncal} &
    \multicolumn{4}{c|}{\bfseries Baseline Calibration Methods} &
    \multicolumn{4}{c}{\bfseries Combination with DAC (Ours)} \\
    & - & TS & ETS & IRM & DIA  & TS & ETS & IRM & DIA \\
    \midrule
    C10 ResNet18 & 0.4547 & 0.3865 & 0.3865 & 0.3913 & 0.3906 & \underline{0.3842} & \textbf{0.3841} & 0.3852 & \underline{0.3842} \\
    C10 VGG16 & 0.4294 & 0.3562 & 0.3564 & 0.3633 & 0.3717 & \textbf{0.3547} & \textbf{0.3547} & \underline{0.3559} & 0.3565 \\
    C10 DenseNet121 & 0.4593 & 0.3940 & 0.3940 & 0.3978 & 0.3986 & \textbf{0.3896} & \textbf{0.3896} & \underline{0.3900} & 0.3926 \\ \hline
    C100 ResNet18 & 0.6902 & 0.6655 & 0.6638 & 0.6696 & 0.6568 & 0.6655 & \underline{0.6562} & 0.6584 & \textbf{0.6525} \\
    C100 VGG16 & 0.8205 & 0.6784 & 0.6784 & 0.6837 & 0.6873 & \textbf{0.6554} & \textbf{0.6554} & \underline{0.6578} & 0.6652 \\
    C100 DenseNet121 & 0.6963 & \underline{0.6261} & 0.6265 & 0.6370 & 0.6312 & \underline{0.6261} & \textbf{0.6254} & 0.6275 & 0.6460 \\ \hline
    IMG ResNet152 & 0.6510 & 0.6360 & 0.6354 & 0.6432 & 0.6432 & \textbf{0.6331} & \textbf{0.6331} & \underline{0.6332} & \textbf{0.6331} \\
    IMG DenseNet169 & 0.6707 & 0.6505 & 0.6507 & 0.6582 & 0.6562 & \underline{ 0.6457} & \textbf{0.6443} & 0.6460 & 0.6494 \\
    IMG Xception & 0.8744 & 0.7539 & 0.7534 & 0.7812 & 0.7547 & 0.7539 & \textbf{0.7519} & 0.7533 & \underline{0.7521} \\ \hline
    IMG BiT-M & 0.6286 & 0.6173 & 0.6164 & 0.6184 & 0.6134 & 0.6067 & \underline{0.6064} & \textbf{0.6057} & \underline{0.6064} \\
    IMG ResNeXt-WSL & 0.5393 & 0.5117 & 0.5117 & 0.5143 & 0.5144 & 0.5090 & 0.5074 & \textbf{0.5023} & \underline{0.5068} \\
    IMG ViT-B & \underline{0.5809} & 0.5816 & 0.5812 & 0.5815 & 0.5869 & \underline{0.5809} & \textbf{0.5806} & 0.5813 & 0.5862 \\ 
\bottomrule
\end{tabular}
\end{sc}
\end{small}
\end{center}
\vskip -0.1in
\end{table*}

Since research regarding state-of-the-art post-hoc methods applied to recent large-scale neural networks is still lacking, we want to attempt to bridge this gap and show results for modern ResNet as well as transformer architectures trained on large corpora of data. We make the same observation as Minderer et al. \cite{minderer2021revisiting} that, in fact, even modern ResNet architectures are not well calibrated in domain-shift settings despite being pre-trained on huge amounts of data, yet transformer architectures perform particularly well. In our experiments, we observe that modern ResNet architectures (BiT-M and ResNeXt) can indeed be further calibrated with post-hoc methods. Especially, SPL+DAC performs best and reduces the ECE by around 37\% for ResNeXt-WSL and 42\% for BiT-M compared to the best-performing standard post-hoc method (Tab.~\ref{tab:mean_ece}). ViT-B, on the other hand, can profit from existing post-hoc methods too, at least in-domain. For domain-shift scenarios, ViT outperforms all standard post-hoc methods, except when combined with DAC, in that case, ETS+DAC outperforms existing methods by 12\%.

\subsection{Calibration in OOD Scenarios}

To complement the previous distributional shift experiments, we conduct additional experiments for the out-of-domain case, incorporating data samples with completely different classes w.r.t.\ the training data. Ideally, a well-calibrated uncertainty-aware model would produce high-confidence predictions for in-domain data and low-confidence ones for the OOD case, allowing for the detection of OOD data samples. Based on this idea, several metrics have been proposed in the OOD literature \cite{hendrycks2017oodbaseline, liang2018odin, lee2018mahalanobis} to quantify the model performance in OOD scenarios, including FPR at 95\% TPR, detection error, AUROC, AUPR-In/AUPR-Out, which we employ for our experiments.

To examine DAC in the OOD scenario, we set up the OOD dataset with in-domain data from ImageNet-1k and OOD data from ObjectNet-OOD (c.f.\ Section~\ref{sec:exp_setup} for dataset descriptions). Top-class confidence predictions are produced by models trained and calibrated on ImageNet-1k with various calibration methods with and without the proposed DAC method. The OOD metrics are computed from the confidence predictions. We summarize the results for the DenseNet169 backbone in Tab.~\ref{tab:ood_metrics_densenet}. Additional results for other backbones can be found in Appendix~\ref{asec:ood_results}.

\begin{table}[ht]
\vskip 0.15in
\centering
\caption{OOD performance with DenseNet169 trained on ImageNet-1k and using ImageNet-1k/ObjectNet-OOD as in-domain/OOD test sets, respectively. We observe that DAC consistently improves all the OOD metrics for all baseline methods.}
\label{tab:ood_metrics_densenet}
    \begin{small}\begin{sc}
        \begin{tabular}{lccccc}
        \toprule
        {} &  FPR &  Det. &    AU- &  AUPR- &  AUPR- \\
        {} &  @95\% ↓ &  Err ↓ &    ROC ↑ &  In ↑ &  Out ↑ \\
        \midrule
        TS & 21.47 & 22.86 & 84.92 & 81.14 & 86.87 \\
         +DAC & \textbf{20.73} & \textbf{22.30} & \textbf{85.50} & \textbf{81.86} & \textbf{87.46} \\ \hline
        ETS & 21.92 & 23.15 & 84.72 & 79.82 & 86.72 \\
         +DAC & \textbf{20.90} & \textbf{22.29} & \textbf{85.51} & \textbf{82.16} & \textbf{87.47} \\ \hline
        IRM & 21.51 & 22.31 & 83.62 & 81.22 & 86.10 \\
         +DAC & \textbf{4.87} & \textbf{20.72} & \textbf{85.81} & \textbf{82.73} & \textbf{87.90} \\ \hline
        DIA & 22.09 & 24.26 & 83.67 & 79.86 & 85.77 \\
         +DAC & \textbf{21.36} & \textbf{23.64} & \textbf{84.35} & \textbf{80.63} & \textbf{86.47} \\ \hline
        SPL & 21.66 & 24.38 & 83.65 & 79.95 & 85.78 \\
         +DAC & \textbf{20.89} & \textbf{22.30} & \textbf{85.54} & \textbf{82.05} & \textbf{87.56} \\
        \bottomrule
        \end{tabular}
    \end{sc}\end{small}
\vskip -0.1in
\end{table}

In general, we see that DAC yields more robust calibration in the OOD scenario, as demonstrated by its consistent improvement of OOD results.

\subsection{Layer Importance for Calibration}
\label{subsec:layer_importance}

\begin{figure}[h]
\vskip 0.2in
\begin{center}
\centerline{\includegraphics[width=1.0\columnwidth]{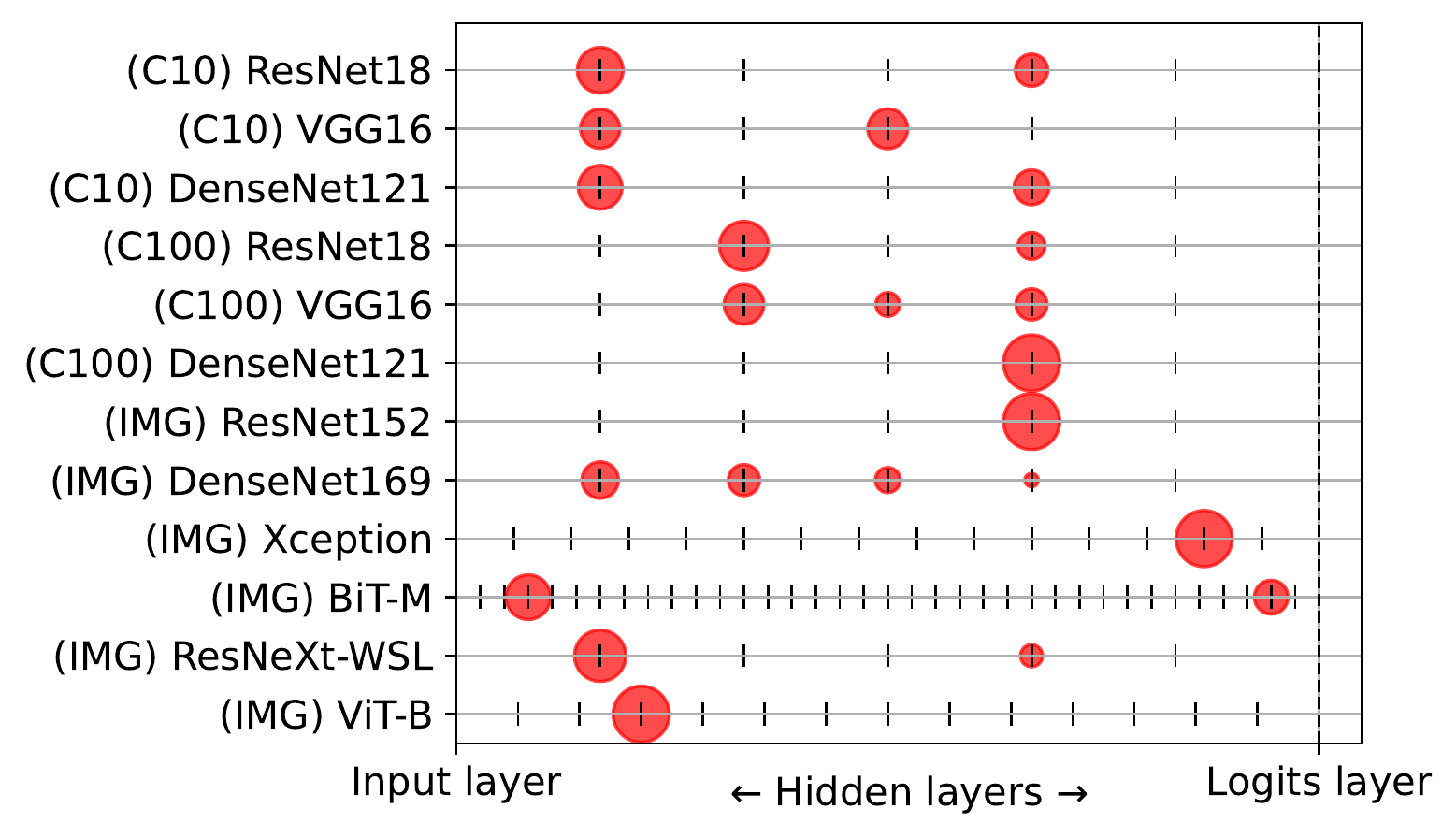}}
\caption{Importance of classifier layers found by DAC. The size of blobs indicates the magnitude of assigned weights for each layer after training of DAC from left (input) to right (logits).
}
\label{fig:layer_weights}
\end{center}
\vskip -0.2in
\end{figure}

Next, we want to investigate which layers of each classifier carry valuable information for DAC to yield calibrated predictions. For each layer, DAC learns a weight based on the importance of the respective layer (equation~\eqref{eq:weighting_scheme}). 
In Fig.~\ref{fig:layer_weights} we demonstrate that DAC focuses on a few important layers, yet the logits layer is never one of them. This is particularly interesting because current state-of-the-art post-hoc calibration methods only focus on the logits vector for recalibration without even considering hidden layers of the classifier. Hence, we can conclude that one reason for DAC's performance improvement can be attributed to its ability to take information from other layers into account apart from the logits layer. 

\begin{figure}[h]
\vskip 0.2in
\begin{center}
\centerline{\includegraphics[width=0.95\columnwidth]{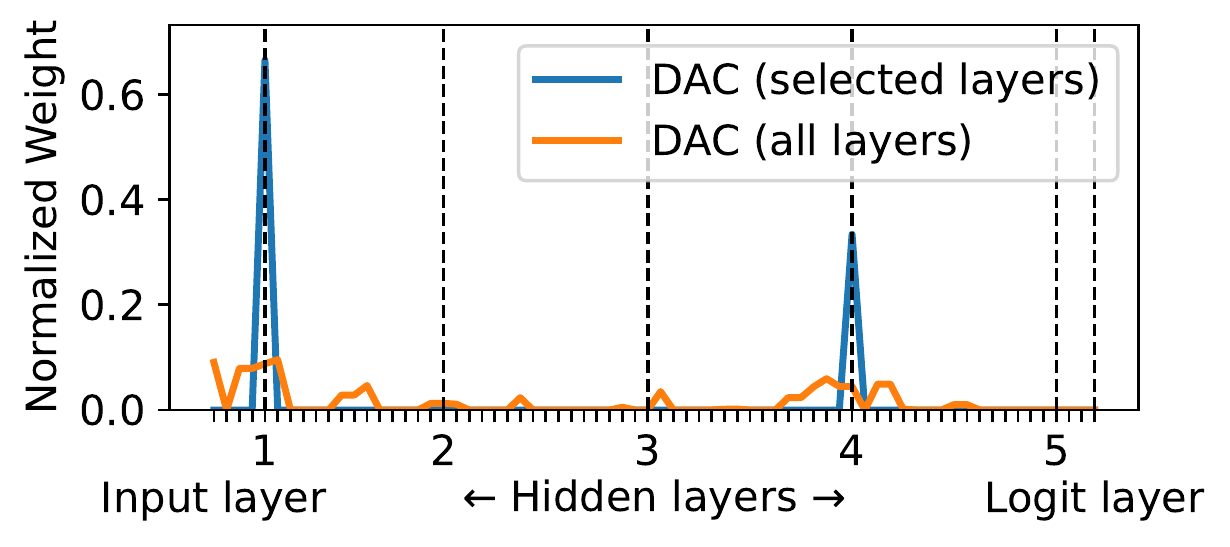}}
\caption{Comparison between our DAC with ``selected layers" to a holistic DAC, which utilizes "all layers" present in ResNet18 trained on CIFAR10. DAC is able to capture the most relevant areas in layer space with important information for calibration.}
\label{fig:resnet18_weight_lineplot}
\end{center}
\vskip -0.2in
\end{figure}

Even though the layers DAC has access to are well distributed throughout the architecture of the classifier, we want to investigate whether DAC can capture all the necessary information present in all the layers of the classifier. To this end, we compare our simple and fast DAC method, which uses a subset of the layers, to a holistic DAC, which utilizes all layers. In Fig.~\ref{fig:resnet18_weight_lineplot}, we illustrate the weights DAC assigns to every layer, which are normalized to add up to 1. The holistic DAC is able to attend to various layers; however, we observe that this does not necessarily result in better calibration performance, which can be attributed to overfitting (see Appendix~\ref{appendix:ablation_layer_choice} for further insights).

\subsection{Sensitivity Analysis of KNN}

\begin{figure}[H]
\vskip 0.2in
\begin{center}
\centerline{\includegraphics[width=1.0\columnwidth]{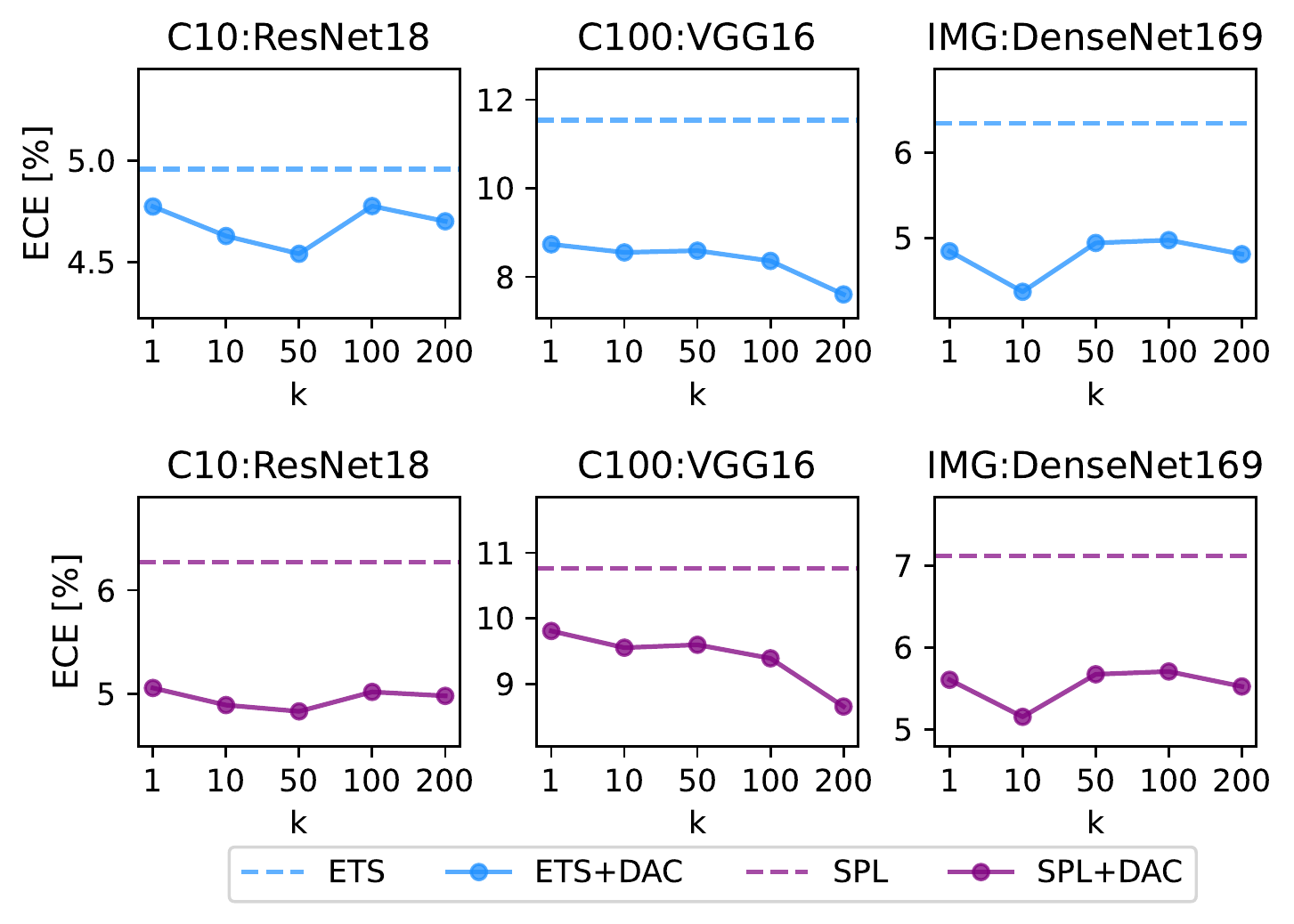}}
\caption{The sensitivity of DAC to the hyperparameter $k$ used for KNN operations in each layer. We combined DAC with ETS (1st row) and SPL (2nd row), varying hyperparameter $k$.}
\label{fig:k_sensitivity}
\end{center}
\vskip -0.2in
\end{figure}

In this section, we evaluate the sensitivity of DAC to the hyperparameter $k$ used for KNN operations in each layer. To this end, we study the resulting calibration performance while varying $k$. Figure~\ref{fig:k_sensitivity} illustrates the impact of $k$ on the calibration performance of DAC when combined with ETS and SPL, with DAC having different values of $k \in$ \{1, 10, 50, 100, 200\}. For each model, the macro-averaged ECE (×$10^2$) is computed across all corruptions from severity=0 (in-domain) until severity=5 (heavily corrupted). We show that DAC consistently boosts the performance of ETS and SPL regardless of the choice of $k$, indicating that the performance of DAC is not overly sensitive to the value of $k$. On the other hand, we observe that using the proper choice for $k$, suggested by \cite{sun2022out}, indeed results in the best calibration performance also for our method, implying that putting in additional effort for hyperparameter tuning can further enhance the performance of DAC.

\subsection{Data Efficiency of DAC}

\begin{figure}[ht!]
\vskip 0.2in
\begin{center}
\centerline{\includegraphics[width=1.0\columnwidth]{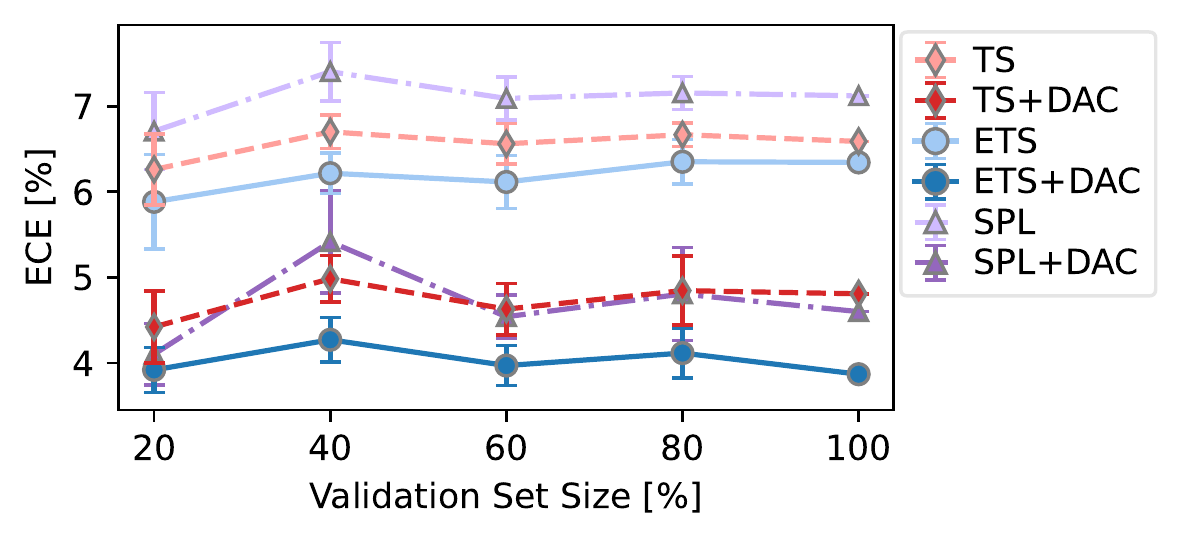}}
\caption{DAC is robust across different validation set sizes (10\% - 100\%) for DenseNet169 trained on ImageNet (ECE ($\times 10^2$)). We conducted five experiments with randomly sampled validation sets.}
\label{fig:data_efficiency_lineplot}
\end{center}
\vskip -0.2in
\end{figure}

Lastly, we investigate how sensitive our method is to various validation set sizes. We focus on the best-performing methods, namely ETS+DAC and SPL+DAC, and compare them to the respective stand-alone methods, ETS and SPL. Additionally, we incorporate TS in our study since this method is least likely to suffer from overfitting because it comprises only one parameter to train. We encounter in Fig.~\ref{fig:data_efficiency_lineplot} that no matter which validation set size we use, combinations with DAC perform better than without DAC. Additionally, DAC does not overfit the data for small validation set sizes. 

\section{Conclusion}
In this work, we have introduced an accuracy-preserving density-aware calibration method that can readily be applied with SOTA post-hoc methods in order to boost domain-shift and OOD calibration performance. We found that our proposed method DAC combined with existing post-hoc calibration methods yields robust predictive uncertainty estimates for any level of domain shift, from in-domain to truly OOD scenarios. In particular, ETS+DAC, as well as SPL+DAC, performed the best. We further demonstrated that hidden layers in classifiers carry valuable information for accurately predicting uncertainty estimates. Lastly, we show that even recently developed large-scale models pre-trained on vast amounts of data can be calibrated effectively by DAC, opening up new research directions within the field of post-hoc calibration for entirely new applications. 
One of the limitations of our method can arise when applying it to highly parametric classifiers with numerous layers, as well as when determining which layers possess the most calibration-related information for DAC.
The amount of calibration-related information within specific layers of a classifier seems to depend not only on the model architecture but also on the relationship between model size and dataset characteristics. 
We hope our findings will encourage further research into developing post-hoc methods that take into account features from the underlying neural network classifier, rather than just the output features.

\bibliography{references}
\bibliographystyle{icml2023}

\newpage
\appendix
\onecolumn

\section{Datasets}

We follow the standard setup for evaluating calibration performance \cite{guo_calibration_2017}.
We split each dataset into train, validation, and test set. 
The train set is used to train classifier neural networks. 
The validation set is used to optimize the post-hoc calibration methods to recalibrate classifiers. 
Then, the calibration performances are evaluated on the test set.

In Tab.~\ref{tab:dataset_description}, we show the numbers of image-label pairs in each split we use. 
For CIFAR10 and CIFAR100 \cite{krizhevsky2009learning_cifar}, following Guo et al.~\yrcite{guo_calibration_2017}, we split the original train set, which contains 50,000 image-label pairs, into 45,000 image-label pairs of train set and 5,000 image-label pairs of the validation set.
For ImageNet \cite{deng2009imagenet}
we split the original validation set, which contains 50,000 image-label pairs, into 12,500 image-label pairs of the validation set and 37,500 image-label pairs of the test set. As a common practice, we split the official validation set from ImageNet into a validation set for training our post-hoc method and test set \cite{minderer2021revisiting}.

\begin{table}[h!]
\vskip 0.15in
\caption{The numbers of image-label pairs we used for each dataset.}
\label{tab:dataset_description}
\begin{center}
\begin{small}
\begin{sc}
\begin{tabular}{cccc}
\toprule
Dataset  & Train     & Val    & Test   \\
\midrule
CIFAR10  & 45,000    & 5,000  & 10,000 \\
CIFAR100 & 45,000    & 5,000  & 10,000 \\
ImageNet & 1,281,167 & 12,500 & 37,500 \\
\bottomrule
\end{tabular}
\end{sc}
\end{small}
\end{center}
\vskip -0.1in
\end{table}

To quantify calibration performance for domain-shift scenarios, we use ImageNet-C as well as CIFAR-C \cite{hendrycks2019benchmarking_imagenet_c}. 
Both datasets have 18 distinct corruption types, each having 5 different levels of severity, mimicking a scenario where the input data to a classifier gradually shifts away from the training distribution.
The 18 corruptions include 4 noise corruptions (gaussian noise, shot noise, speckle noise, and impulse noise), 4 blur corruptions (defocus blur, gaussian blur, motion blur, and zoom blur), 5 weather corruptions (snow, fog, brightness, spatter, and frost), and 5 digital corruptions (elastic transform, pixelate, JPEG compression, contrast, and saturate).

\section{Classifiers}

In this section we describe the implementation of the classifiers we used in our work.
\begin{itemize}
    \item CIFAR10: 
        \begin{itemize}
            \item ResNet18 \cite{he2016deep_resnet}/VGG16 \cite{simonyan2014very_vgg}/DenseNet121 \cite{huang2017densely_densenet}: We use PyTorch's official implementation to obtain the model architectures. We trained all models for 200 epochs: 100 epochs at a learning rate of 0.01, 50 epochs at a learning rate of 0.005, 30 epochs at a learning rate of 0.001, and 20 epochs at a learning rate of 0.0001. We use a basic data augmentation technique of random cropping and horizontal flipping.
        \end{itemize}
   \item CIFAR100:
       \begin{itemize}
            \item ResNet18 \cite{he2016deep_resnet}/VGG16 \cite{simonyan2014very_vgg}/DenseNet121 \cite{huang2017densely_densenet}: We obtain architectures from a github repository\footnote{https://github.com/weiaicunzai/pytorch-cifar100} that provides PyTorch implementation of the optimal architectures for CIFAR100 dataset. We trained all models for 200 epochs, with an initial learning rate of 0.01, which decays 0.2 times at the 60, 120, and 160th epochs.  We use a basic data augmentation technique of random cropping and horizontal flipping.
        \end{itemize}
  \item ImageNet-1k: We use pre-trained models.
    \begin{itemize}
        \item ResNet152 \cite{he2016deep_resnet}/DenseNet169 \cite{huang2017densely_densenet}: We use the ImageNet-1k pre-trained models from the torchvision library.
        \item Xception \cite{chollet2017xception}: We use the ImageNet-1k pre-trained model provided by the timm library\footnote{\label{footnote:timm}https://github.com/rwightman/pytorch-image-models}.
        \item BiT-M \cite{kolesnikov2020big}: We use the ImageNet-21k pre-trained and ImageNet-1k fine-tuned model provided by the timm library\footref{footnote:timm}. Specifically, we use the BiT-M based on ResNetV2 101x1 architecture.
        \item ResNeXt-WSL \cite{wslimageseccv2018}: We use the Instagram pre-trained and ImageNet-1k fine-tuned model provided by Meta's research group\footnote{https://github.com/facebookresearch/WSL-Images}. Specifically, we use a model with ResNeXt101 32x8d architecture.
        \item ViT-Base \cite{dosovitskiy2020image_vit}: We use the ImageNet-21k pre-trained and ImageNet-1k fine-tuned model provided by the timm library\footref{footnote:timm}. Specifically, we use the ViT-Base that expects an input image size of 224 and have 16 patch-embeddings.
    \end{itemize}
\end{itemize}

\section{Post-hoc methods}

\subsection{Density-Aware Calibration (DAC)}
\label{appendix:DAC}
\begin{table}[h]
\vskip 0.15in
\caption{Layers from classifiers used for DAC. For neural networks that have block structures, we pick the very last layer of each block (denoted by \textsc{Block}-i for $i_{th}$ block), and, additionally, the immediate layer before the first block (denoted by \textsc{Pre-block}), the layer just before the fully-connected layers at the end of neural networks (denoted by \textsc{Penultimate}), and the logits layer (denoted by \textsc{Logits}). For VGG, we use feature vectors after every max-pooling layer where the resolution of the feature map changes (denoted by \textsc{Maxpool}-i).}
\label{tab:layers}
\begin{center}
\begin{small}
\begin{sc}
\begin{tabular}{c|l|l}
\toprule
\multicolumn{1}{c|}{Dataset} & \multicolumn{1}{c|}{Model} & \multicolumn{1}{c}{Used layers} \\
\midrule
 & ResNet18 & Pre-block, Block-1, ..., Block-4, Logits \\ \cline{2-3} 
CIFAR10 & VGG16 & Maxpool-1, ..., Maxpool-5, logits \\ \cline{2-3} 
 & DenseNet121 & Pre-block, Block-1, ..., Block-3, Penultimate, logits \\ 
 \hline
 & ResNet18 & Pre-block, Block-1, ..., Block-4, Logits  \\ \cline{2-3} 
 CIFAR100 & VGG16 & Maxpool-1, ..., Maxpool-5, logits \\ \cline{2-3} 
 & DenseNet121 & Pre-block, Block-1, ..., Block-3, Penultimate, logits \\
 \hline
& ResNet152 & Pre-block, Block-1, ..., Block-4, Logits  \\ \cline{2-3} 
 & DenseNet169 & Pre-block, Block-1, ..., Block-3, Penultimate, logits \\ \cline{2-3} 
 & Xception & Pre-block, Block-1, ..., Block-12, Penultimate, logits \\ \cline{2-3} 
 ImageNet & \begin{tabular}[c]{@{}l@{}}BiT-M \\ (ResNetV2)\end{tabular} & \begin{tabular}[c]{@{}l@{}}Pre-block, Block-1, ..., Block-34, Penultimate, logits\end{tabular} \\ \cline{2-3} 
 & \begin{tabular}[c]{@{}l@{}}ResNeXt-WSL \\ (ResNeXt101 32x8d)\end{tabular} & Pre-block, Block-1, ..., Block-4, logits \\ \cline{2-3} 
 & ViT-Base & Pre-block, Block-1, ..., Block-24, Penultimate, logits \\
 \bottomrule
\end{tabular}
\end{sc}
\end{small}
\end{center}
\vskip -0.1in
\end{table}

Even though DAC is capable of using every layer in a classifier due to its weighting scheme, we opt for a much simpler and faster version that uses a subset of layers.
We follow a structured approach for choosing layers to end up with a well-distributed subset of layers: We choose (1) the last layer of each ``block" in a neural network, e.g., ResNet or Transformer block, or, (2) the layer where the resolution or channel size of the feature vector changes, e.g., VGG.
The intuition behind this approach is based on the fact that layers in neural networks represent an image at increasing levels of abstraction from low-level to high-level representation, and thus, different blocks or layers with different resolutions are expected to have different levels of representation.
The approach of choosing a subset of layers to obtain different levels of representations has been applied to a wide range of works in the computer vision field  \cite{zhang2018perceptual, gatys2016image, chen2017photographic}: Zhang et al.~\yrcite{zhang2018perceptual} leverage different levels of representations for better measurement of the perceptual difference between two images; Gatys et al.~\yrcite{gatys2016image} leverage different levels of representations to separate image content from style for image style transfer task; Chen et al.~\yrcite{chen2017photographic} applied the strategy to image synthesis task.

Based on the above idea, we select layers for each classifier as described in Tab.~\ref{tab:layers}. 
For neural networks that have block structures, we pick the very last layers of each block (i.e., \textsc{block}-i for $i_{th}$ block), and, additionally, the immediate layer before the first block (i.e., \textsc{pre-block}), e.g., the first max-pooling layer in ResNet, the layer just before the fully-connected layers at the end of neural networks (i.e., \textsc{Penultimate}), and the logits layer (i.e., \textsc{Logits}).
For VGG, which does not have a block structure, we choose 5 different resolutions of layers, similar to  
Gatys et al.~\yrcite{gatys2016image} and Chen et al.~\yrcite{chen2017photographic}: Specifically, we use feature vectors after every max-pooling layer where the resolution of the feature map changes (i.e., \textsc{Maxpool}-i).

Note that the extracted feature vector from each layer is always converted to a single-dimensional vector by a pooling operation. 
For convolutional neural networks, we do spacial avg-pooling for each layer output.
For Transformer architectures, we do avg-pooling w.r.t. token length.

\subsection{Baseline Post-hoc Methods}

Here, we describe the details of the implementation of the baseline post-hoc methods.

\begin{itemize}

\item Temperature scaling (TS) \cite{guo_calibration_2017}: We use the implementation from the GitHub repository\footnote{\label{footnote:zhang-github}https://github.com/zhang64-llnl/Mix-n-Match-Calibration} provided by Zhang et al. \cite{zhang2020mix}.

\item Ensemble Temperature scaling (ETS) \cite{zhang2020mix}: Ensemble version of TS with 4 parameters. We use the official implementation from their GitHub repository\footref{footnote:zhang-github}.

\item Isotonic Regression for multi-class (IR) \cite{zhang2020mix}: Decomposes the problem as one-versus-all problem to extend Isotonic Regression for multi-class setting. We use the official implementation from their GitHub repository\footref{footnote:zhang-github}.

\item Accuracy preserving version of Isotonic Regression (IRM) \cite{zhang2020mix}: We use the official implementation from their GitHub repository\footref{footnote:zhang-github}.

\item Parameterized Temperature scaling (PTS) \cite{tomani2022parameterized}: Sample-wise version of TS. Following their paper, PTS was trained as a neural network with 2 fully connected hidden layers with 5 nodes each, using a learning rate of 0.00005, batch size of 1000, and step size of 100,000. The top 10 most confident predictions were used as input.

\item Dirichlet calibration (DIR) \cite{milios2018dirichlet}: Matrix scaling with off-diagonal and intercept regularization. 
Among their variants, we use MS-ODIR, which is intended for calibrating logits rather than probabilities and is claimed as the best-performing variant by the authors. 
We use the official implementation from their GitHub repository\footnote{\label{footnote:dirichlet}\text{https://github.com/dirichletcal/experiments\_dnn}}. 
However, we encountered some difficulties and could not adapt the code to run ImageNet experiments.

\item Intra-order preserving calibration (DIA) \cite{rahimi2020intra}: Among their variants, we use the method DIAG (diagonal intra-order-preserving), which works the best on average for various datasets and classifiers.
We use their official implementation \footnote{https://github.com/AmirooR/IntraOrderPreservingCalibration}. DIA can be run with or without hyperparameter optimization; for a fair comparison, since our other baselines, including DAC, have fixed hyperparameters, we show results without hyperparameter optimization. 

\item Calibration using Splines (SPL) \cite{gupta2020calibration}: We use their official implementation \footnote{https://github.com/kartikgupta-at-anu/spline-calibration}. Following their paper, we use the natural cubic spline fitting method with 6 knots for all our experiments.

\end{itemize}

\subsection{Training and Evaluation of DAC}

Our post-hoc calibration method DAC does not require any GPU for the training phase as well as the inference phase.
For computing $s_l$ via the k-nearest neighbor method (equation~\eqref{eq:weighting_scheme}),
we use the Faiss library\footnote{https://github.com/facebookresearch/faiss} \cite{johnson2019billion} for efficient and fast similarity search, which can be run with a CPU or GPU.
To minimize the loss function described in equation~\eqref{eq:lossece}, we use Scipy optimization using a CPU.

\section{Results for Additional Baseline-Calibration Methods}
\label{appendix:additional_methods}

\begin{figure*}[!htbp]
    \centering
    \begin{tabular}{cccc}
    \includegraphics[width=0.47\textwidth]{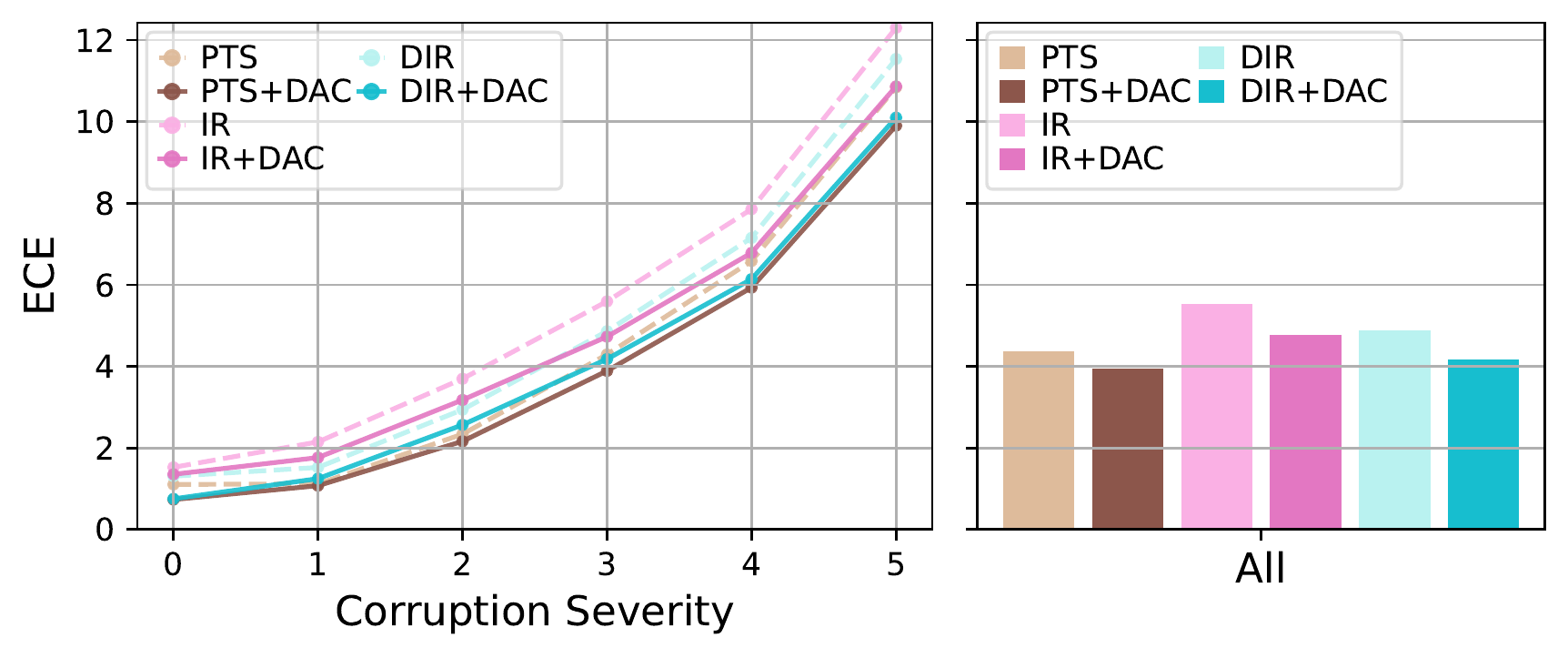} &
    \includegraphics[width=0.47\textwidth]{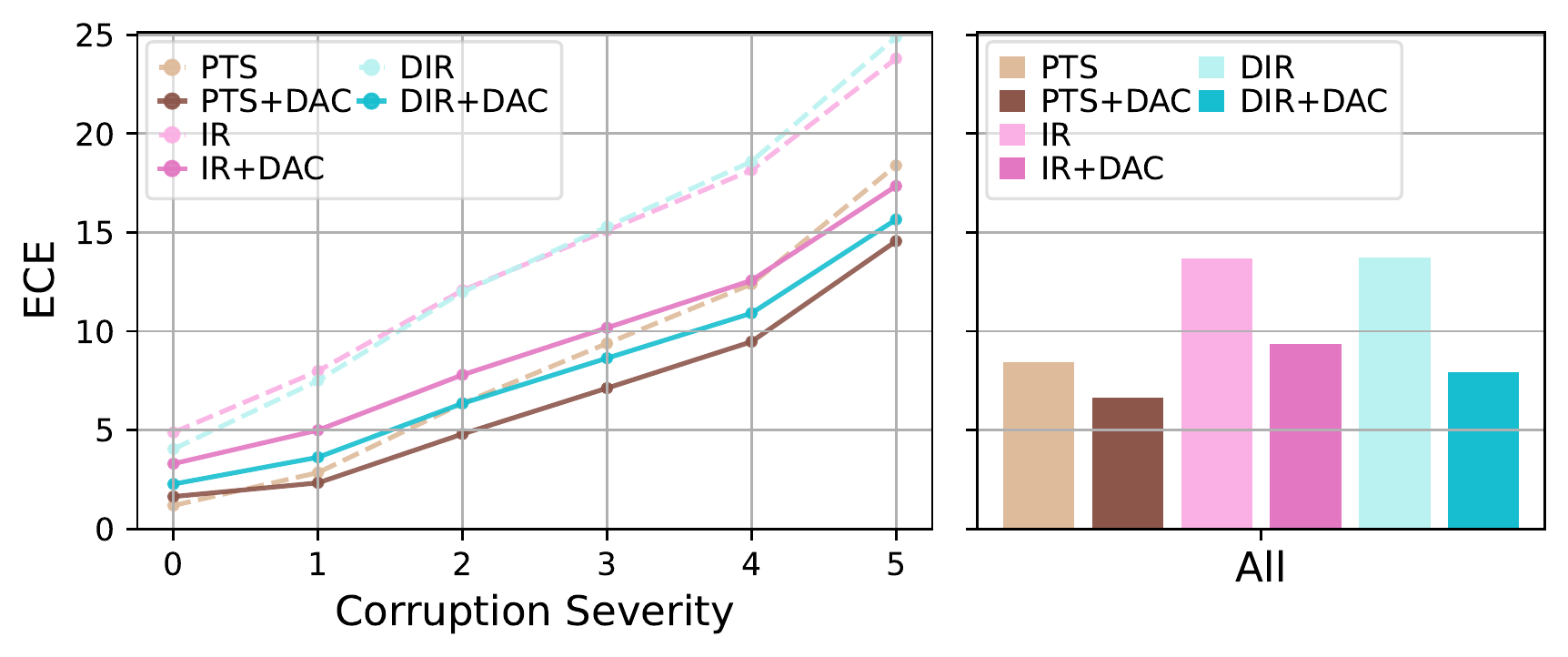} \\  
    {CIFAR10-ResNet18} & {CIFAR100-VGG16}\\  

    \includegraphics[width=0.47\textwidth]{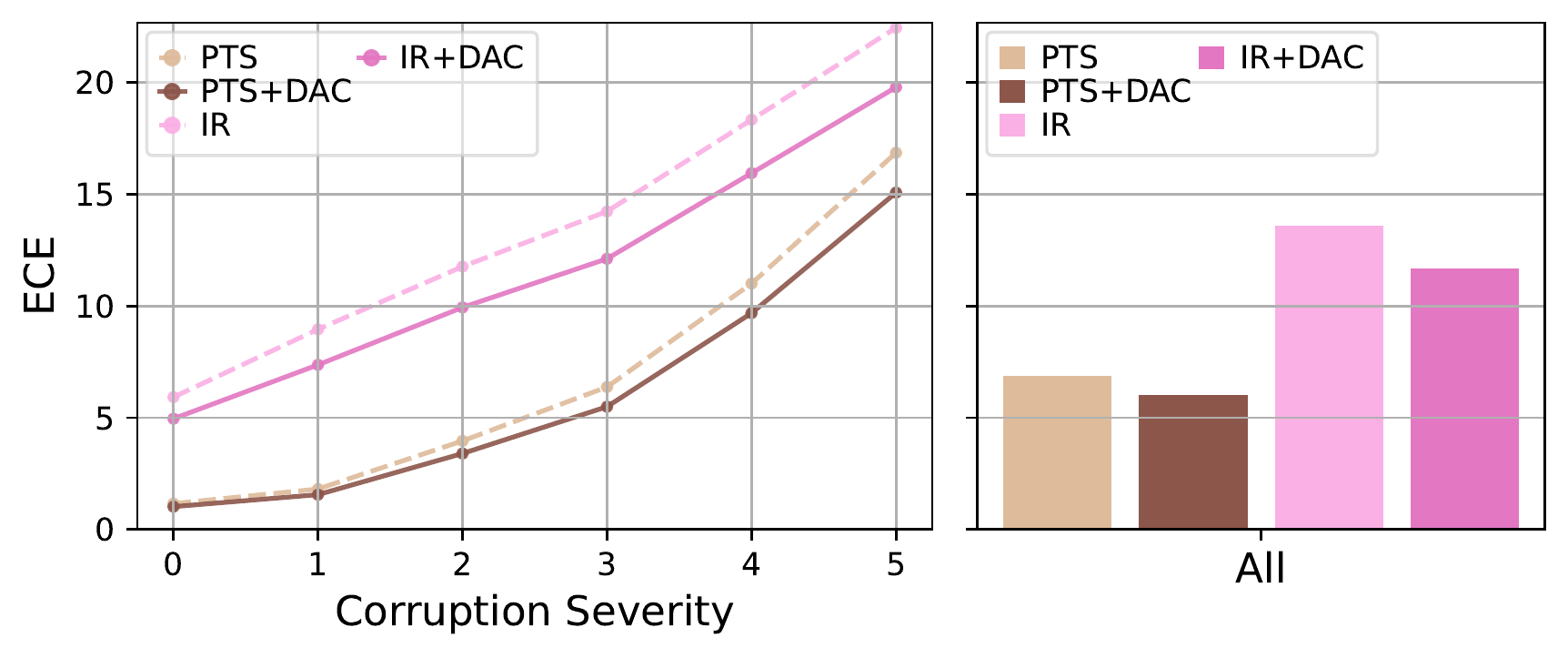} &
    \includegraphics[width=0.47\textwidth]{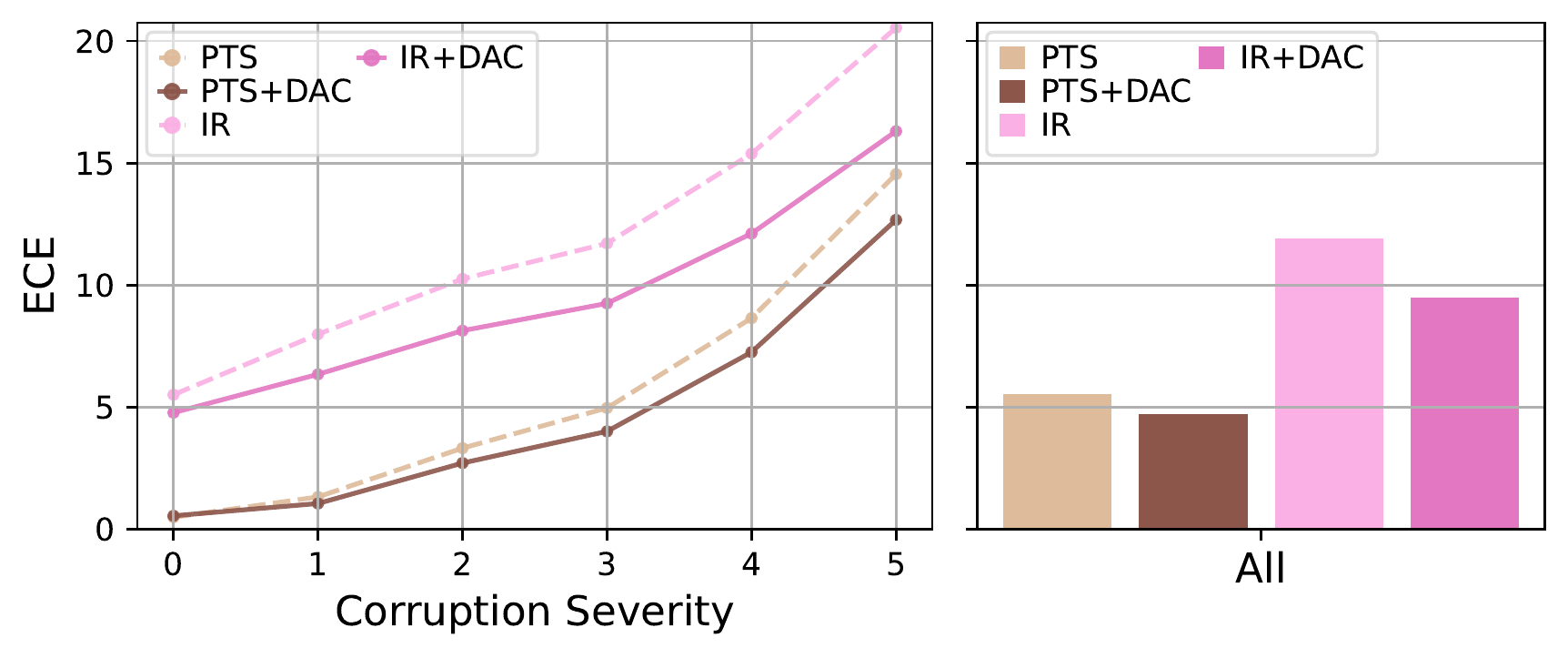} \\  
    {ImageNet-DenseNet169} & {ImageNet-BiT-M}\\  

    \includegraphics[width=0.47\textwidth]{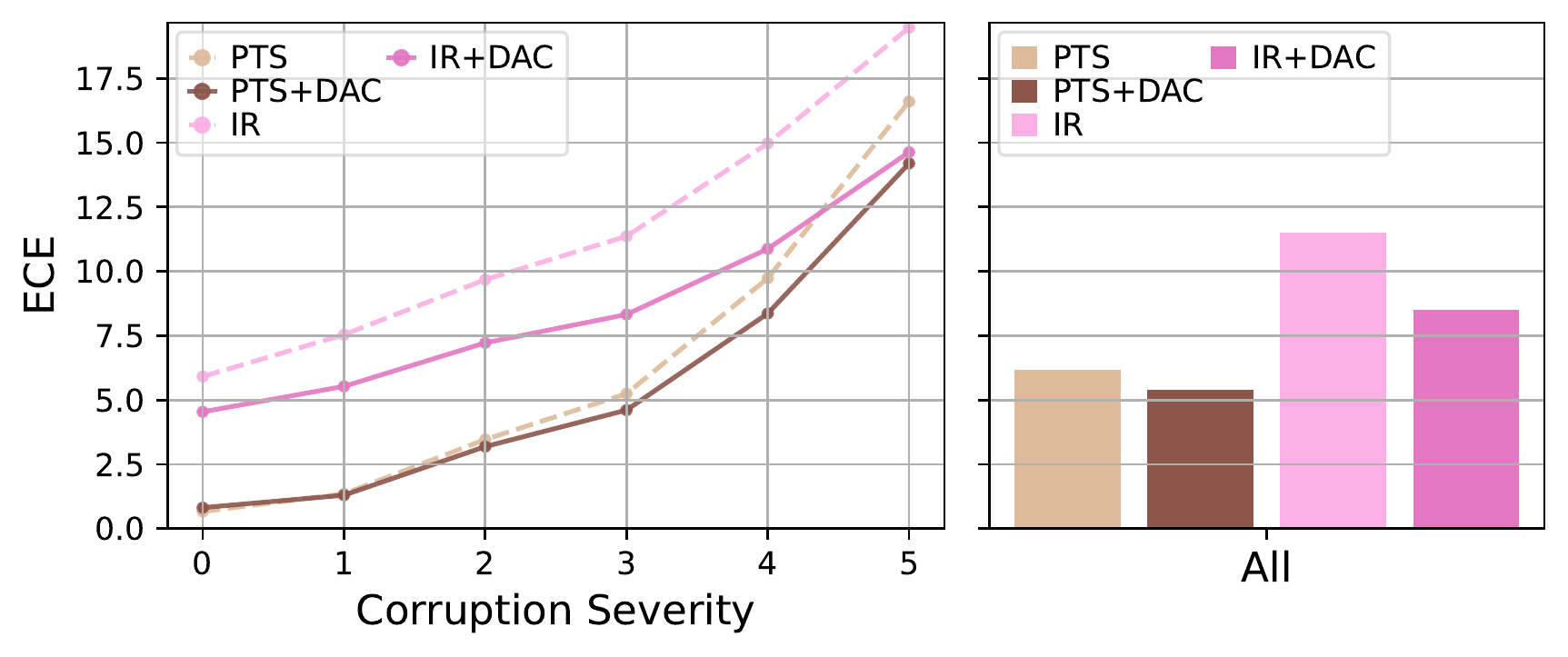} &
    \includegraphics[width=0.47\textwidth]{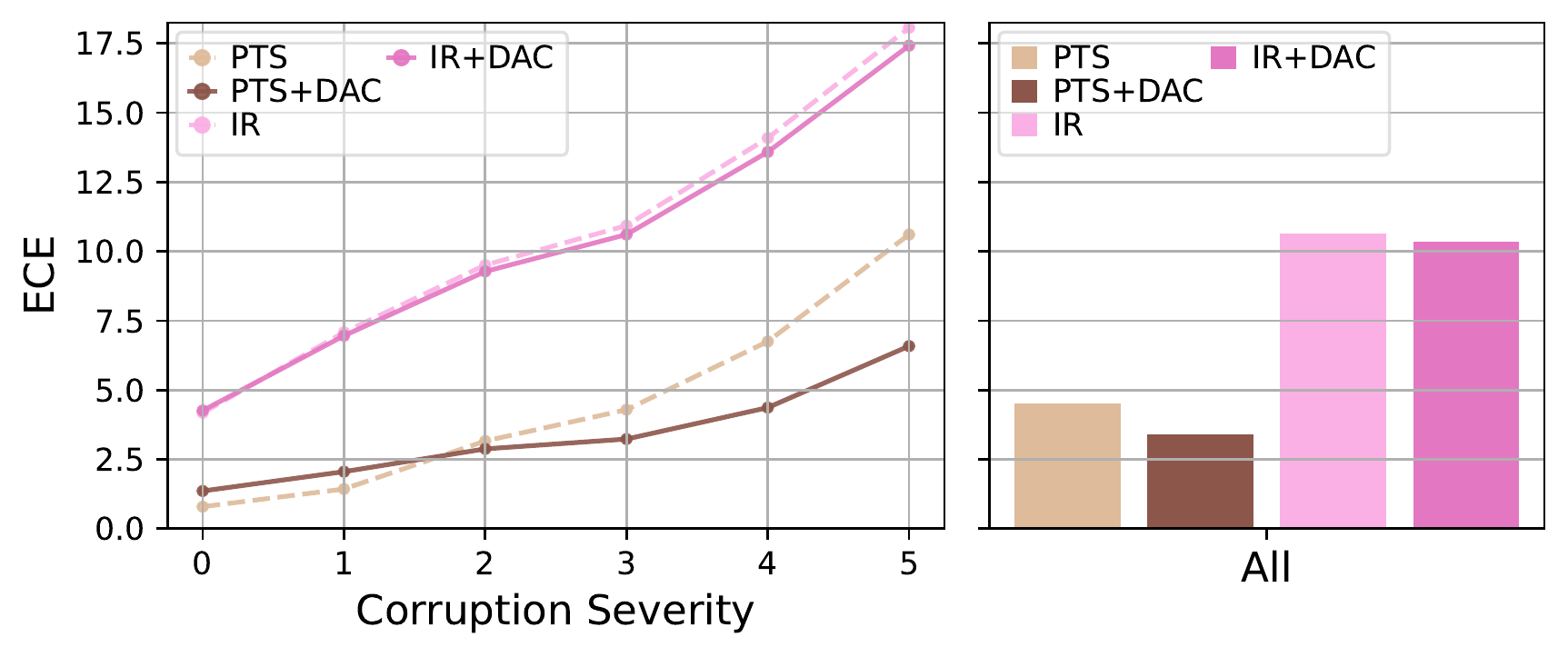} \\  
    {ImageNet-ResNeXt-WSL} & {ImageNet-ViT-B}\\  
    \end{tabular}
	\caption{\textbf{ECE for additional post-hoc methods:} Expected calibration error ($\times 10^2$) of post-hoc methods  with and without our method DAC for different model and dataset combinations. \textbf{Line plots}: Macro-averaged ECE across all corruption types shown for each corruption severity, from in-domain to OOD. \textbf{Bar plots}: Macro-averaged ECE across all corruption types as well as across all severities (lower ECE is better).}
\label{fig:ece_line_bar_combined_assessment_additional}
\end{figure*}

In addition to the state-of-the-art post-hoc calibration method which we compared in the main text of the paper, we show another 3 commonly used methods in this section and combine them with our proposed DAC method. Across all methods, classifiers, and datasets we see consistent improvements in post-hoc methods combined with DAC for macro-averaged ECE (calculated across all corruptions and levels of severity) in Tab.~\ref{tab:mean_ece_additional}. 
Moreover, in Tab.~\ref{tab:delta_ece_additional} we show the additional gain of the model and in Fig.~\ref{fig:ece_line_bar_combined_assessment_additional} we demonstrate how ECE behaves across different severity levels of corruption. 

\begin{table*}[!htbp]
\vskip 0.15in
\caption{\textbf{ECE ($\times 10^2$) for Additional calibration post-hoc methods:} Mean expected calibration error across all test domain-shift scenarios. For each model, the macro-averaged ECE (with equal-width binning and 15 bins) is computed across all corruptions from severity=0 (in-domain) until severity=5 (heavily corrupted). 
Also for these additional post-hoc calibration methods results are consistently better calibrated when paired with our method. (lower ECE is better)}
\label{tab:mean_ece_additional}
\begin{center}
\begin{small}
\begin{sc}
\begin{tabular}{l|c|ccc|ccc}
    \toprule
    & \multicolumn{1}{|c|}{\bfseries Uncal} &
    \multicolumn{3}{c|}{\bfseries Baseline Calibration Methods} &
    \multicolumn{3}{c}{\bfseries Combination with DAC (Ours)} \\
    & - & PTS &    IR &   DIR & PTS &    IR &   DIR\\
\midrule
    C10 ResNet18 &            19.27 &              4.99 &  6.27 &  5.55 &     \textbf{4.55} &   5.42 & \underline{4.80} \\
       C10 VGG16 &            19.05 &  \underline{6.74} &  7.92 &  8.48 &     \textbf{5.66} &   6.91 &             7.62 \\
 C10 DenseNet121 &            19.26 &  \underline{4.75} &  6.61 &  7.47 &     \textbf{4.64} &   5.88 &             6.38 \\
 \hline
   C100 ResNet18 &            16.44 &             11.48 & 14.40 & 12.02 & \underline{10.81} &  12.04 &   \textbf{10.58} \\
      C100 VGG16 &            34.41 &              9.78 & 15.31 & 15.52 &     \textbf{7.59} &  10.50 & \underline{8.97} \\
C100 DenseNet121 &            23.83 & \underline{10.67} & 13.74 & 13.02 &    \textbf{10.53} &  12.11 &            11.77 \\    \hline
   IMG ResNet152 &            10.50 &  \underline{7.32} & 12.41 &     - &     \textbf{4.78} &   9.91 &                - \\
 IMG DenseNet169 &            13.28 &  \underline{7.93} & 15.05 &     - &     \textbf{6.98} &  12.94 &                - \\
    IMG Xception &            30.49 &  \underline{9.15} & 17.94 &     - &     \textbf{9.13} &  14.29 &                - \\
     \hline
       IMG BiT-M &            11.71 &  \underline{6.50} & 13.10 &     - &     \textbf{5.49} &  10.37 &                - \\
 IMG ResNeXt-WSL &            15.44 &  \underline{7.21} & 12.53 &     - &     \textbf{6.27} &   9.26 &                - \\
       IMG ViT-B &    \textbf{3.78} &              5.20 & 11.85 &     - &  \underline{3.80} &  11.49 &                - \\
    \bottomrule
\end{tabular}
\end{sc}
\end{small}
\end{center}
\vskip -0.1in
\end{table*}

\begin{table*}[!htbp]
\centering
\vskip 0.15in
\caption{\textbf{Deltas for Additional post-hoc methods:} Difference in expected calibration error ($\times 10^2$) of post-hoc calibration methods with and without our method DAC. \textbf{$InD$}: In-domain, \textbf{$SEV. 5$}: Heavily corrupted (severity of 5) and \textbf{$ALL$}: Macro-averaged ECE across all corruptions from severity=0 (in-domain) until severity=5 (heavily corrupted). For \textbf{$ALL$}, we additionally report the ratio to indicate overall performance gain. (negative deltas are better)}
\vskip 0.15in
\label{tab:delta_ece_additional}
    \begin{small}
    \begin{sc}
    \begin{tabular}{crrr}
    \toprule
    & \multicolumn{1}{c}{InD} & \multicolumn{1}{c}{Sev. 5} & \multicolumn{1}{c}{All} \\ 
    \midrule
    PTS & \minusbar{-0.37}{-0.1287}& \minusbar{-0.94}{-0.3267}& \minusbar{-0.43}{-0.1497}\;\;(9.8\%) \\
IR & \minusbar{-0.17}{-0.0596}& \minusbar{-1.44}{-0.4995}& \minusbar{-0.74}{-0.2577}(13.4\%) \\
DIR & \minusbar{-0.56}{-0.1930}& \minusbar{-1.44}{-0.5000}& \minusbar{-0.72}{-0.2511}(14.8\%) \\
    \bottomrule
    \multicolumn{4}{c}{a.) CIFAR10 ResNet18}
    \end{tabular}
    \end{sc}
    \end{small}
\medskip
    \begin{small}
    \begin{sc}
    \begin{tabular}{crrr}
    \toprule
    & \multicolumn{1}{c}{InD} & \multicolumn{1}{c}{Sev. 5} & \multicolumn{1}{c}{All} \\ 
    \midrule
    PTS & \plusbar{0.45}{0.0244}& \minusbar{-3.83}{-0.2075}& \minusbar{-1.77}{-0.0960}(21.0\%) \\
IR & \minusbar{-1.58}{-0.0857}& \minusbar{-6.45}{-0.3498}& \minusbar{-4.30}{-0.2332}(31.5\%) \\
DIR & \minusbar{-1.77}{-0.0961}& \minusbar{-9.22}{-0.5000}& \minusbar{-5.80}{-0.3147}(42.3\%) \\
    \bottomrule
    \multicolumn{4}{c}{b.) CIFAR100 VGG16}
    \end{tabular}
    \end{sc}
    \end{small}
\hfil
    \begin{small}
    \begin{sc}
    \begin{tabular}{crrr}
    \toprule
    & \multicolumn{1}{c}{InD} & \multicolumn{1}{c}{Sev. 5} & \multicolumn{1}{c}{All} \\ 
    \midrule
PTS & \minusbar{-0.14}{-0.0260}& \minusbar{-1.79}{-0.3355}& \minusbar{-0.82}{-0.1548}(12.0\%) \\
IR & \minusbar{-0.97}{-0.1816}& \minusbar{-2.66}{-0.5000}& \minusbar{-1.93}{-0.3618}(14.1\%) \\
\bottomrule
    \multicolumn{4}{c}{c.) ImageNet DenseNet169}
    \end{tabular}
    \end{sc}
    \end{small}
\medskip
\begin{small}
    \begin{sc}
    \begin{tabular}{crrr}
    \toprule
    & \multicolumn{1}{c}{InD} & \multicolumn{1}{c}{Sev. 5} & \multicolumn{1}{c}{All} \\ 
    \midrule
PTS & \plusbar{0.06}{0.0070}& \minusbar{-1.87}{-0.2207}& \minusbar{-0.84}{-0.0992}(15.1\%) \\
IR & \minusbar{-0.73}{-0.0868}& \minusbar{-4.24}{-0.5000}& \minusbar{-2.41}{-0.2845}(20.2\%) \\
\bottomrule
    \multicolumn{4}{c}{d.) ImageNet BiT-M}
    \end{tabular}
    \end{sc}
    \end{small}
\hfil
    \begin{small}
    \begin{sc}
    \begin{tabular}{crrr}
    \toprule
    & \multicolumn{1}{c}{InD} & \multicolumn{1}{c}{Sev. 5} & \multicolumn{1}{c}{All} \\ 
    \midrule
PTS & \plusbar{0.15}{0.0156}& \minusbar{-2.40}{-0.2484}& \minusbar{-0.76}{-0.0790}(12.4\%) \\
IR & \minusbar{-1.37}{-0.1415}& \minusbar{-4.84}{-0.5000}& \minusbar{-2.97}{-0.3069}(25.8\%) \\
\bottomrule
    \multicolumn{4}{c}{e.) ImageNet ResNeXt-WSL}
    \end{tabular}
    \end{sc}
    \end{small}
\medskip
\begin{small}
    \begin{sc}
    \begin{tabular}{crrr}
    \toprule
    & \multicolumn{1}{c}{InD} & \multicolumn{1}{c}{Sev. 5} & \multicolumn{1}{c}{All} \\ 
    \midrule
PTS & \plusbar{0.57}{0.0704}& \minusbar{-4.02}{-0.5000}& \minusbar{-1.09}{-0.1358}(24.2\%) \\
IR & \plusbar{0.08}{0.0097}& \minusbar{-0.64}{-0.0802}& \minusbar{-0.29}{-0.0362}\;\;(2.7\%) \\
\bottomrule
    \multicolumn{4}{c}{f.) ImageNet ViT-B}
    \end{tabular}
    \end{sc}
    \end{small}
\end{table*}

\setlength{\tabcolsep}{0.5pt}
\renewcommand{\arraystretch}{0.5}
\setlength{\tabcolsep}{6pt}
\renewcommand{\arraystretch}{1}

\clearpage

\section{Results for Additional Calibration Measures}
\label{appendix:additional_measures}

Even though ECE is most commonly used for evaluating calibration performance, other metrics have been proposed as well. Here we want to show that our results on ECE with equal-width binning are consistent with various other calibration methods. We additionally evaluate based on: \textbf{a.} ECE using equal-mass binning (Tab.~\ref{tab:mean_ece_hist}), \textbf{b.} ECE based on kernel density estimation (ECE-KDE) \cite{zhang2020mix} (Tab.~\ref{tab:mean_ece_kde}) and \textbf{c.} class-wise ECE \cite{kull2019beyond} (Tab.~\ref{tab:mean_ece_class_wise}). Moreover, we show that negative log-likelihood (Tab.~\ref{tab:mean_nll}) is kept similar for our method compared to baseline calibration methods without DAC. In each of these tables, we report macro-averaged ECE or NLL scores. For each model, we compute the macro-averaged ECE or NLL across all corruptions from severity=0 (in-domain) until severity=5 (OOD). For spline authors didn't provide ways to calibrate the probabilistic predictions; that is, why we were not able to calculate the negative log-likelihood, and class-wise ECE for this method.

\begin{table*}[!h]
\vskip 0.15in
\caption{Mean \textbf{ECE ($\times 10^2$) with equal-mass binning} (15 bins) computed across all corruptions from severity=0 (in-domain) until severity=5 (heavily corrupted). (lower ECE is better).}
\label{tab:mean_ece_hist}
\begin{center}
\begin{small}
\begin{sc}
\begin{tabular}{l|c|ccccc|ccccc}
    \toprule
    & \multicolumn{1}{|c|}{\bfseries Uncal} &
    \multicolumn{5}{c|}{\bfseries Baseline Calibration Methods} &
    \multicolumn{5}{c}{\bfseries Combination with DAC (Ours)} \\
    & - & TS & ETS  & IRM & DIA & SPL & TS & ETS  & IRM & DIA & SPL\\
    \midrule
      C10 ResNet18 & 19.26 &          4.93 &             4.93 &  5.91 &  6.45 &             6.26 & \underline{4.45} &    \textbf{4.37} &             4.81 &          4.60 &             4.77 \\
       C10 VGG16 & 19.05 &          6.31 &             6.36 &  7.44 &  9.88 &             7.52 &    \textbf{5.60} &    \textbf{5.60} &             6.18 &          6.33 &             5.68 \\
 C10 DenseNet121 & 19.26 &          5.20 &             5.20 &  6.61 &  7.84 &             6.59 & \underline{4.57} & \underline{4.57} &             5.63 &          6.62 &    \textbf{4.42} \\ \hline
   C100 ResNet18 & 16.41 &         11.37 &            10.72 & 12.25 &  9.24 &            10.39 &            10.65 & \underline{9.11} &            10.01 & \textbf{8.45} &             9.75 \\
      C100 VGG16 & 34.40 &         11.57 &            11.57 & 13.21 & 14.69 &            10.75 & \underline{6.49} &    \textbf{6.48} &             8.06 &         10.30 &             7.48 \\
C100 DenseNet121 & 23.83 &          8.78 &             8.76 & 12.03 & 11.02 &             9.72 & \underline{8.74} &    \textbf{8.40} &             9.96 &         15.07 &             9.17 \\ \hline
   IMG ResNet152 & 10.49 &          4.47 &             4.05 &  5.18 &  7.16 &             5.56 & \underline{3.48} &    \textbf{3.37} &             3.49 &          3.64 &             3.65 \\
 IMG DenseNet169 & 13.27 &          6.58 &             6.39 &  7.36 &  8.43 &             7.11 &             4.80 &    \textbf{3.91} & \underline{4.50} &          6.29 &             4.60 \\
    IMG Xception & 30.48 &          8.83 &             8.42 & 12.93 &  9.83 &            10.79 &             8.81 &    \textbf{8.00} & \underline{8.37} &          8.99 &             8.49 \\ \hline
       IMG BiT-M & 11.71 &          7.15 &             6.54 &  6.92 &  7.46 &             6.60 &             4.40 & \underline{3.99} &             4.19 &          5.52 &    \textbf{3.75} \\
 IMG ResNeXt-WSL & 15.40 &          8.01 &             8.01 &  8.03 &  8.31 &             6.17 &             7.29 &             5.91 & \underline{5.73} &          6.30 &    \textbf{3.97} \\
       IMG ViT-B &  3.77 &          4.22 &             3.76 &  4.21 &  5.84 &             3.94 &             3.79 &    \textbf{3.39} &             3.96 &          5.51 & \underline{3.58} \\
    \bottomrule
\end{tabular}
\end{sc}
\end{small}
\end{center}
\vskip -0.1in
\end{table*}

\begin{table*}[!h]
\vskip 0.15in
\caption{Mean \textbf{ECE-KDE} ($\times 10^2$) computed across all corruptions from severity=0 (in-domain) until severity=5 (heavily corrupted). (lower ECE-KDE is better).}
\label{tab:mean_ece_kde}
\begin{center}
\begin{small}
\begin{sc}
\begin{tabular}{l|c|ccccc|ccccc}
    \toprule
    & \multicolumn{1}{|c|}{\bfseries Uncal} &
    \multicolumn{5}{c|}{\bfseries Baseline Calibration Methods} &
    \multicolumn{5}{c}{\bfseries Combination with DAC (Ours)} \\
    & - & TS & ETS  & IRM & DIA & SPL & TS & ETS  & IRM & DIA & SPL\\
    \midrule
     C10 ResNet18 & 18.63 &          4.93 &  4.93 &             5.78 &  6.34 &             6.27 & \underline{4.35} &    \textbf{4.30} &             4.67 &          4.58 &             4.75 \\
       C10 VGG16 & 18.44 &          6.25 &  6.30 &             7.15 &  9.74 &             7.53 &    \textbf{5.56} &    \textbf{5.56} &             5.94 &          6.31 &             5.67 \\
 C10 DenseNet121 & 18.64 &          5.17 &  5.17 &             6.30 &  7.61 &             6.51 & \underline{4.51} & \underline{4.51} &             5.20 &          6.47 &    \textbf{4.35} \\ \hline
   C100 ResNet18 & 15.99 &         11.08 & 10.55 &            11.93 &  9.13 &            10.37 &            10.39 & \underline{9.05} &             9.69 & \textbf{8.33} &             9.67 \\
      C100 VGG16 & 33.83 &         11.54 & 11.54 &            13.14 & 14.46 &            10.80 & \underline{6.50} &    \textbf{6.49} &             7.81 &         10.14 &             7.37 \\
C100 DenseNet121 & 23.26 &          8.79 &  8.83 &            11.66 & 10.99 &             9.79 & \underline{8.74} &    \textbf{8.47} &             9.58 &         14.86 &             9.15 \\ \hline
   IMG ResNet152 & 10.28 &          4.44 &  4.16 &             5.11 &  7.07 &             5.66 &    \textbf{3.49} &             3.55 & \underline{3.52} &          3.72 &             3.71 \\
 IMG DenseNet169 & 13.04 &          6.49 &  6.39 &             7.25 &  8.32 &             7.20 &             4.75 &    \textbf{4.01} & \underline{4.43} &          6.21 &             4.66 \\
    IMG Xception & 30.18 &          8.89 &  8.54 &            12.93 &  9.84 &            10.93 &             8.87 &    \textbf{8.15} & \underline{8.35} &          9.01 &             8.53 \\ \hline
       IMG BiT-M & 11.60 &          7.14 &  6.59 &             6.86 &  7.45 &             6.66 &             4.44 & \underline{4.10} &             4.13 &          5.51 &    \textbf{3.80} \\
 IMG ResNeXt-WSL & 14.90 &          7.55 &  7.55 &             7.74 &  8.07 &             6.32 &             6.82 &             5.66 & \underline{5.43} &          6.13 &    \textbf{4.01} \\
       IMG ViT-B &  3.72 &          4.15 &  3.78 &             4.07 &  5.75 &             3.96 &             3.74 &    \textbf{3.42} &             3.82 &          5.44 & \underline{3.63} \\
    \bottomrule
\end{tabular}
\end{sc}
\end{small}
\end{center}
\vskip -0.1in
\end{table*}

\begin{table*}[!h]
\vskip 0.15in
\caption{Mean \textbf{Class-wise ECE} ($\times 10^2$) computed across all corruptions from severity=0 (in-domain) until severity=5 (heavily corrupted). (lower class-wise ECE is better). Note that since SPL only calibrates the highest predicted confidence, it is not directly possible to evaluate class-wise ECE. }
\label{tab:mean_ece_class_wise}
\begin{center}
\begin{small}
\begin{sc}
\begin{tabular}{l|c|cccc|cccc}
    \toprule
    & \multicolumn{1}{|c|}{\bfseries Uncal} &
    \multicolumn{4}{c|}{\bfseries Baseline Calibration Methods} &
    \multicolumn{4}{c}{\bfseries Combination with DAC (Ours)} \\
    & - & TS & ETS  & IRM & DIA  & TS & ETS  & IRM & DIA \\
    \midrule
     C10 ResNet18 &          41.33 &          22.96 &             22.96 & 22.94 & 24.46 & \underline{22.45} &    \textbf{22.43} &             22.84 &          22.66 \\
       C10 VGG16 &          40.72 &          24.44 &             24.46 & 24.80 & 28.41 &    \textbf{24.07} &    \textbf{24.07} &             24.37 &          24.68 \\
 C10 DenseNet121 &          41.49 &          24.70 &             24.70 & 24.63 & 25.82 &    \textbf{24.04} &    \textbf{24.04} &             24.20 &          24.86 \\ \hline
   C100 ResNet18 &          54.32 &          48.89 &             48.42 & 54.09 & 48.44 &             48.35 & \underline{46.73} &             52.22 & \textbf{46.53} \\
      C100 VGG16 &          80.43 &          47.73 &             47.73 & 48.30 & 52.59 &    \textbf{42.11} &    \textbf{42.11} &             45.63 &          46.22 \\
C100 DenseNet121 &          61.97 &          45.78 &             45.82 & 51.99 & 48.94 & \underline{45.70} &    \textbf{45.56} &             51.74 &          56.51 \\ \hline
   IMG ResNet152 &          58.49 &          54.61 &             54.25 & 62.23 & 58.77 & \underline{52.76} &    \textbf{52.73} &             54.18 &          55.49 \\
 IMG DenseNet169 &          61.05 &          56.71 &             56.29 & 61.32 & 59.90 & \underline{55.37} &    \textbf{54.68} &             56.54 &          58.20 \\ 
    IMG Xception &          84.33 &          54.22 & \underline{53.93} & 80.07 & 64.24 &             54.19 &    \textbf{53.38} &             61.96 &          63.65 \\ \hline
       IMG BiT-M &          55.78 &          53.45 &             53.05 & 58.28 & 56.64 & \underline{50.63} &    \textbf{50.44} &             54.16 &          55.11 \\
 IMG ResNeXt-WSL &          46.54 &          38.87 &             38.87 & 47.64 & 45.08 &    \textbf{37.60} & \underline{37.77} &             40.40 &          43.00 \\
       IMG ViT-B &          49.95 &          50.10 & \underline{49.91} & 51.23 & 53.36 &             49.96 &    \textbf{49.78} &             51.35 &          53.17 \\
\bottomrule
\end{tabular}
\end{sc}
\end{small}
\end{center}
\vskip -0.1in
\end{table*}

\begin{table*}[!h]
\vskip 0.15in
\caption{Mean \textbf{negative log-likelihood} computed across all corruptions from severity=0 (in-domain) until severity=5 (heavily corrupted). Note that since SPL only calibrates the highest predicted confidence, it is not directly possible to evaluate negative log-likelihood.}
\label{tab:mean_nll}
\begin{center}
\begin{small}
\begin{sc}
\begin{tabular}{l|c|cccc|cccc}
    \toprule
    & \multicolumn{1}{|c|}{\bfseries Uncal} &
    \multicolumn{4}{c|}{\bfseries Baseline Calibration Methods} &
    \multicolumn{4}{c}{\bfseries Combination with DAC (Ours)} \\
    & - & TS & ETS  & IRM & DIA & TS & ETS  & IRM & DIA \\
    \midrule
    C10 ResNet18 & 1.4496 & 0.8713 & 0.8713 & 0.8718 & 0.9227 & 0.8666 &  0.8657 &  0.8883 &  0.8675 \\
       C10 VGG16 & 1.4500 & 0.8036 & 0.8040 & 0.8008 & 0.8912 & 0.8029 &  0.8029 &  0.8065 &  0.8068 \\
 C10 DenseNet121 & 1.3284 & 0.8918 & 0.8918 & 0.8905 & 0.9045 & 0.8827 &  0.8827 &  0.9000 &  0.8868 \\ \hline
   C100 ResNet18 & 2.4690 & 2.3330 & 2.3088 & 2.3575 & 2.2879 & 2.3356 &  2.2830 &  2.3159 &  2.2721 \\
      C100 VGG16 & 3.6090 & 2.4193 & 2.4193 & 2.4340 & 2.5827 & 2.3232 &  2.3231 &  2.3585 &  2.3899 \\
C100 DenseNet121 & 2.4672 & 2.1404 & 2.1427 & 2.1535 & 2.1400 & 2.1417 &  2.1413 &  2.1890 &  2.2775 \\ \hline
   IMG ResNet152 & 2.6844 & 2.5870 & 2.5887 & 2.4654 & 2.6669 & 2.5691 &  2.5772 &  2.5611 &  2.5763 \\
 IMG DenseNet169 & 2.7371 & 2.5975 & 2.6006 & 2.5211 & 2.6745 & 2.5677 &  2.5650 &  2.5565 &  2.6126 \\
    IMG Xception & 4.2438 & 3.3895 & 3.3977 & 3.2663 & 3.4935 & 3.3893 &  3.3915 &  3.4136 &  3.4733 \\ \hline
       IMG BiT-M & 2.3315 & 2.2874 & 2.2914 & 2.2256 & 2.3145 & 2.2689 &  2.2775 &  2.2363 &  2.2588 \\
 IMG ResNeXt-WSL & 2.1246 & 1.9379 & 1.9379 & 1.7881 & 1.9260 & 1.9407 &  1.9370 &  1.8354 &  1.8381 \\
       IMG ViT-B & 2.2560 & 2.2585 & 2.2617 & 2.3009 & 2.3107 & 2.2561 &  2.2597 &  2.2900 &  2.3044 \\
\bottomrule
\end{tabular}
\end{sc}
\end{small}
\end{center}
\vskip -0.1in
\end{table*}

\clearpage

\section{Ablation Study: How Information from Hidden Layers Benefits Calibration with DAC} 
\label{appendix:ablation_layer_choice}

\begin{figure}[h]
\centering
\begin{tabular}{cc}
\includegraphics[width=0.9\textwidth]{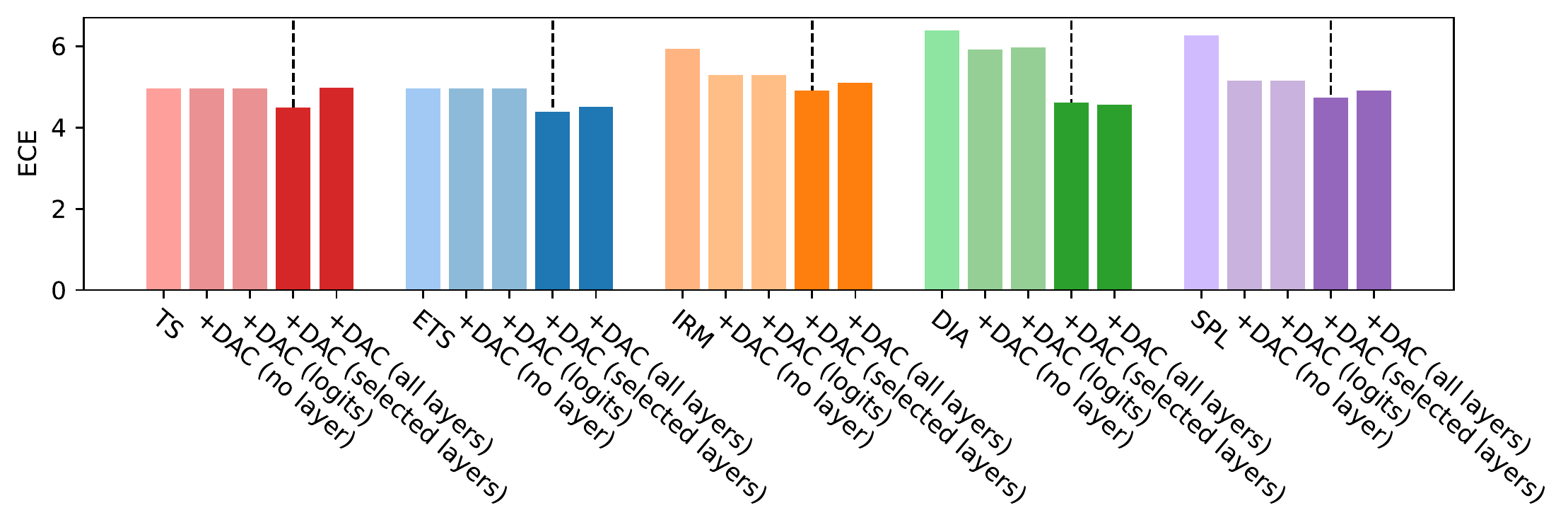} \\
CIFAR10-ResNet18 \\
\includegraphics[width=0.9\textwidth]{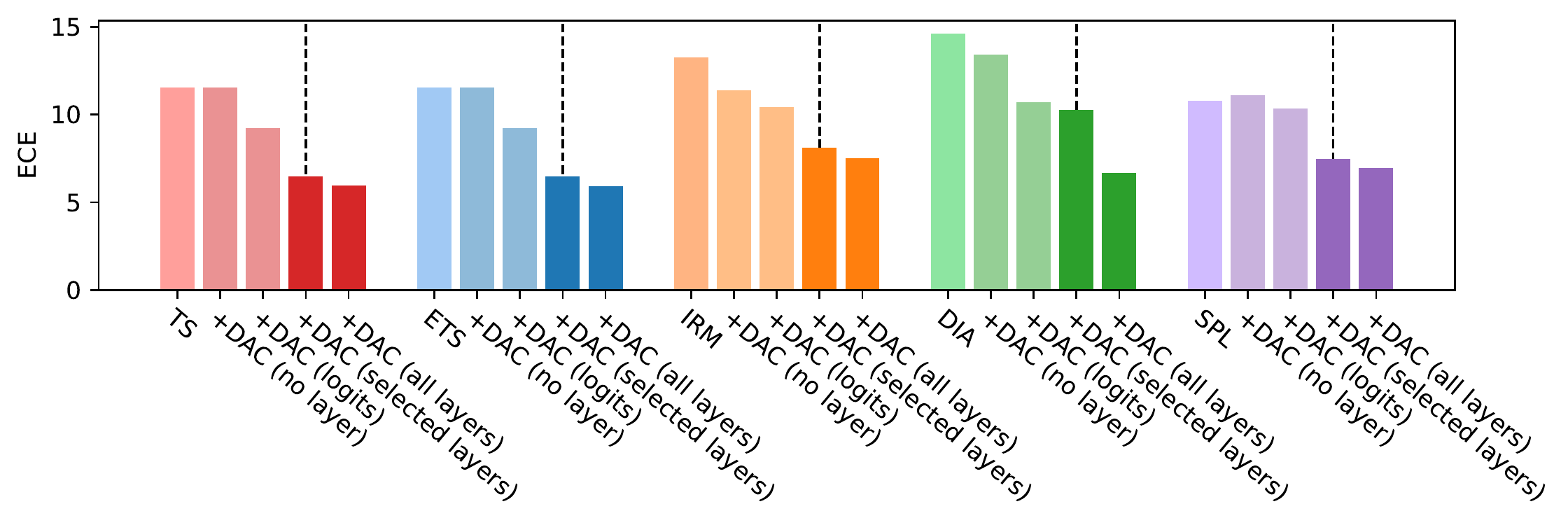} \\
CIFAR100-VGG16 \\
\end{tabular}
\caption{Ablation study on how information from hidden layers benefits the calibration performance of DAC. 
The y-axis shows the macro-averaged ECE ($\times 10^2$) computed across all corruptions from severity=0 (in-domain) until severity=5 (heavily corrupted).
DAC boosts calibration performance significantly as soon as information from hidden layers is available ("selected layers" and "all layers") compared to DAC with access to only the "logits layer".}
\label{fig:DAC_layer_choice_difference}
\end{figure}

In order to investigate the importance of hidden layers from classifiers towards the calibration performance of DAC, we conduct an ablation study. In this study, we analyze the sensitivity of ECE with regard to different subsets of hidden layers that our DAC has access to.

We compare the calibration performance of the following variations:
\begin{itemize}
    \item \textbf{w/o DAC}: Baseline post-hoc calibration method without DAC.
    \item \textbf{DAC (no layer)}: Baseline method + DAC using no layer at all. If DAC has no access to any layer in the classifier, only the bias term $w_0$ remains in equation \eqref{eq:eucledian}, thus in this setting DAC is equivalent to temperature scaling.
    \item \textbf{DAC (logits layer)}: Base method  + DAC using only the logits layer.
    \item \textbf{DAC (selected layers)}: Base method  + DAC using selected layers (a well distributed subset of layers throughout the classifier). This is the default DAC used throughout the paper.
    \item \textbf{DAC (all layers)}: Base method + DAC using all layers of the classifiers. 
\end{itemize}

Fig.~\ref{fig:DAC_layer_choice_difference} shows macro-averaged ECE across all corruptions from severity=0 (in-domain) until severity=5 (OOD).
The results indicate that DAC using solely logits layer can already boost calibration compared to the respective stand-alone baseline methods. However, purely relying on the logits layer as a basis for post-hoc calibration is still suboptimal. DAC boosts the calibration performance significantly as soon as information from hidden layers is available to DAC. However, by increasing the access further from selected layers to all layers, an increase in calibration performance can not be observed in all scenarios. This indicates that more layers can benefit DAC's performance; however, we assume that too many layers can also cause an overfitting issue due to having too many parameters to optimize.

\section{Reliability Diagrams}

Reliability diagrams allow insights into calibration performance by showing the difference between a method’s confidence in its predictions and its accuracy \cite{guo_calibration_2017}. Following Maddox et al. \cite{maddox2019simple_rel_d}, we show the following reliability diagrams.
We split the test data into 15 bins based on the confidence values, so that each bin contains a uniform number of data points. Then, we evaluate the accuracy and mean confidence for each bin. For a well-calibrated model, this difference should be close to zero for each bin.

\begin{figure*}[ht!]
    \centering
    \begin{tabular}{ccc}
    \includegraphics[width=0.30\textwidth]{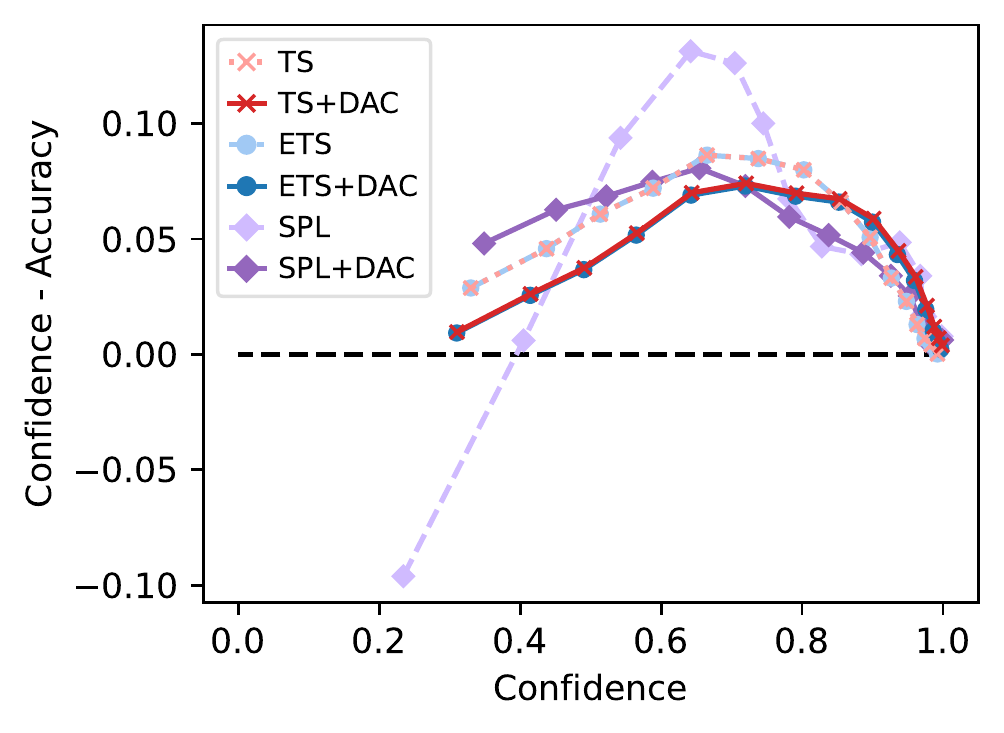} & 
    \includegraphics[width=0.30\textwidth]{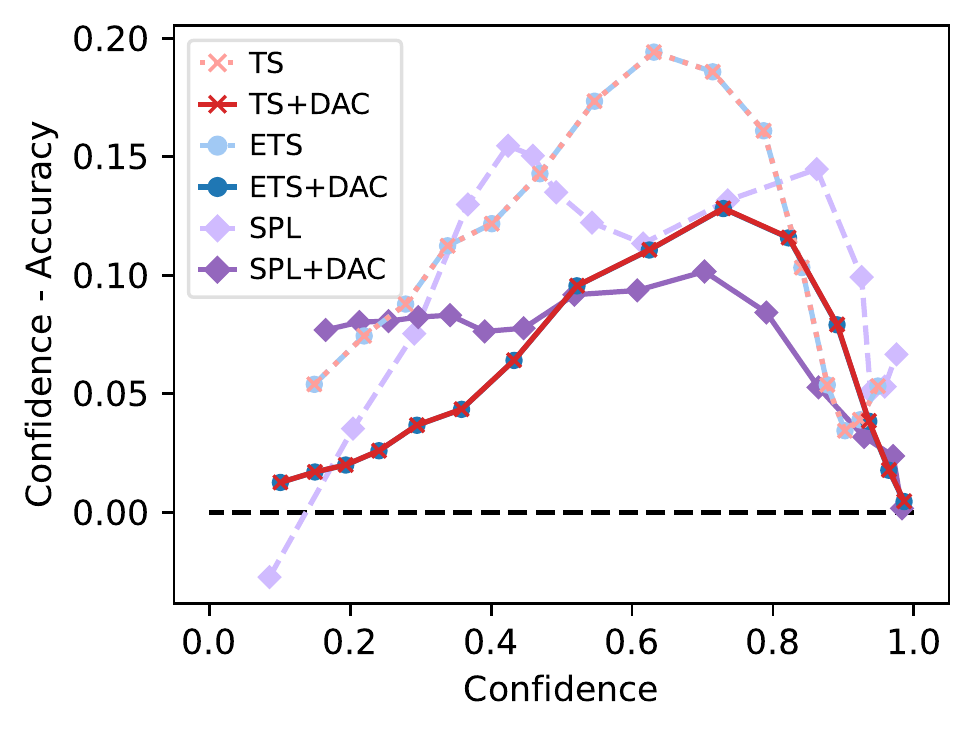} & 
    \includegraphics[width=0.30\textwidth]{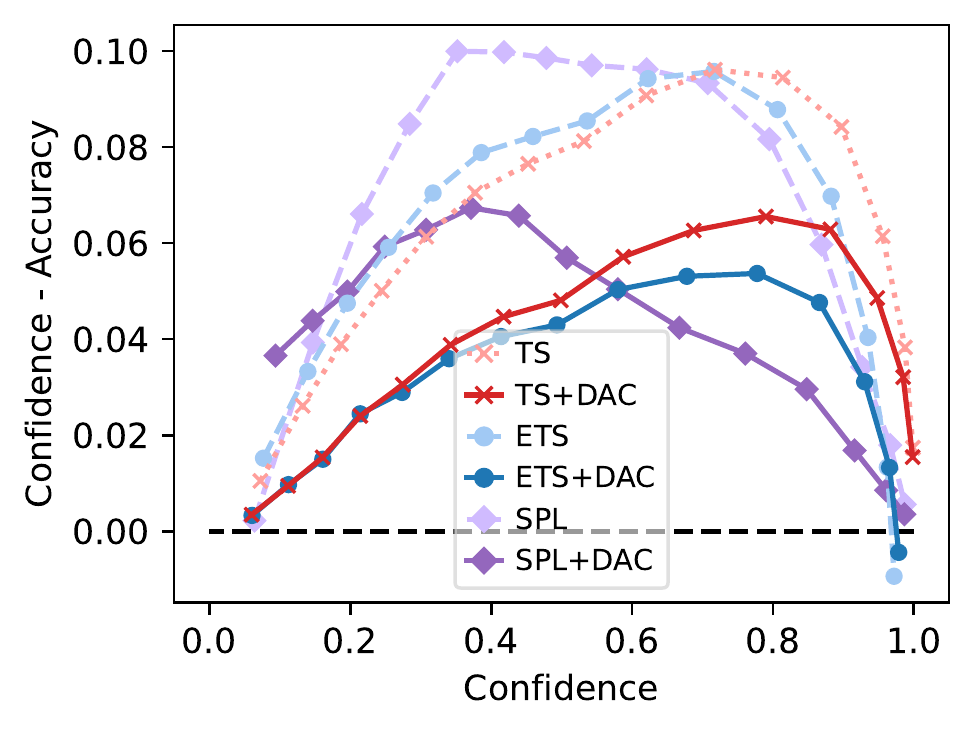} \\  
    \multicolumn{1}{c}{CIFAR10-ResNet18} & \multicolumn{1}{c}{CIFAR100-VGG16} & \multicolumn{1}{c}{ImageNet-DenseNet169}\\  
    \includegraphics[width=0.30\textwidth]{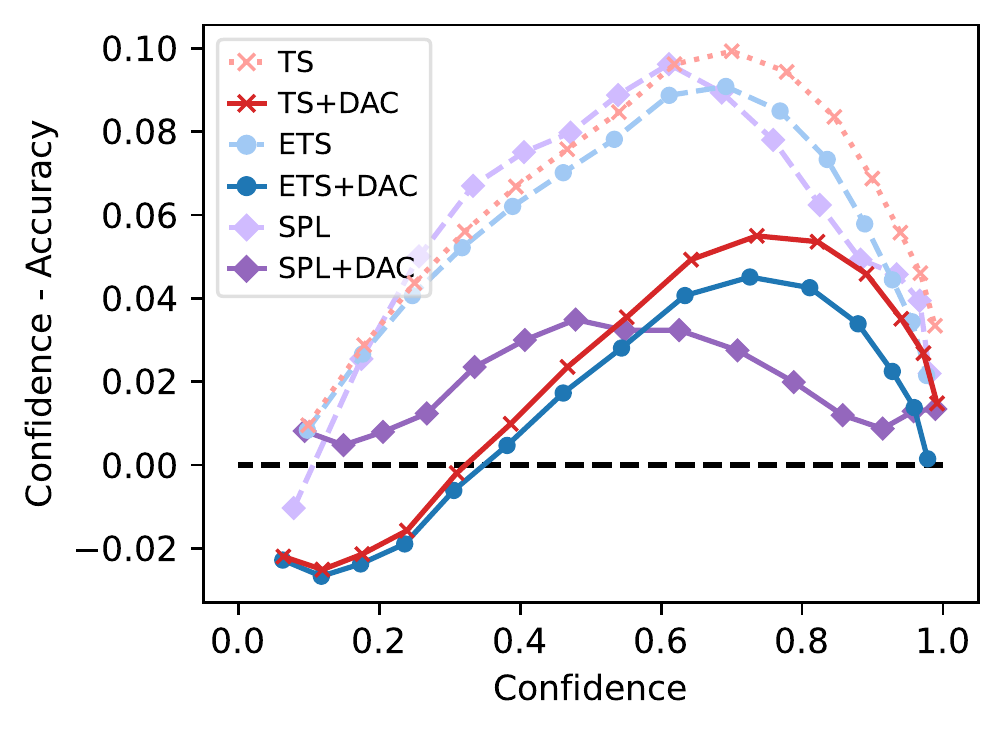} & 
    \includegraphics[width=0.30\textwidth]{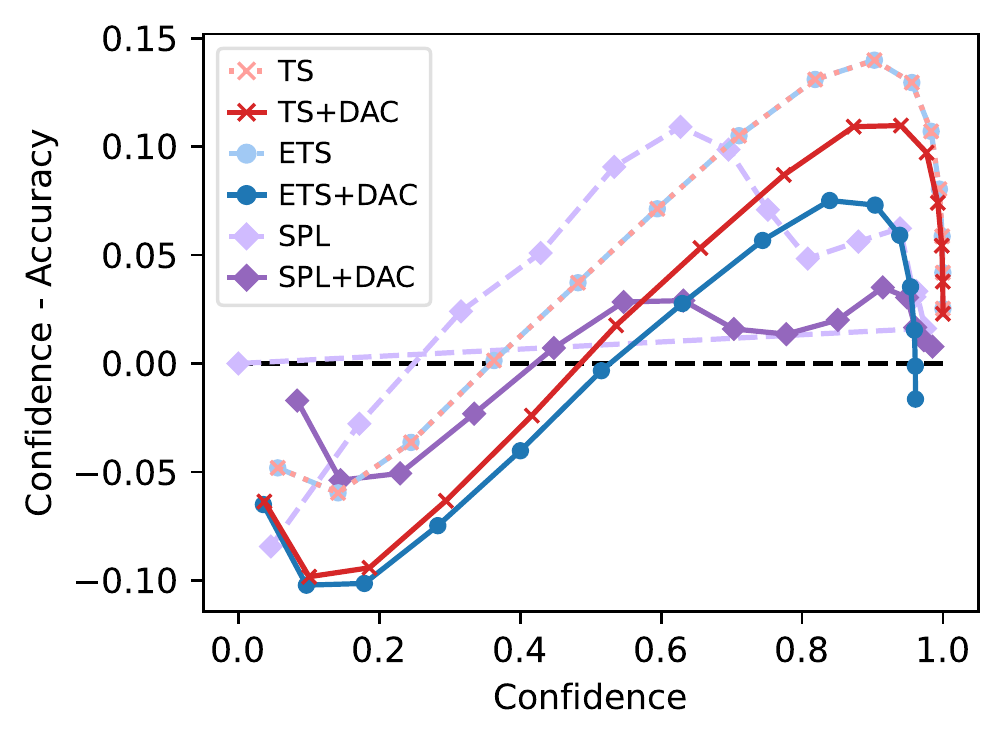} & 
    \includegraphics[width=0.30\textwidth]{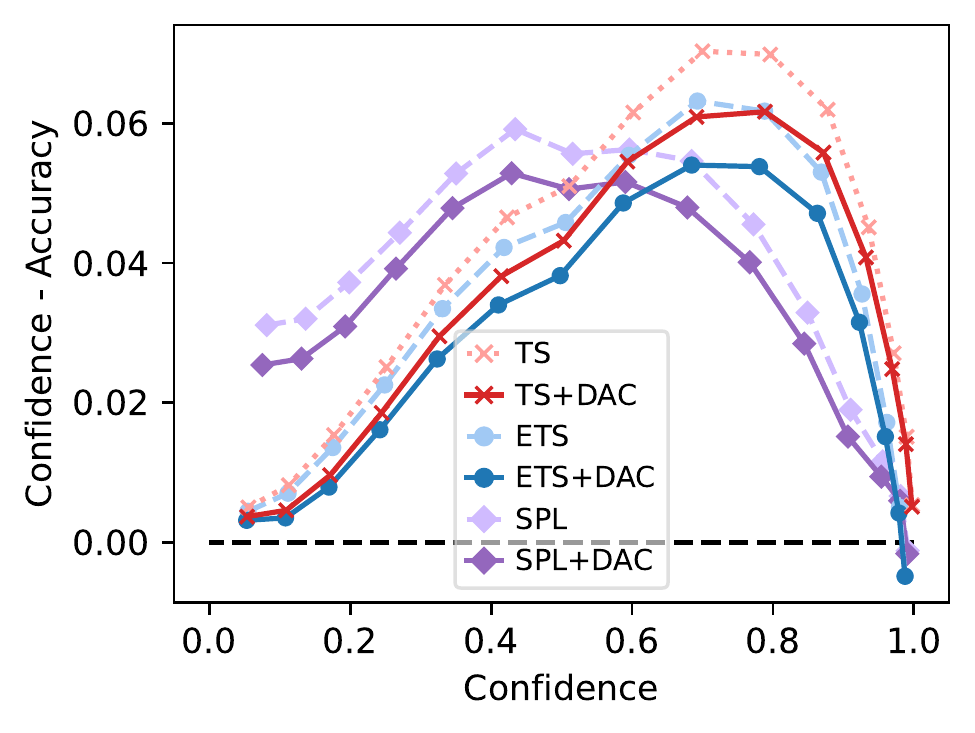} \\  
    \multicolumn{1}{c}{ImageNet-BiT-M} & \multicolumn{1}{c}{ImageNet-ResNeXt-WSL} & \multicolumn{1}{c}{ImageNet-ViT-B}\\   
    \end{tabular}
	\caption{Reliability Diagrams for all corruptions and severity levels combined (with equal mass-binning for 15 bins).}
\label{fig:appendix_rel_diagram}
\end{figure*}

In Fig.~\ref{fig:appendix_rel_diagram}, we show the reliability diagrams for 6 different dataset-classifier pairs on CIFAR-C or ImageNet-C dataset, across all corruptions and severity levels combined in one reliability diagram. The figure shows that our post-hoc calibration method DAC successfully boosts the existing post-hoc methods' calibration performance in domain-shift scenarios.

\section{Additional Results: OOD Scenarios} \label{asec:ood_results}

\begin{figure*}[t!]
    \centering
    \begin{tabular}{ccc}
    \includegraphics[width=0.45\textwidth]{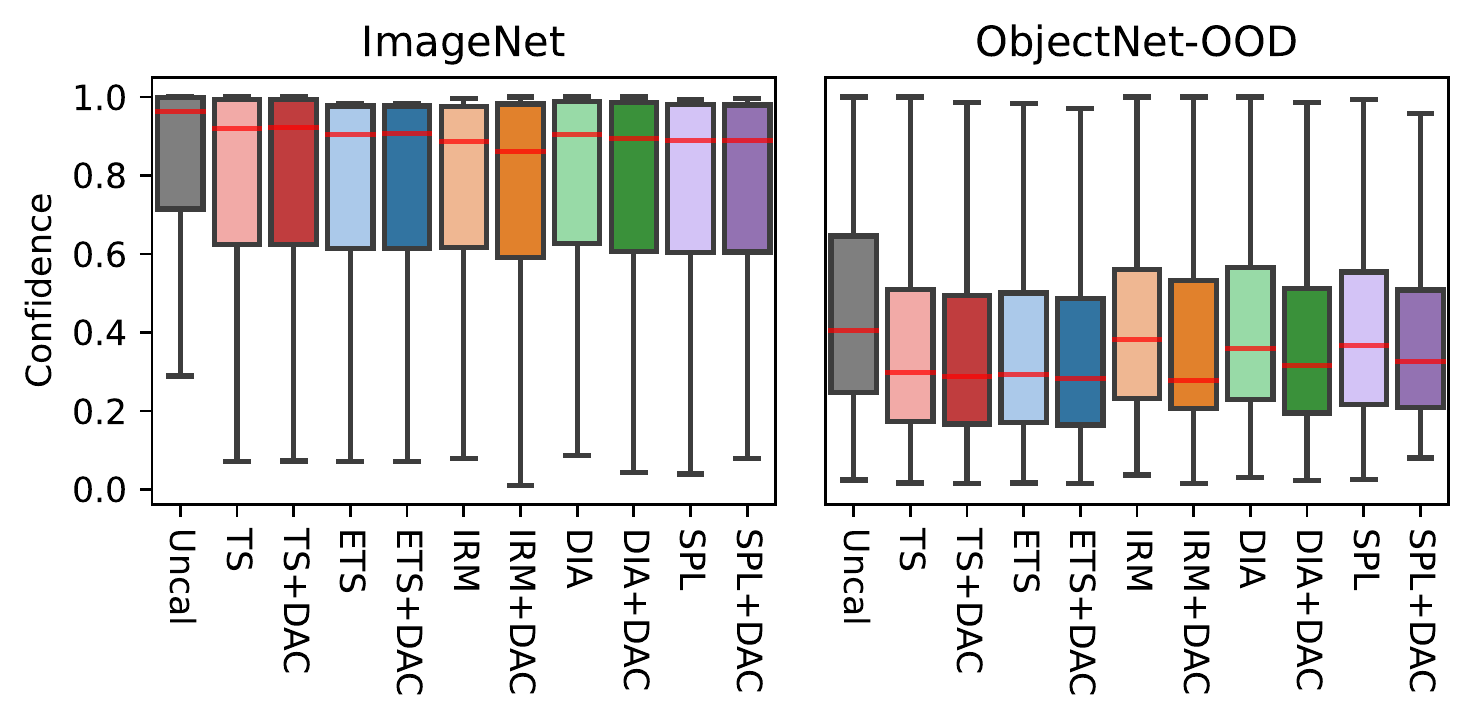} & 
    \includegraphics[width=0.45\textwidth]{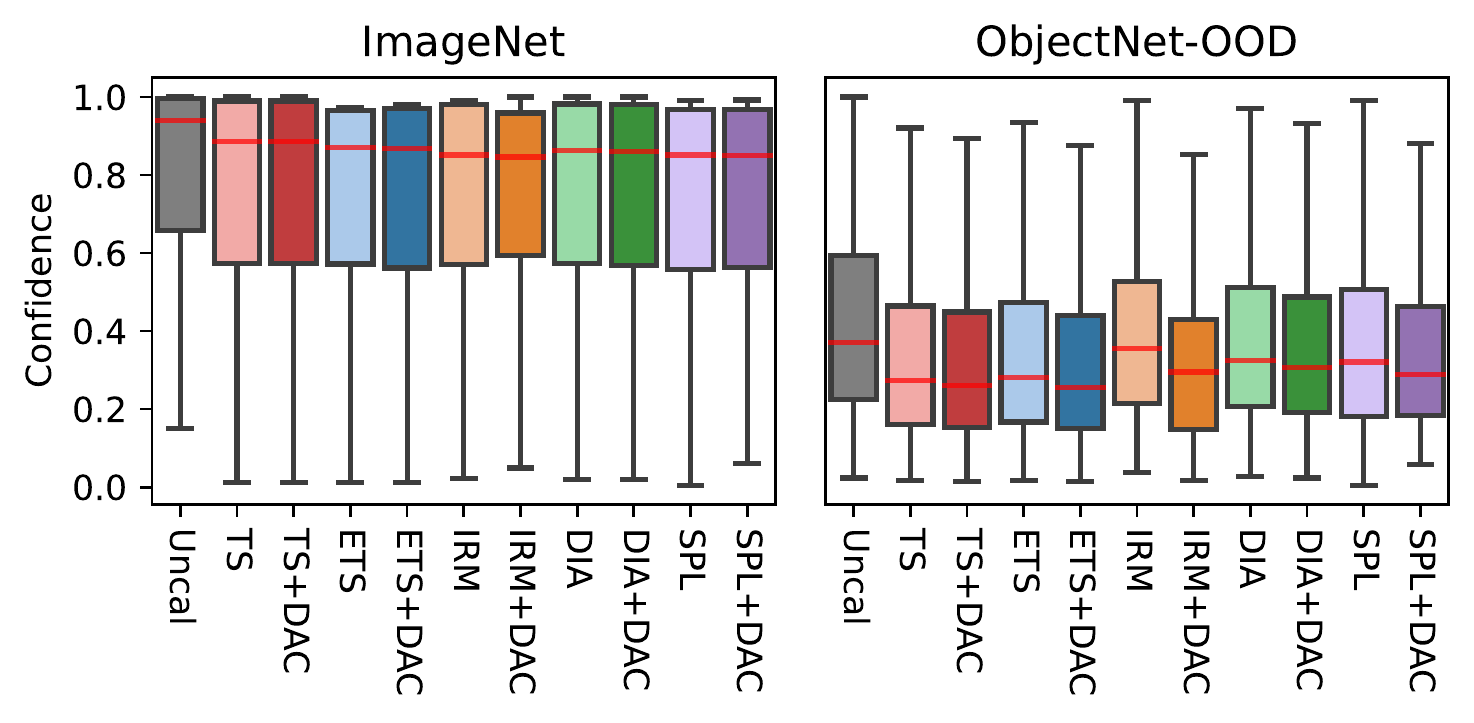}
    \\ 
    \multicolumn{1}{c}{ImageNet-ResNet152} & \multicolumn{1}{c}{ImageNet-DenseNet169}
    \\
    \includegraphics[width=0.45\textwidth]{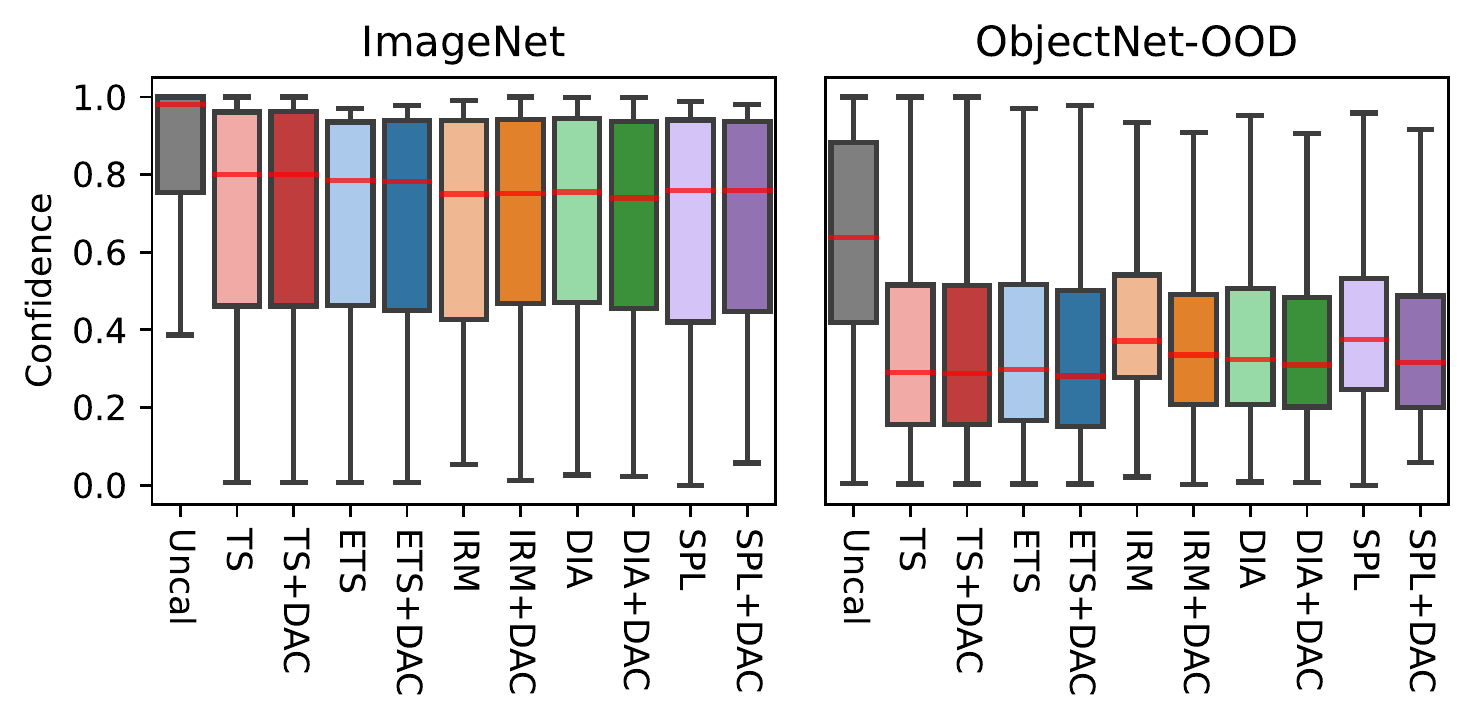}
    &
    \includegraphics[width=0.45\textwidth]{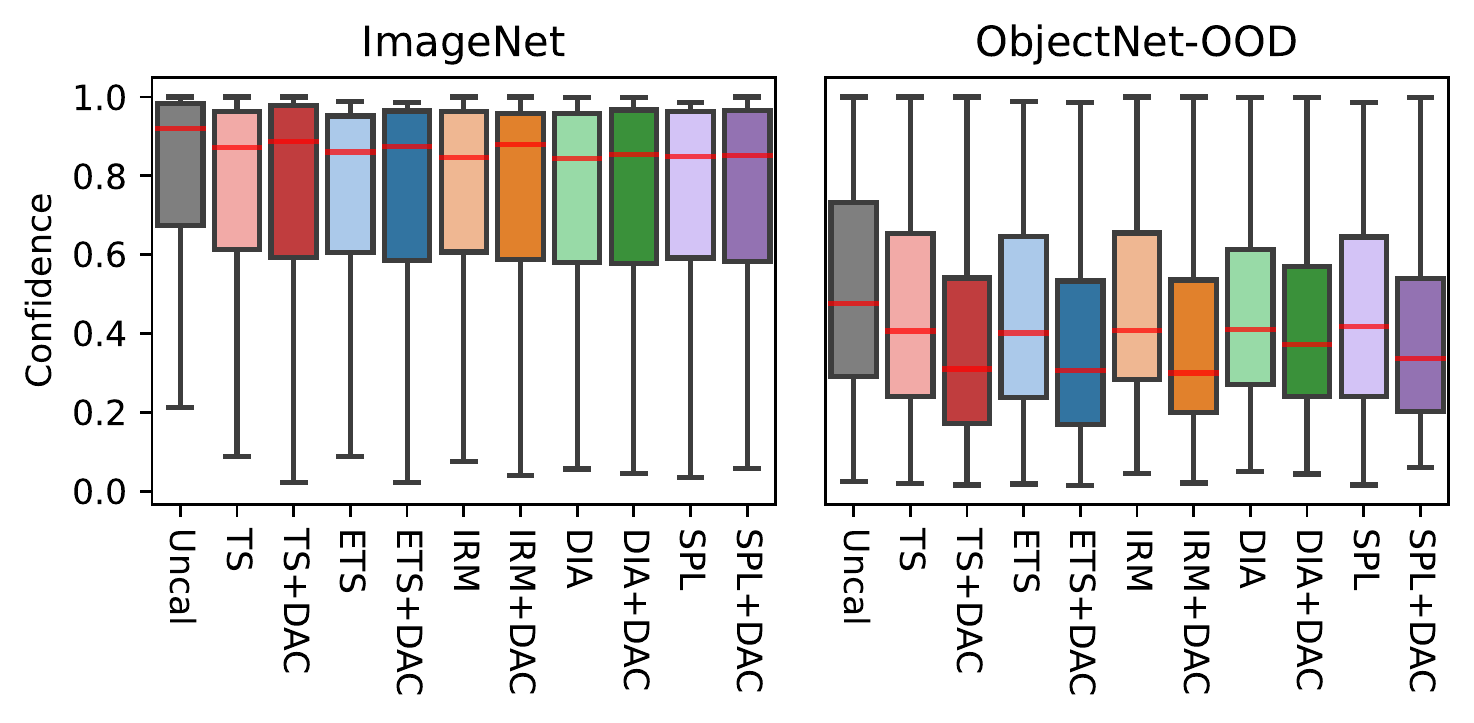}
    \\ 
    \multicolumn{1}{c}{ImageNet-Xception} & \multicolumn{1}{c}{ImageNet-BiT-M} 
    \\  
    \includegraphics[width=0.45\textwidth]{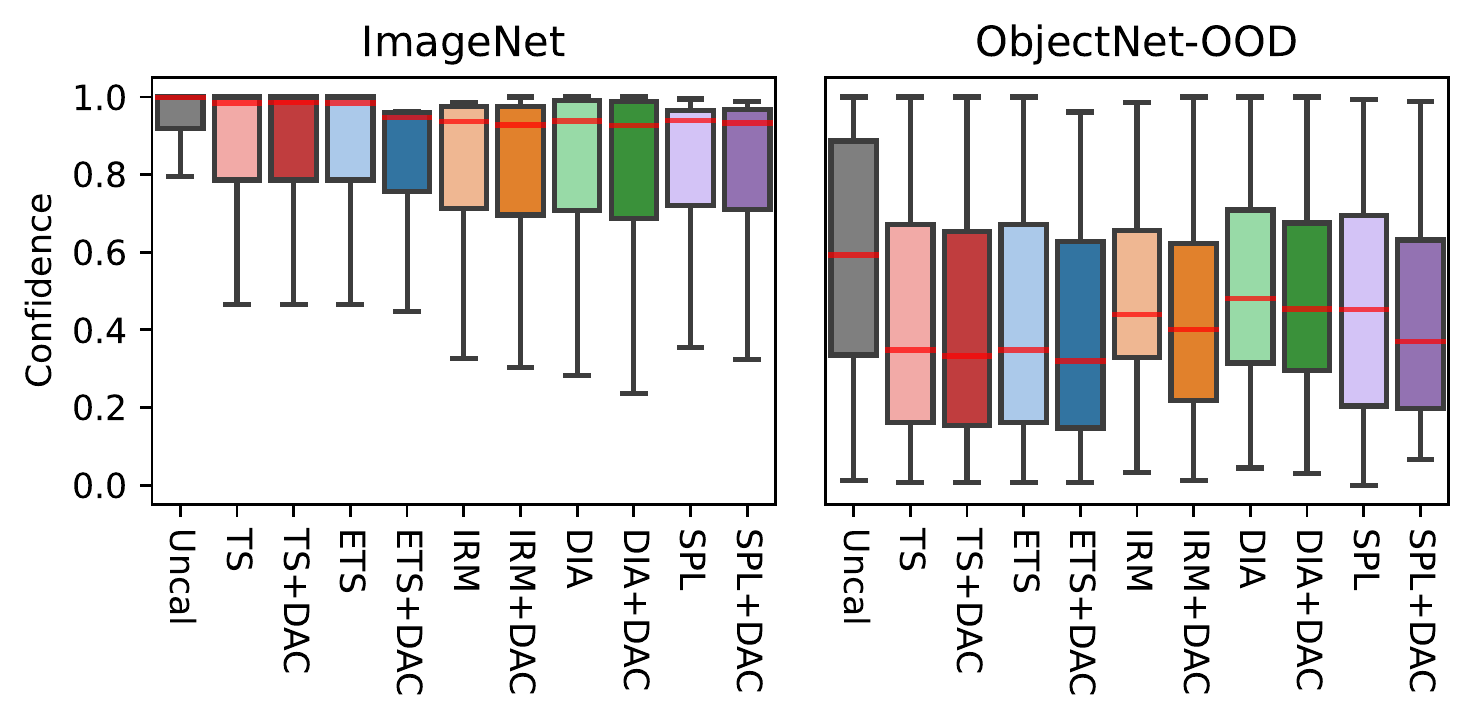} & 
    \includegraphics[width=0.45\textwidth]{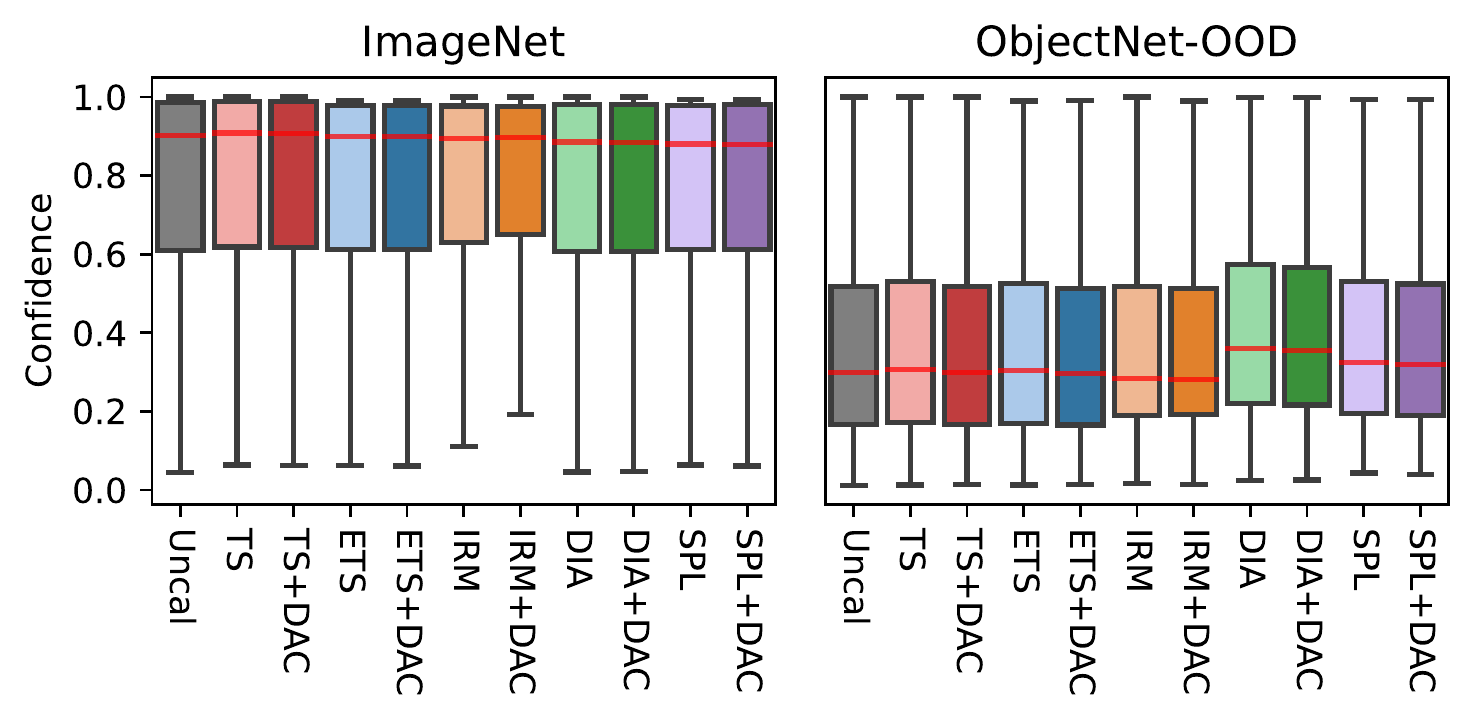} 
    \\  
    \multicolumn{1}{c}{ImageNet-ResNeXt-WSL} & \multicolumn{1}{c}{ImageNet-ViT-B}\\   
    \end{tabular}
	\caption{Boxplots of confidence values for ImageNet test set, and ObjectNet-OOD dataset. By combining the existing calibration methods with our DAC method, all classifiers output lower confidence values for novel-class images.}
\label{fig:appendix_boxplot_objectnet_conf}
\end{figure*}

Here, we include additional results from the OOD experiments. Fig.~\ref{fig:appendix_boxplot_objectnet_conf} summarizes the in-domain/OOD confidence distributions in various cases, where we see that DAC can in general better separate the in-domain data from OOD data, which is desirable. Tab.~\ref{tab:objectnet_grid} provides a more comprehensive summary of the quantitative OOD results for various network backbones. We observe consistent improvement with DAC in majority of the cases.

Note that BiT-M, ResNeXt-WSL, and ViT-B are pre-trained with additional data, as mentioned in Section~\ref{sec:exp_setup}. This might influence the outcome and the validity of the OOD experimental setup. The effect of backbone pre-training on the OOD tasks could be an interesting topic to investigate for future work.

\begin{table*}[h]
\vskip 0.15in
\caption{Additional OOD results using ImageNet-1k/ObjectNet-OOD as in-
domain/OOD test set, respectively.}
\vskip 0.15in
    \label{tab:objectnet_grid}
    \begin{small}
    \begin{sc}
        \begin{tabular}{lccccc}
        \toprule
        {} &  FPR &  Det. &    AU- &  AUPR- &  AUPR- \\
        {} &  @95\% ↓ &  Err ↓ &    ROC ↑ &  In ↑ &  Out ↑ \\
        \midrule
        TS & 21.52 & 23.18 & 84.75 & 80.41 & 86.83 \\
         +DAC & \textbf{21.03} & \textbf{22.70} & \textbf{85.27} & \textbf{81.20} & \textbf{87.30} \\ \hline
        ETS & 21.52 & 23.18 & 84.75 & \textbf{80.41} & 86.83 \\
         +DAC & \textbf{21.36} & \textbf{22.70} & \textbf{85.24} & 80.11 & \textbf{87.29} \\  \hline
        IRM & 16.10 & 21.94 & 83.55 & 80.37 & 86.32 \\
         +DAC & \textbf{14.90} & \textbf{21.56} & \textbf{85.19} & \textbf{80.75} & \textbf{87.25} \\ \hline
        DIA & 22.15 & 24.61 & 83.38 & 79.07 & 85.65 \\
         +DAC & \textbf{21.06} & \textbf{23.52} & \textbf{84.49} & \textbf{80.34} & \textbf{86.64} \\ \hline
        SPL & 21.70 & 24.54 & 83.45 & 79.29 & 85.72 \\
         +DAC & \textbf{21.13} & \textbf{22.69} & \textbf{85.23} & \textbf{81.25} & \textbf{87.22} \\
        \bottomrule
        \multicolumn{4}{c}{a.) ResNet152}
    \end{tabular}
    \end{sc}
    \end{small}
\medskip
    \begin{small}
    \begin{sc}
        \begin{tabular}{lccccc}
        \toprule
        {} &  FPR &  Det. &    AU- &  AUPR- &  AUPR- \\
        {} &  @95\% ↓ &  Err ↓ &    ROC ↑ &  In ↑ &  Out ↑ \\
        \midrule
        TS & 21.47 & 22.86 & 84.92 & 81.14 & 86.87 \\
         +DAC & \textbf{20.73} & \textbf{22.30} & \textbf{85.50} & \textbf{81.86} & \textbf{87.46} \\ \hline
        ETS & 21.92 & 23.15 & 84.72 & 79.82 & 86.72 \\
         +DAC & \textbf{20.90} & \textbf{22.29} & \textbf{85.51} & \textbf{82.16} & \textbf{87.47} \\ \hline
        IRM & 21.51 & 22.31 & 83.62 & 81.22 & 86.10 \\
         +DAC & \textbf{4.87} & \textbf{20.72} & \textbf{85.81} & \textbf{82.73} & \textbf{87.90} \\ \hline
        DIA & 22.09 & 24.26 & 83.67 & 79.86 & 85.77 \\
         +DAC & \textbf{21.36} & \textbf{23.64} & \textbf{84.35} & \textbf{80.63} & \textbf{86.47} \\ \hline
        SPL & 21.66 & 24.38 & 83.65 & 79.95 & 85.78 \\
         +DAC & \textbf{20.89} & \textbf{22.30} & \textbf{85.54} & \textbf{82.05} & \textbf{87.56} \\
        \bottomrule
        \multicolumn{4}{c}{b.) DenseNet169}
        \end{tabular}
    \end{sc}
    \end{small}
\hfil
    \begin{small}
    \begin{sc}
        \begin{tabular}{lccccc}
        \toprule
        {} &  FPR &  Det. &    AU- &  AUPR- &  AUPR- \\
        {} &  @95\% ↓ &  Err ↓ &    ROC ↑ &  In ↑ &  Out ↑ \\
    \midrule
    TS & 28.89 & 28.57 & 77.83 & 71.57 & 79.28 \\
     +DAC & \textbf{28.71} & \textbf{28.51} & \textbf{77.91} & \textbf{71.71} & \textbf{79.35} \\ \hline
    ETS & 28.87 & 28.66 & 77.75 & 71.53 & 79.23 \\
     +DAC & \textbf{28.67} & \textbf{28.50} & \textbf{77.91} & \textbf{71.79} & \textbf{79.35} \\ \hline
    IRM & 19.95 & 28.85 & 74.82 & 71.89 & 77.33 \\
     +DAC & \textbf{15.70} & \textbf{27.58} & \textbf{78.09} & \textbf{73.02} & \textbf{79.51} \\ \hline
    DIA & 28.36 & 28.09 & 78.84 & 74.98 & 80.09 \\
     +DAC & \textbf{28.27} & \textbf{27.82} & \textbf{79.04} & \textbf{74.96} & \textbf{80.25} \\ \hline
    SPL & 30.90 & 30.83 & 75.36 & 70.72 & 77.21 \\
     +DAC & \textbf{28.71} & \textbf{28.50} & \textbf{77.90} & \textbf{71.56} & \textbf{79.31} \\
        \bottomrule
        \multicolumn{4}{c}{c.) Xception}
        \end{tabular}
    \end{sc}
    \end{small}
\medskip
    \begin{small}
    \begin{sc}
        \begin{tabular}{lccccc}
        \toprule
        {} &  FPR &  Det. &    AU- &  AUPR- &  AUPR- \\
        {} &  @95\% ↓ &  Err ↓ &    ROC ↑ &  In ↑ &  Out ↑ \\
    \midrule
    TS & 26.42 & 28.59 & 78.47 & 70.72 & 81.88 \\
     +DAC & \textbf{23.68} & \textbf{25.23} & \textbf{82.79} & \textbf{78.77} & \textbf{84.99} \\ \hline
    ETS & 26.42 & 28.59 & 78.47 & 70.72 & 81.88 \\
     +DAC & \textbf{23.69} & \textbf{25.21} & \textbf{82.82} & \textbf{78.96} & \textbf{85.00} \\ \hline
    IRM & 33.44 & 27.43 & 77.73 & 72.45 & 81.79 \\
     +DAC & \textbf{9.14} & \textbf{23.54} & \textbf{82.84} & \textbf{79.41} & \textbf{85.19} \\ \hline
    DIA & 26.67 & 28.26 & 79.39 & 74.55 & 82.06 \\
     +DAC & \textbf{25.32} & \textbf{26.58} & \textbf{81.40} & \textbf{77.92} & \textbf{83.64} \\ \hline
    SPL & 26.66 & 29.01 & 78.07 & 70.84 & 81.45 \\
     +DAC & \textbf{23.64} & \textbf{25.25} & \textbf{82.81} & \textbf{79.31} & \textbf{84.96} \\
        \bottomrule
        \multicolumn{4}{c}{d.) BiT-M}
        \end{tabular}
    \end{sc}
    \end{small}
\hfil
    \begin{small}
    \begin{sc}
        \begin{tabular}{lccccc}
        \toprule
        {} &  FPR &  Det. &    AU- &  AUPR- &  AUPR- \\
        {} &  @95\% ↓ &  Err ↓ &    ROC ↑ &  In ↑ &  Out ↑ \\
    \midrule
    TS & 22.52 & 25.47 & 82.30 & 75.30 & 85.83 \\
     +DAC & \textbf{22.05} & \textbf{25.05} & \textbf{82.83} & \textbf{76.00} & \textbf{86.26} \\ \hline
    ETS & \textbf{22.52} & 25.47 & 82.30 & 75.30 & 85.83 \\
     +DAC & 23.15 & \textbf{25.05} & \textbf{83.07} & \textbf{78.77} & \textbf{86.33} \\ \hline
    IRM & 23.20 & 25.61 & 80.66 & \textbf{77.06} & 84.80 \\
     +DAC & \textbf{10.86} & \textbf{23.50} & \textbf{82.75} & 76.13 & \textbf{86.06} \\ \hline
    DIA & 25.19 & 27.96 & 79.37 & 72.47 & 82.84 \\
     +DAC & \textbf{24.77} & \textbf{27.48} & \textbf{79.94} & \textbf{73.10} & \textbf{83.37} \\ \hline
    SPL & 28.23 & 26.69 & 80.84 & \textbf{76.46} & 84.17 \\
     +DAC & \textbf{22.63} & \textbf{25.04} & \textbf{82.84} & 75.44 & \textbf{86.29} \\
        \bottomrule
        \multicolumn{4}{c}{e.) ResNext-WSL}
        \end{tabular}
    \end{sc}
    \end{small}
\medskip
    \begin{small}
    \begin{sc}
        \begin{tabular}{lccccc}
        \toprule
        {} &  FPR &  Det. &    AU- &  AUPR- &  AUPR- \\
        {} &  @95\% ↓ &  Err ↓ &    ROC ↑ &  In ↑ &  Out ↑ \\
    \midrule
    TS & 22.52 & 24.16 & 83.77 & 79.95 & 85.54 \\
     +DAC & \textbf{22.15} & \textbf{23.75} & \textbf{84.24} & \textbf{80.74} & \textbf{85.95} \\ \hline
    ETS & 22.52 & 24.16 & 83.77 & 79.95 & 85.54 \\
     +DAC & \textbf{22.18} & \textbf{23.76} & \textbf{84.22} & \textbf{80.42} & \textbf{85.94} \\ \hline
    IRM & 20.76 & 24.01 & 83.69 & 79.31 & 85.30 \\
     +DAC & \textbf{5.42} & \textbf{21.22} & \textbf{83.87} & \textbf{80.46} & \textbf{85.48} \\ \hline
    DIA & 23.53 & 25.73 & 82.22 & 78.12 & 84.50 \\
     +DAC & \textbf{23.28} & \textbf{25.50} & \textbf{82.53} & \textbf{78.63} & \textbf{84.77} \\ \hline
    SPL & \textbf{21.55} & 23.98 & 83.90 & 80.09 & 85.63 \\
     +DAC & 22.12 & \textbf{23.73} & \textbf{84.21} & \textbf{80.77} & \textbf{85.88} \\ 
        \bottomrule
        \multicolumn{4}{c}{f.) ViT-B}
        \end{tabular}
    \end{sc}
    \end{small}
\end{table*}

\clearpage

\section{Data Efficiency Analysis}

\subsection{Additional Results}

We show additional data efficiency diagrams for models on CIFAR10 and CIFAR100. As for ImageNet in the main text, we encounter no drastic sensitivity to validation set size. We conclude that these post-hoc methods combined with DAC can be trained on very small validation set sizes. 

\begin{figure*}[htbp!]
    \centering
    \begin{tabular}{cc}
    \includegraphics[width=0.47\textwidth]{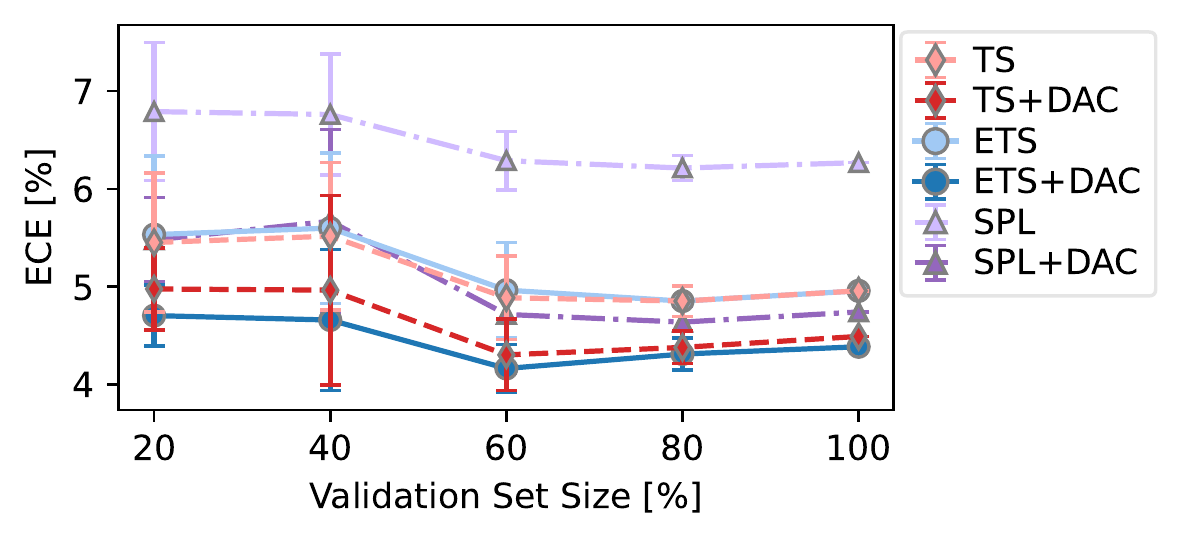} & 
    \includegraphics[width=0.47\textwidth]{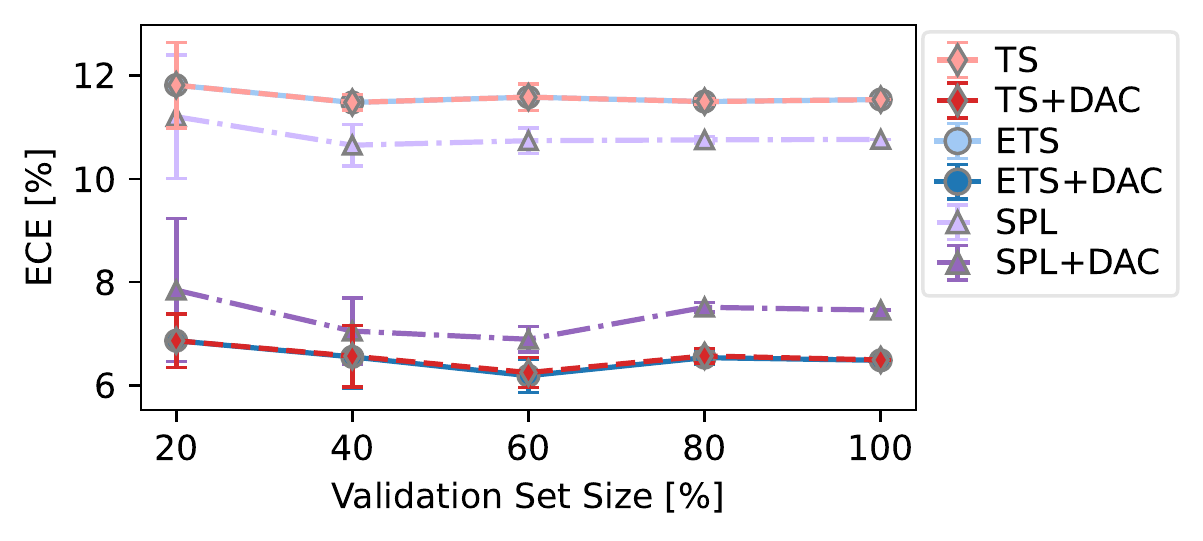} \\  
    \multicolumn{1}{c}{CIFAR10-ResNet18} & \multicolumn{1}{c}{CIFAR100-VGG16}\\  
    \end{tabular}
	\caption{Data efficiency diagrams for macro-averaged ECE ($\times 10^2$) (with 15 bins across all corruptions from severity=0 (in-domain) until severity=5 (heavily corrupted)) from 10\% to 100\% validation set size.}
\label{fig:appendix_data_eff}
\end{figure*}

\subsection{Trade-off between in-domain and domain-shift performance}
\begin{figure*}[htbp!]
    \centering
    \begin{tabular}{cc}
    \includegraphics[width=0.47\textwidth]{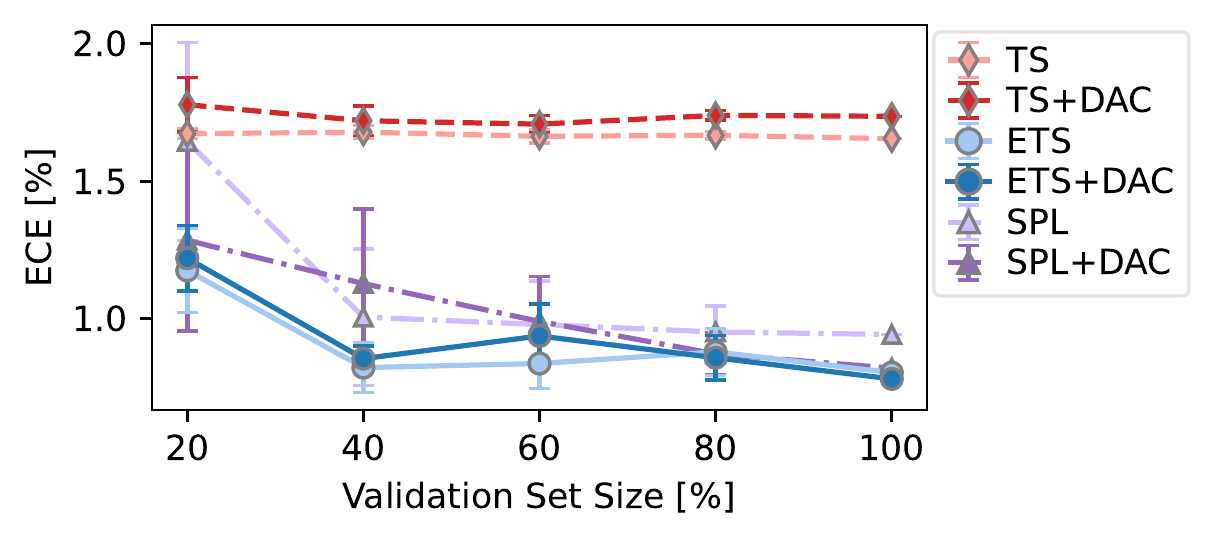} & 
    \includegraphics[width=0.47\textwidth]{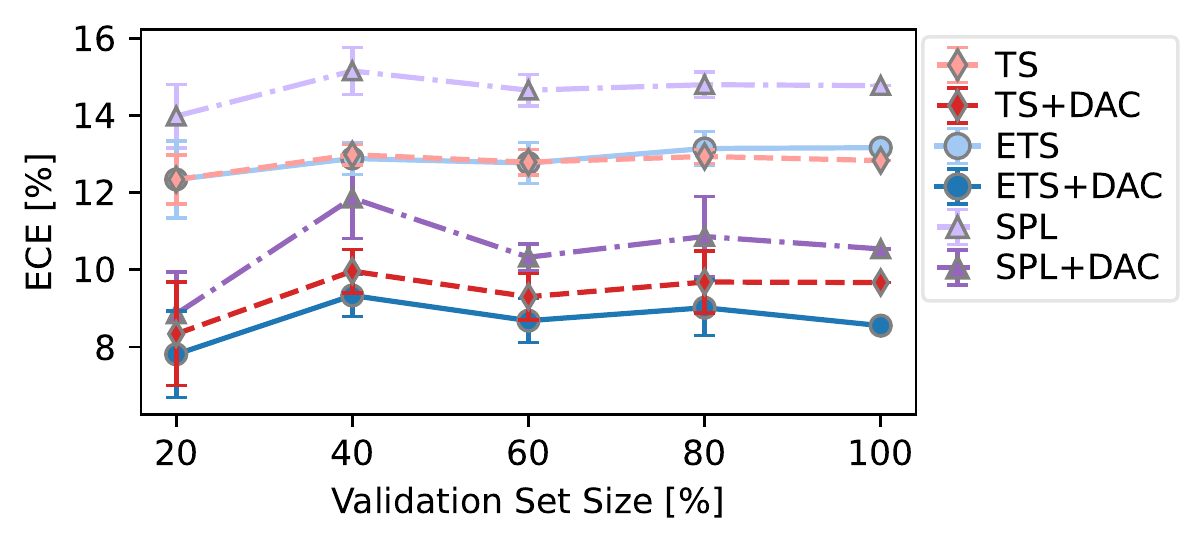} \\  
    \multicolumn{1}{c}{IMG-DenseNet169 (In-domain)} & \multicolumn{1}{c}{IMG-DenseNet169 (Sev. 5)}\\  
    \end{tabular}
    \caption{Data efficiency diagrams for ECE ($\times 10^2$) with 15 bins from 10\% to 100\% validation set size. Left: image corruption severity=0 (in-domain). Right: image corruption severity=5 (heavily corrupted)). The figures show the trade-off between in-domain and out-of-domain calibration errors of DAC.}
\label{fig:trade-off_dac}
\end{figure*}

We observe that the larger the validation set gets, the more likely the calibration method is to overfit to in-domain data, leading to a degradation in performance on domain shift and OOD data. 
In Figure~\ref{fig:trade-off_dac}, we show the data efficiency diagrams for ECE ($\times 10^2$) with 15 bins from 10\% to 100\% validation set size, for the in-domain case and the case of corruption severity=5 (heavily corrupted), respectively.
We show that while the in-domain calibration performance benefits from the size of the validation set, the domain-shift calibration performance is worsen due to overfitting.
Fig.~\ref{fig:data_efficiency_lineplot} in the main text shows the mean ECE across all test-domain shift scenarios, including in-domain, and thus incorporates this overfitting behavior on large validation set sizes. 

\section{Computational Cost}
In this section, we compare the computational cost of DAC and the existing methods. In short, the computational cost of DAC is similar to that of existing post-hoc methods during both the training and inference phases. 

We first compare the training speed of DAC, DIA, ETS, and SPL for DenseNet169 trained on ImageNet, using an NVIDIA Titan X (12GB) GPU. The table below shows that DAC achieves a total training time of only about 14 minutes. Compared to the overall training process of the classifier, which takes at least several hours, the additional computational cost added by DAC is minor.

\begin{table}[!ht]
\centering
\begin{tabular}{cc|ccc}
\toprule
 & Total Training time (s) & \begin{tabular}[c]{@{}c@{}}1. Extract features \\ for train set (s)\end{tabular} & \begin{tabular}[c]{@{}c@{}}2. Extract features/logits\\  for validation set (s)\end{tabular} & 3. Optimization (s) \\ \midrule
DAC & 825.98 & 175.72 & 452.91 (include KNN) & 197.35 \\  \hline
DIA & 11090.09 & - & 130.49 & 10959.60 \\
ETS & 149.35 & - & 130.49 & 18.86 \\ 
SPL & 131.04 & - & 130.49 & 0.55 \\ 
\bottomrule
\end{tabular}
\caption{Training time comparison: Our proposed method DAC vs. existing post-hoc calibration methods (DIA, ETS, SPL).}
\label{fig:computation_cost_training}
\end{table}

In addition, we compare the per-sample inference speed. As shown in the table below, we found that the inference speed of DenseNet169 combined with DAC is comparable to the existing calibration methods. This can be attributed to the efficient GPU-accelerated KNN search in DAC, enabled by the Faiss library \footnote{https://github.com/facebookresearch/faiss} \cite{johnson2019billion}.

\begin{table}[!ht]
    \centering
    \begin{tabular}{cc}
    \toprule
        ~ & Total per-sample inference time (ms) \\
        \midrule
        DAC & 37.12 (+-0.77) \\ \hline
        DIA & 38.23 (+-0.97) \\
        ETS & 34.05 (+-0.96) \\ 
        SPL & 34.00 (+-0.96) \\ 
    \bottomrule
    \end{tabular}
\caption{Per-sample inference time comparison: Our proposed method DAC vs. existing post-hoc calibration methods (DIA, ETS, SPL).}
\label{fig:computation_cost_inference}
\end{table}

\end{document}